\documentclass[10pt,twocolumn,letterpaper]{article}

\usepackage{cvpr}              %

\usepackage{booktabs}
\usepackage{xspace}
\usepackage{multirow}
\usepackage{amsmath}
\usepackage{amssymb}
\usepackage{pifont}
\usepackage{colortbl}
\usepackage{tablefootnote}
\usepackage{makecell}
\usepackage{wrapfig}
\usepackage{caption}
\usepackage{pgfplotstable}
\usepackage[dvipsnames]{xcolor}
\usepackage[hang,flushmargin]{footmisc} %

\newcommand{\IDDDataset}{ISSU\xspace}%
\newcommand{\IDDDatasetTrain}{\IDDDataset-{Train}\xspace}%
\newcommand{\IDDDatasetTemporal}{\IDDDataset-Test-Temporal\xspace}%
\newcommand{\IDDDatasetStatic}{\IDDDataset-Test-Static\xspace}%

\newcommand{\cmark}{\ding{51}}%
\newcommand{\xmark}{\ding{55}}%

\newcommand{\ph}{\phantom{0}}

\newcommand{\boldparagraph}[1]{\noindent{\bf #1}}
\newcommand{\scatterwidthstats}{0.246\textwidth}
\newcommand{\scatterheightstats}{0.25\textwidth}

\pgfplotsset{compat = 1.3}

\newenvironment{customlegend}[1][]{%
    \begingroup
    \csname pgfplots@init@cleared@structures\endcsname
    \pgfplotsset{#1}%
}{%
    \csname pgfplots@createlegend\endcsname
    \endgroup
}%
\def\addlegendimage{\csname pgfplots@addlegendimage\endcsname}

\definecolor{cvprblue}{rgb}{0.21,0.49,0.74}
\usepackage[pagebackref,breaklinks,colorlinks,allcolors=cvprblue]{hyperref}

\usepackage{silence}
\makeatletter
\robustify\@latex@warning@no@line
\makeatother
\usepackage{authblk}

\newcommand\blfootnote[1]{%
  \begingroup
  \renewcommand\thefootnote{}\footnote{#1}%
  \addtocounter{footnote}{-1}%
  \endgroup
}

\def\paperID{4945} %
\def\confName{CVPR}
\def\confYear{2025}

\title{A Dataset for Semant\underline{i}c \underline{S}egmentation in the Pre\underline{s}ence of \underline{U}nknowns\vspace{-0.5em}}

\begin{document}

\author[a]{Zakaria Laskar$^{\ast,}$}
\author[a]{Tomáš Vojíř$^{\ast,}$}
\author[b]{Matej Grcic$^{\ast,}$}
\author[c]{Iaroslav Melekhov$^{\bullet,}$}
\author[e]{Shankar Gangisetty}
\author[c,d]{Juho~Kannala}
\author[a]{Jiri Matas}
\author[a]{Giorgos Tolias}
\author[e]{C.V. Jawahar}

\affil[a]{VRG, FEE, Czech Technical University in Prague, Czechia}
\affil[b]{University of Zagreb Faculty of Electrical Engineering and Computing, Croatia}
\affil[c]{Aalto University, Finland}
\affil[e]{IIIT Hyderabad, India}
\affil[d]{University of Oulu, Finland}

\twocolumn[{%
\renewcommand\twocolumn[1][]{#1}%
\maketitle
\begin{center}
    \vspace{-1.5em}
    \centering
    \captionsetup{type=figure}
    \includegraphics[width=0.96\textwidth]{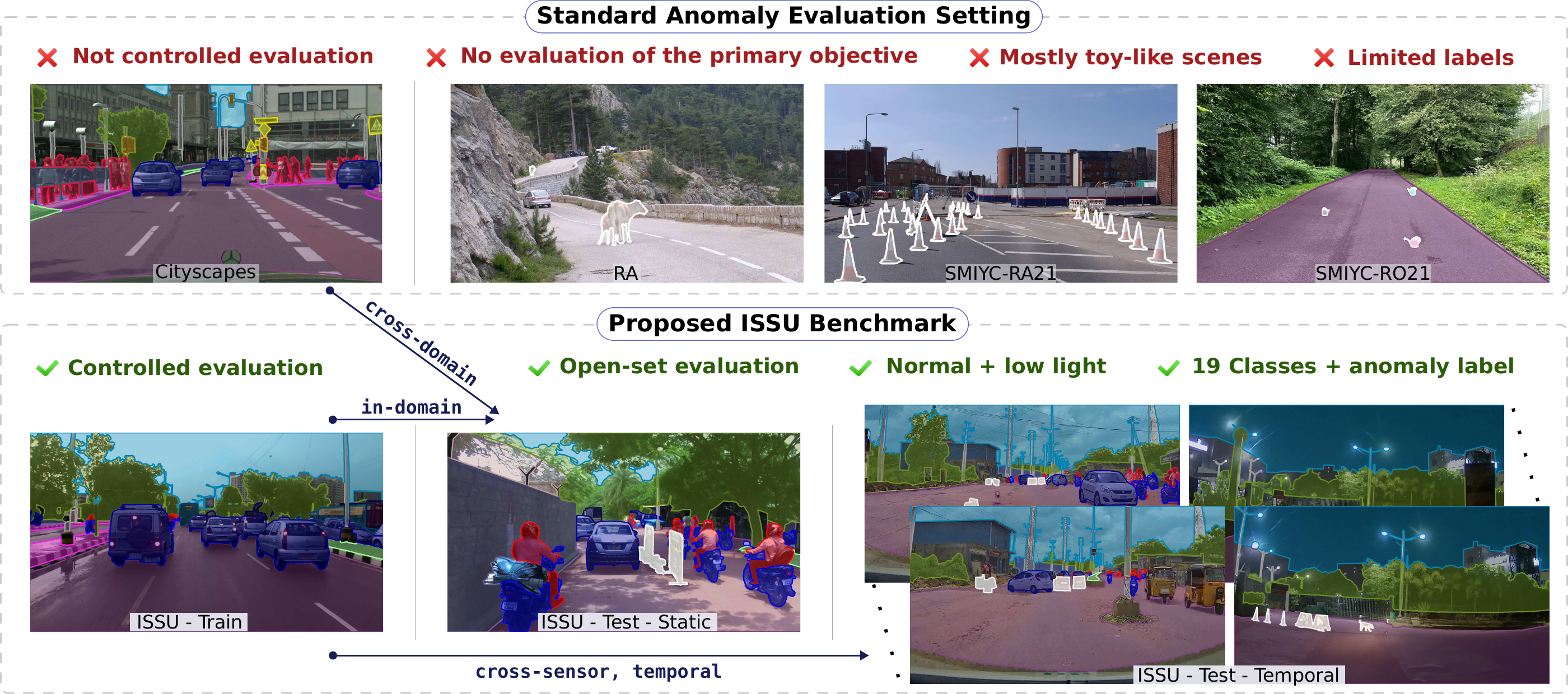}
    \captionof{figure}{Standard benchmarks cannot separate the effects of domain shift, lighting conditions, and anomaly size during evaluation. The proposed dataset allows controlled evaluation of these effects and supports evaluation of both closed-set and anomaly segmentation.
    }
    \label{fig:teaser}
\end{center}%
}] 
\maketitle

\blfootnote{$^\ast$ Equal Contribution. The corresponding author {\it zakaria.nits@gmail.com}\\$^\bullet$ The work was done prior to joining Amazon}

\vspace{-10pt}
\begin{abstract}
Before deployment in the real-world deep neural networks require
thorough evaluation of how they handle both knowns, inputs represented in the training data,  and unknowns (anomalies). 
This is especially important for scene understanding tasks with safety critical
applications, such as in autonomous driving. Existing datasets allow evaluation
of only knowns or unknowns - but not both, which is required to establish
``in the wild" suitability of deep neural network models.
To bridge this
gap, we propose a novel anomaly segmentation dataset, ISSU, that features
a diverse set of anomaly inputs from cluttered real-world environments. 
The dataset is twice larger than existing anomaly segmentation datasets,
and provides a training, validation and test set for controlled in-domain evaluation. The test set consists
of a static and temporal part, with the latter comprised of videos. The
dataset provides annotations for both closed-set (knowns) and anomalies,
enabling closed-set and open-set evaluation.
The dataset covers diverse conditions, such as domain and cross-sensor shift, illumination variation
and allows ablation of anomaly detection methods with respect to these
variations. 
Evaluation results of current state-of-the-art methods
confirm the need for improvements especially in domain-generalization, small
and large object segmentation. 
The code and the dataset
are available at~{\footnotesize\url{https://github.com/vojirt/benchmark_issu}}.
\end{abstract}
    
\vspace{-5mm}
\section{Introduction} \label{sec:intro}
Many successful computer vision applications rely heavily or entirely on deep
neural networks trained on extensive, fully or partially labeled training
data~\cite{everingham10ijcv,cordts16cvpr,zhou19ijcv,kuznetsova20ijcv}. 
The validation part of the training process provides performance 
estimates for situations well covered in the training data.

However, when exposed to data not well represented in training, predictions of a 
deep neural network model may become arbitrary. Therefore, a crucial capability for these
models is the ability to detect such \textit{unknown} or {\it anomalous
inputs}\footnote{We use these term interchangeably.}. There are multiple
reasons why an input may be ``unknown'' -- it might be a rare case, belonging
to the long-tail missed for statistical reasons, a result of domain shifts such
as introduction of a  novel classes (\eg, a segway on a highway) or a result of
optical sensor defects (\eg, a broken or dirty lens in a surveillance camera).
 
Deep neural networks, which lack the ability to recognize unknowns, assign
to these anomalous inputs a label that corresponds to one of the known classes of
the training set, potentially with high
confidence~\cite{hendrycks17iclr}. This may result in suboptimal or even dangerous
behavior in deployed systems. Thus, the importance of anomaly detection is
critical in safety-sensitive applications, such as autonomous driving, where an
undetected anomaly could lead to accidents. 

Autonomous driving is a very complex task, with one of its core elements being
the perception of the environment surrounding the vehicle, often referred to as
scene understanding. Scene understanding is typically defined as a closed-set
semantic segmentation task, where each pixel in an image is assigned to one of
\textit{K} known classes. Progress in this area has been greatly advanced by
large semantic segmentation datasets~\cite{cordts16cvpr, varma19wacv, yu20cvpr,neuhold17iccv} along with
the development of powerful deep learning models~\cite{cheng22cvpr,chen18eccv} specifically
designed for semantic segmentation. However, these datasets overlook the
anomaly detection problem. Neither anomalous data nor evaluation protocols are
provided with their test sets, limiting the ability to evaluate segmentation
models in real-world settings where unknowns may occur. 

To address these limitations, specialized datasets focusing on anomaly
detection in driving scenarios have been developed, including LostAndFound~\cite{pinggera16iros},
Fishyscapes~\cite{blum21ijcv}, RoadAnomaly~\cite{lis19iccv}, and SMIYC~\cite{chan21neurips}. However, these datasets
typically use a binary evaluation approach (``known" \vs ``anomaly"), where all
pixels belonging to closed-set classes are assigned a single ``known" label,
while unknowns are labeled as ``anomaly". This approach diverges from   open
set $K+1$ evaluation  (``closed-set" \vs ``anomaly"), which is essential for
real-world applications. Moreover, these datasets lack in-domain training data
and are often collected under controlled conditions - usually in clear daylight
and in simplified environments such as empty roads or parking areas. This setup
leads to limited scene diversity and the absence of clutter from other traffic
actors.

In this paper, we introduce \IDDDataset, a fully annotated semantic
segmentation dataset that provides \textit{both} closed-set and anomaly labels
for images in the test set. This allows for the joint evaluation of closed-set
and open-set semantic segmentation, with the anomaly label forming
an additional class. Our dataset comprises real-world images collected from
roads in India, which, due to its unstructured traffic conditions, present
a wide range of anomalies in diverse sizes, shapes, lighting conditions, and
complex backgrounds cluttered with on-road traffic agents. \IDDDataset consists
of three parts: \IDDDatasetTrain, \IDDDatasetStatic and \IDDDatasetTemporal. \IDDDatasetTrain
includes training and validation sets, while \IDDDatasetStatic forms the test split for controlled in-domain
evaluation (\ie, with train and test data from the same distribution). \IDDDatasetTemporal contains temporal test data in the form of short
video clips collected using a different sensor setup than \IDDDatasetTrain.
Semantic annotations with anomaly labels, along with the specific design of
the dataset, allow controlled evaluations that isolate the effects of
various nuisance factors, such as different anomaly sizes, lighting conditions, and camera sensors. Beyond standard evaluations,
our dataset enables cross-domain evaluations, facilitating an analysis of how
models trained on datasets from structured environments, such as
Cityscapes~\cite{cordts16cvpr}, generalize to unstructured settings, and vice versa.
\begin{table}[t]
\centering
\resizebox{0.45\textwidth}{!}{%
    \begin{tabular}{@{}lcccc@{}l}
        \toprule

        \textbf{Dataset-Year} & 
        \begin{tabular}[c]{@{}c@{}}\textbf{Size}\\ \textbf{(annotated)}\end{tabular} & 
        \begin{tabular}[c]{@{}c@{}}\textbf{Weather/Env.}\\ \textbf{Cond.}\end{tabular} & 
        \textbf{Location} &
        \textbf{Clutter} \\ 

        \midrule

        AppoloScape'18~\cite{Huang2020}          & 145k & Diverse     & China  & High \\
        Mapillary Vistas'17~\cite{neuhold17iccv} & 25k  & Diverse     & World  & Diverse\\
        BDD100K'20~\cite{yu20cvpr}               & 10k  & Diverse     & US     & Low \\
        IDD'19~\cite{varma19wacv}                & 10k  & Good        & India  & High \\
        Cityscapes'16~\cite{cordts16cvpr}        & 5k   & Good        & Europe & Low \\
        WildDash 2'22~\cite{Zendel2022}          & 4.3k  & Diverse     & World  & Diverse \\
        ACDC'21~\cite{sakaridis21iccv}           & 4k   & Adverse     & Europe & Low \\

        \cmidrule{1-5}
        \IDDDatasetTrain'24                      & 3.4k  & Diverse     & India  & High \\

        \bottomrule
    \end{tabular}%
}%
    \caption{\textbf{Comparison of existing datasets for semantic segmentation} for
    driving scenarios.}
\vspace{-5mm}
\label{tab:overview_seq}
\end{table}%
\begin{table*}[htb]
\resizebox{\textwidth}{!}{%
    \begin{tabular}{@{}lcrccrccccc}
        \toprule

        \textbf{Dataset-Year} & \textbf{Domain} & \textbf{Size} & \textbf{Anom. Size} & 
        \textbf{Modality} & 
        \begin{tabular}[c]{@{}c@{}}\textbf{\%Anom.}\\ \textbf{Pixels}\end{tabular} & 
        \begin{tabular}[c]{@{}c@{}}\textbf{\%Non-Anom.}\\ \textbf{Pixels}\end{tabular} & 
        \textbf{Classes} & 
        \begin{tabular}[c]{@{}c@{}}\textbf{Weather/Env.}\\ \textbf{Cond.}\end{tabular} & 
        \textbf{Clutter} & 
        \textbf{oIoU} \\

        \midrule

        Street-hazards'22~\cite{hendrycks22icml}   & Synthetic & 1500 & Diverse & Static             & 1.00  & 98.90 & 13 & Day     & Low  & \cmark\\
        Fishyscapes-static'21~\cite{blum21ijcv}    & Hybrid    & 1000 & Diverse & Static             & 2.10  & 85.80 & 2  & Diverse & Low  & \xmark \\
        \cmidrule{1-11}

        LostAndFound'16~\cite{pinggera16iros}      & Real      & 1000 & Small   & Static             & 0.12  & 39.10 & 2  & Day     & None & \xmark \\
        RoadAnomaly'19~\cite{lis19iccv}            & Real      & 60   & Diverse & Static             & 9.85  & 33.16 & 2  & Day     & High & \xmark \\
        Fishyscapes-LaF'21~\cite{blum21ijcv}       & Real      & 275  & Small   & Static             & 0.23  & 81.13 & 2  & Day     & None & \xmark \\
        SOS'22~\cite{maag22accv}                     & Real      & 1129 & Diverse & Temporal           & 0.21  & 23.30 & 2  & Day     & None & \xmark \\
        WOS'22~\cite{maag22accv}                     & Real      & 938  & Diverse & Temporal           & 0.88  & 41.80 & 2  & Day     & None & \xmark \\
        SMIYC-RoadAnomaly'21~\cite{chan21neurips}  & Real      & 100  & Diverse & Static             & 13.80 & 82.20 & 2  & Day     & Low  & \xmark \\
        SMIYC-RoadObstacle'21~\cite{chan21neurips} & Real      & 327  & Small   & Temporal$^\dagger$ & 0.12  & 39.10 & 2  & Diverse & None & \xmark \\

        \cmidrule{1-11}
        \IDDDatasetStatic'24                       & Real      & 980  & Diverse & Static             & 2.18  & 89.60 & 20 & Diverse & High & \cmark \\
        \IDDDatasetTemporal'24                     & Real      & 1140 & Diverse & Temporal           & 1.20  & 85.60 & 20 & Diverse & High & \cmark \\
        
        \bottomrule
    \end{tabular}%
}
    \caption{\textbf{Comparison of existing datasets for anomaly detection} in driving scenarios. Datasets are compared in terms of dataset properties (\textit{Domain, Size, Modality}, number of \textit{Classes}), anomaly statistics (\textit{Anom}aly size, \%\textit{Anom}aly and \textit{Non-Anom}aly pixels), diversity of conditions (\textit{Weather/Environment}, \textit{Clutter}) and support for open-set evaluation (\textit{oIoU}). $^\dagger$ indicates low frame-per-second in sequences. The \emph{void} class is not considered in the class count reported in the table.}
    \vspace{-2mm}
\label{tab:overview}
\end{table*}%

\noindent The contributions are as follows.
\begin{enumerate}
    \item We introduce the first real-world segmentation dataset with both
        closed-set and anomaly labels with defined static and temporal test
        splits. %
    \item We present a comprehensive evaluation of the best performing state-of-the-art anomaly
        segmentation models, based on the standard and most commonly used SMIYC 
        benchmark leaderboard\footnote{\url{https://segmentmeifyoucan.com/leaderboard}}, covering 
        approaches from pixel-based to mask-based methods.
    \item We provide in-depth analysis of how in-domain, cross-domain,
        cross-sensor, lighting variations, and anomaly size affect performance. 
        Our results indicate that current methods
        struggle under these challenging conditions, highlighting the
        need for further research.
    \item The proposed \IDDDatasetTemporal, which consists of short video clips,
        opens up new directions for future research - particularly in test-time
        adaptation of anomaly segmentation models in real-world. %
\end{enumerate}

\section{Related Work} \label{sec:related_work}
\boldparagraph{Semantic segmentation driving datasets}
aggregate images from the driver's front view and label
them into the $19$ most relevant classes to driving tasks (such as road, curb, pedestrian, \etc), as originally proposed in~\cite{cordts16cvpr}. Some datasets include additional class labels tailored to the specific locations where the data was collected, such as a ``tricycle'' class in the China region~\cite{Huang2020}. However, all of them adhere to the basic $19$ classes for compatibility reasons. More recent datasets focus on increasing task difficulty by capturing scenes on a larger scale~\cite{neuhold17iccv,Huang2020}, incorporating unstructured traffic environments~\cite{varma19wacv}, or including adverse driving conditions~\cite{sakaridis21iccv,yu20cvpr,dai20ijcv,Zendel2022}. 

Despite these advancements, all datasets ignore pixels outside of the predefined training classes, and their evaluation protocols assess only closed-set performance, \ie,
performance on the classes specified during training. This lack of annotation for unknown objects in test sets and the closed-set
evaluation methodology limit their ability to validate models in realistic scenarios involving unknown objects. In contrast, \IDDDataset allows the evaluation of semantic perception models in the presence of unknowns by providing labels that include an unknown class, thus supporting an open-set evaluation. The statistics of the commonly used driving semantic segmentation datasets are shown in \cref{tab:overview_seq}.

\boldparagraph{Anomalies in road-driving scenes.}
Limited evaluation of standard semantic segmentation road-driving datasets gave rise to specialized datasets that benchmark the detection of unknowns as a standalone task \cite{blum21ijcv,chan21neurips}. The Fishyscapes \cite{blum21ijcv} benchmark evaluates obstacle detection in a subset of the LostAndFound~\cite{pinggera16iros} dataset and a subset of Cityscapes {\it val} injected with synthetic anomalies. The SMIYC \cite{chan21neurips} benchmark is fully based on real-world images and validates the detection of anomalies on drivable surfaces as well as on the whole images. Several other standalone test datasets were proposed in conjunction with novel methods, such as RoadAnomaly~\cite{lis19iccv} which was later merged into the SMIYC benchmark or synthetic Street-hazards~\cite{hendrycks22icml} which is not widely used due to large domain shift between not photorealistic synthetic and
real-world images. Most recently, WOS and SOS~\cite{maag22accv} datasets were introduced. These datasets include video sequences but only focus on the evaluation on drivable regions of interest. Unlike all
these datasets, \IDDDataset contains labels with 19 known classes and an unknown class which enables evaluation of anomaly detection performance in various regions of interest (such as the whole image, drivable surface only or anything in-between), and joint evaluation of the performance in open-set setting ($K+1$ class evaluation).

Acquiring detailed annotations (\eg, 19 known classes and an anomaly class) requires extensive manual effort. Thus, several datasets \cite{hendrycks22icml,bogdoll24arxiv} attempt to simplify the labeling efforts by simulating real-world traffic in synthetic environments. However, the 
quality of synthetic images diverges from the real-world data, leading to domain shifts that complicate the evaluation. Existing road driving anomaly datasets are summarized in
\cref{tab:overview}. 

\vspace{-2mm}
\section{The Proposed \IDDDataset Dataset}  \label{sec:dataset}

\begin{figure*}[t!]
\centering
  \setlength{\tabcolsep}{0.2em}
    \renewcommand{\arraystretch}{0.35}
    \hfill{}\hspace*{-0.5em}
    \footnotesize
    \begin{tabular}{@{}ccc@{}}
        \includegraphics[width=0.33\textwidth]{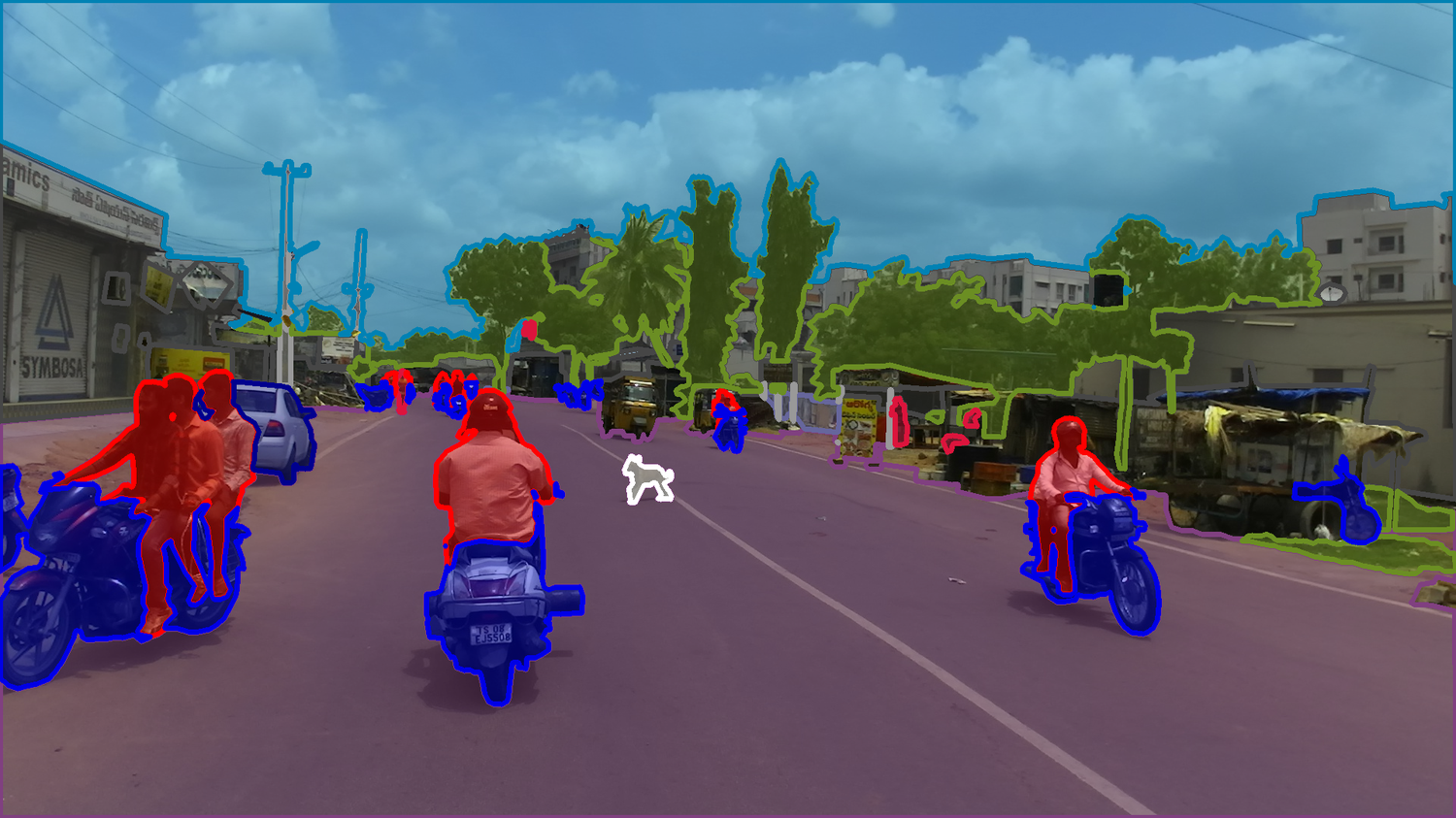}&
        \includegraphics[width=0.33\textwidth]{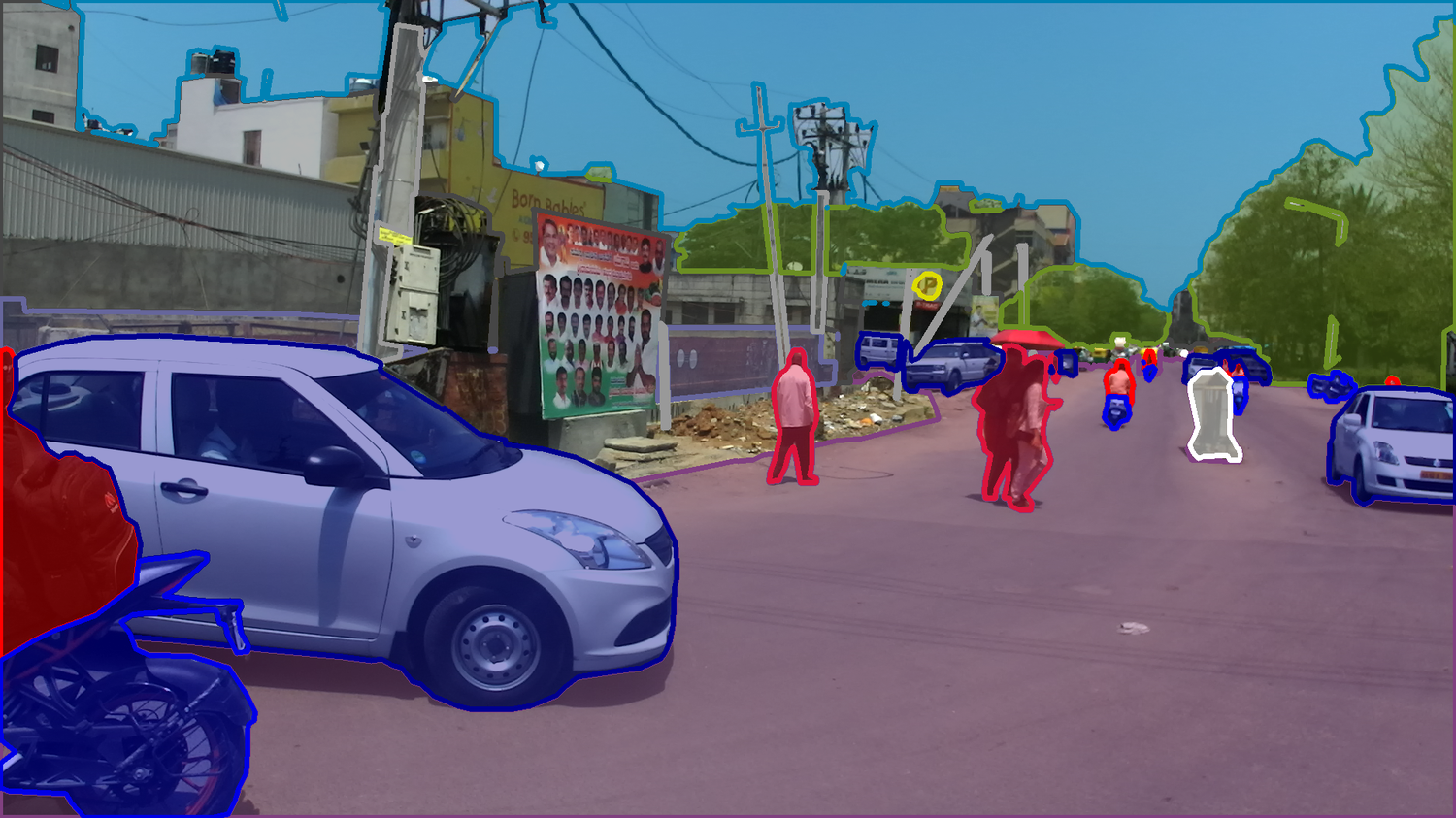}&
        \includegraphics[width=0.33\textwidth,height=0.144\textheight]{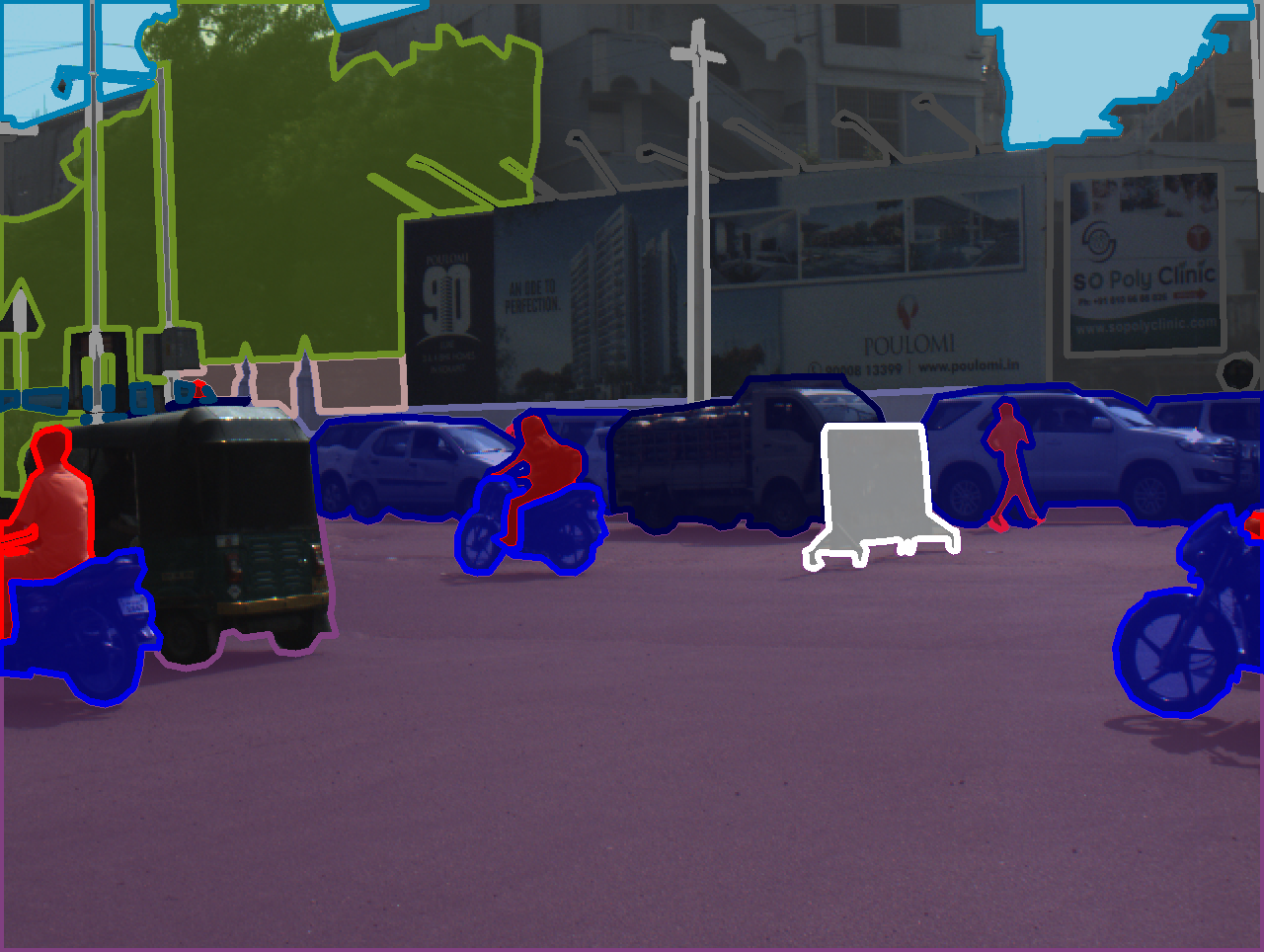}\\
         &  & \\
        \includegraphics[width=0.33\textwidth]{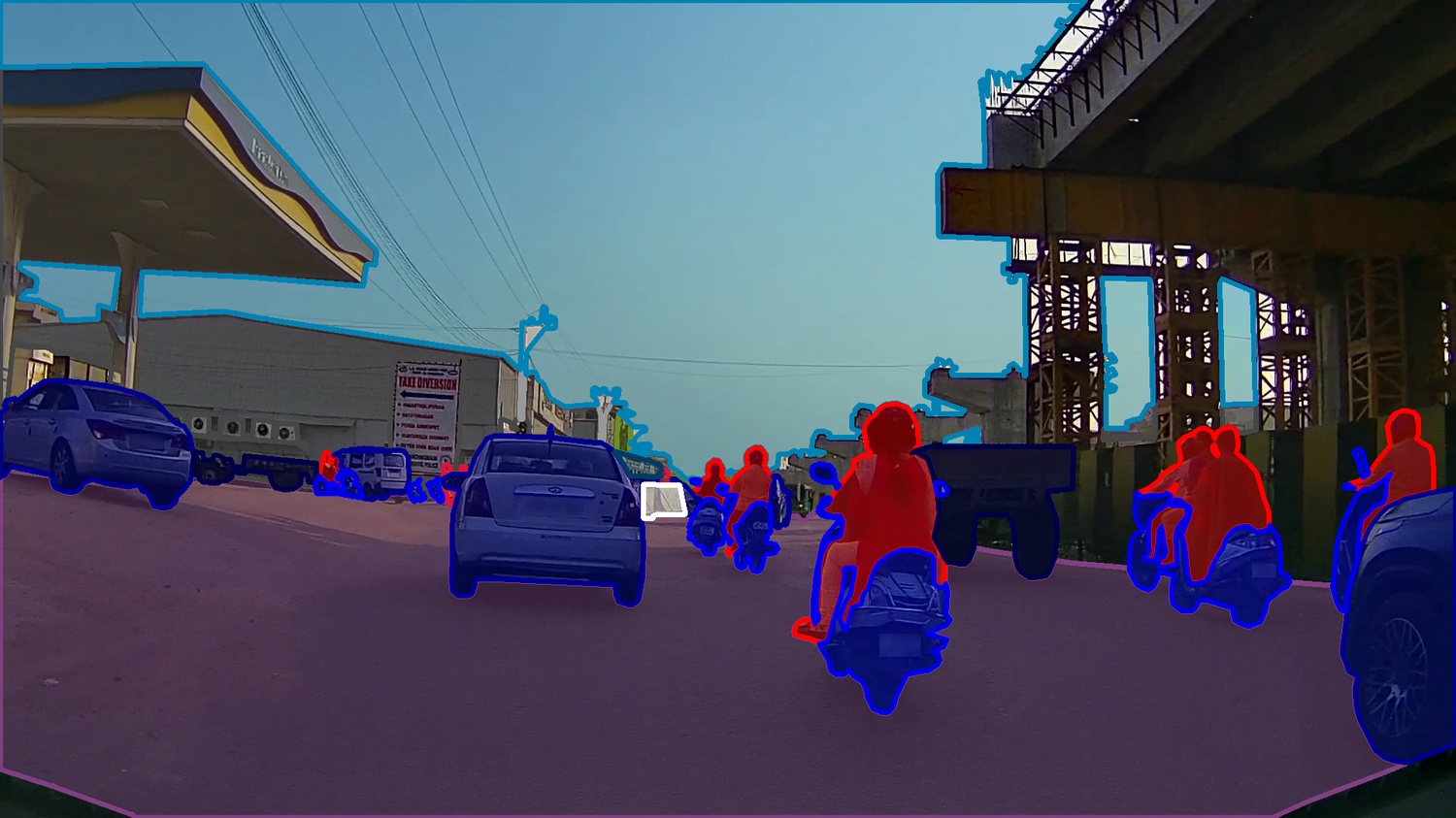}&
        \includegraphics[width=0.33\textwidth]{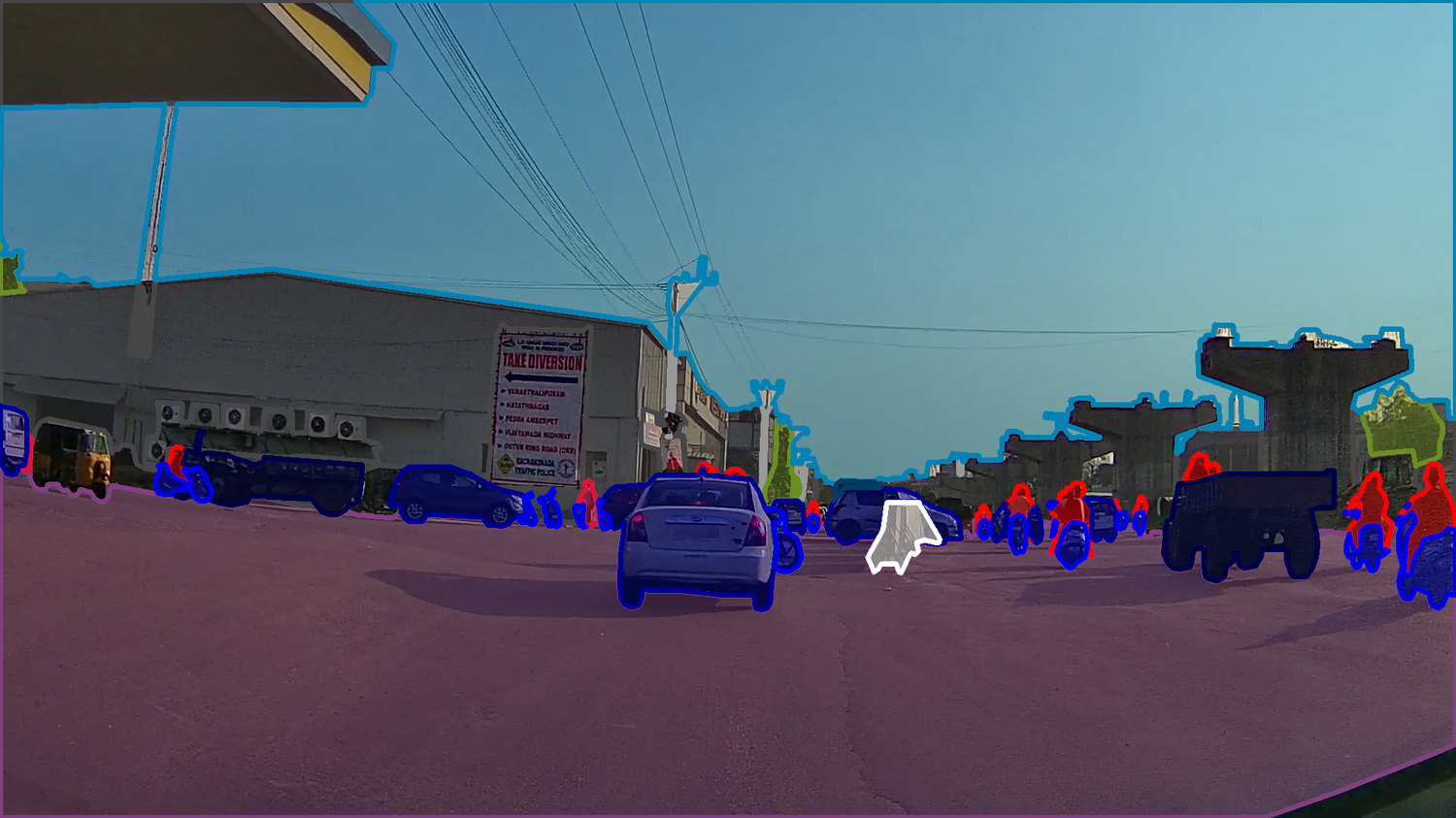}&
        \includegraphics[width=0.33\textwidth]{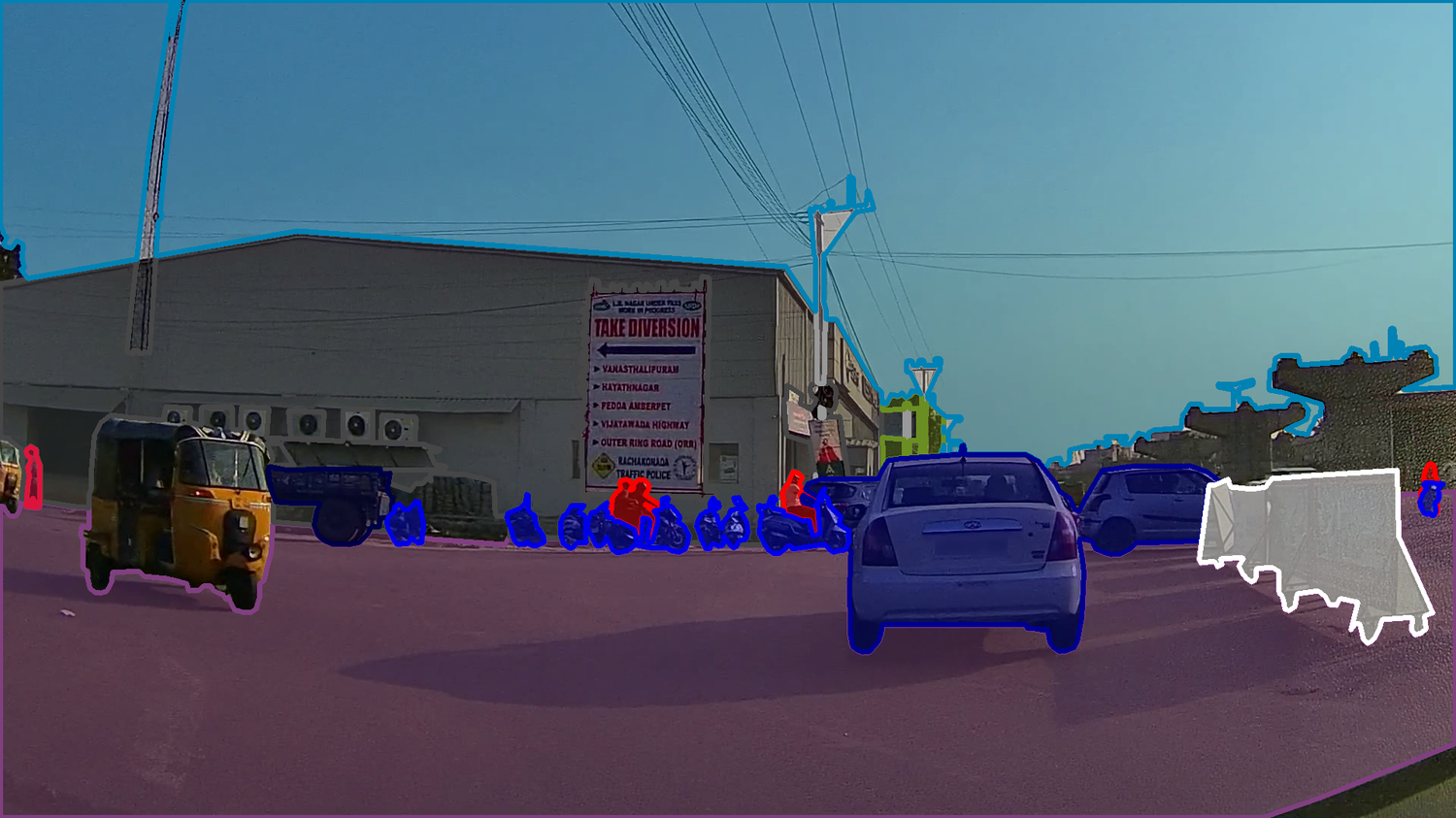}\\
    \end{tabular}
\caption{\textbf{Examples of anomalies (shown in white) in the \IDDDataset dataset.} The anomalous examples are ordered from small (left) to very large (right). Top: examples of anomalies of different size and shape at approximately the same distance from the ego-vehicle in \IDDDatasetStatic. Bottom: temporal view of an anomaly observed at different time-steps in \IDDDatasetTemporal.}
\vspace{-2mm}
\label{fig:examples}
\end{figure*}

Unstructured driving environments, such as those seen on Indian roads, are challenging for the task of semantic segmentation. The density of on-road and near-road traffic agents such as cars, pedestrians, road-side shops create a cluttered environment as shown in~\cref{fig:examples}. 

\subsection{Dataset Composition}
\label{sec:dataset_comp}
 We compose our dataset using images collected on Indian roads~\cite{varma19wacv,shaik24wacv,parikh24icra} with new and detailed annotations of known class and anomaly labels (\cf~\cref{sec:dataset_anno}). 
 The dataset consists of three parts, a training set (\IDDDatasetTrain), and two test sets (\IDDDatasetStatic and \IDDDatasetTemporal).
 
\noindent{\bf Training set (\IDDDatasetTrain).} It consists of images collected from different parts of Indian cities~\cite{varma19wacv,shaik24wacv}. 
The training set images contain {\it only} objects from the known classes.

\noindent{\bf Static test set (\IDDDatasetStatic).} It consists of images collected in the same way as the training set, but the test set images contains both known and anomalous objects. 

\noindent{\bf Temporal test set (\IDDDatasetTemporal).} It consist of short video clips that are also collected on Indian roads~\cite{parikh24icra}. Images in each clip contains both known and anomalous objects. This set is collected using a consumer grade dashcam~\cite{parikh24icra} which are ubiquitous but may produce lower image quality, \eg, due to firmware issues\footnote{https://dashcamtalk.com/forum/threads/ddpai-mini3-video-quality.37223/}. On the other hand, training sets \IDDDatasetTrain and CityScapes consist of images captured using higher quality cameras. 

\noindent{\bf Challenging examples}. We focus on studying the affect of challenging viewing conditions, such as extreme lighting conditions and weather variations. To achieve this, all three aforementioned dataset parts include several images and video clips collected in lowlight or in rainy conditions. Detailed examples are shown in the supplementary. This subset has many challenges, such as light burst from oncoming cars and low visibility. The images collected in rainy conditions contain rain droplets and wiper movements, making the segmentation task even more challenging.

\subsection{Training and Evaluation Setups}

\boldparagraph{In-domain Static Evaluation.}
Training on \IDDDatasetTrain and testing on \IDDDatasetStatic account for an \emph{in-domain Static evaluation setup}. 

\boldparagraph{Cross-domain Static Evaluation.}
Training models on the CityScapes dataset, and testing on \IDDDatasetStatic forms a \emph{cross-domain Static evaluation setup}.
The comparison with the in-domain setup allows us to evaluate the impact of such a domain shift in anomaly segmentation performance. 

\boldparagraph{Cross-sensor Temporal Evaluation.}
Testing on \IDDDatasetTemporal allows \textit{cross-sensor temporal evaluation} of methods trained on \IDDDatasetTrain or CityScapes due to the quality discrepancy between such sensors. Modeling domain shifts from such image corruptions isw an active field of research~\cite{hendrycks2019robustness}. We argue that it is necessary to benchmark anomaly segmentation methods in such real-world settings. 
This setup can be evaluated in an in-domain or cross-domain fashion, \ie training models on the CityScapes / \IDDDatasetTrain datasets, and testing on \IDDDatasetTemporal forms a \emph{cross / in-domain Temporal evaluation setup}.

\begin{figure*}[t!]
\centering
\vspace{-2mm}
  \includegraphics[width=0.99\textwidth]{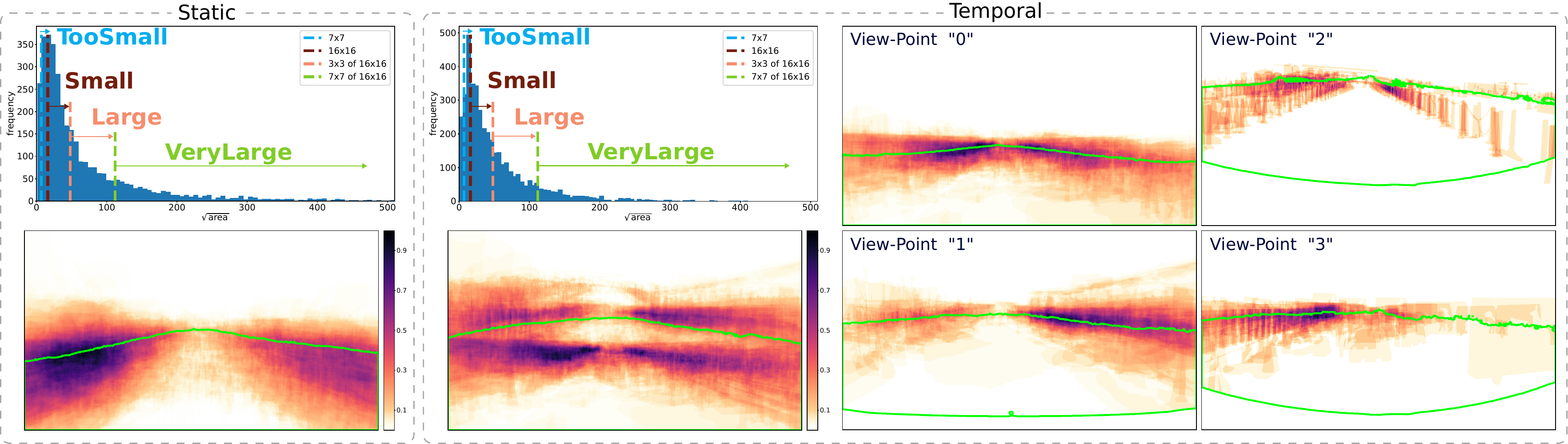}
    \caption{{\bf Distributions of anomalies with respect to their size and spatial location within images.} The anomalies are quantized to four different size
    intervals that are used in the ablation. Anomalies less than $7\times7$
    (black dashed line) are ignored during all evaluations. 
    The spatial distributions are visualized as a probability heatmap for each
    image location. Green line outlines road pixels that appeared in more than
    $50\%$ of dataset images. For temporal dataset the spatial distribution is also visualized for different view-points.}
    \vspace{-2mm}
    \label{fig:spatial}
\end{figure*}

\subsection{Annotation}\label{sec:dataset_anno}
The annotation process is performed to assign three types of labels: i) semantic class labels representing the known set of classes in the training set, ii) anomaly label denoting unknowns, and iii) void label for pixels that should not be taken into account during evaluation. 

\noindent\textbf{Labels of known classes.}
We follow the 19 CityScapes labels to define our known classes, but make adjustments for the context of Indian roads. As Indian roads often have blurry road boundaries, we assign both the road and the nearby drivable region as road class label. Unlike semantic segmentation datasets~\cite{varma19wacv}, we exclude traffic cones and short on-road traffic-poles from the known "traffic-sign" class following standard anomaly segmentation datasets~\cite{chan21neurips}.

\boldparagraph{Anomaly label.}
We conducted a rigorous multi-step annotation process (details are provided supplementary~\cref{sec:supp_dataset_curation}), identifying anomalies as objects outside the known 19 CityScapes classes and within a pre-defined region of interest (ROI) - on or within 2 meters of the road (based on visual inspection). Although this process may not cover all unknowns, it was a deliberate design choice to avoid unknowns outside the region of interest as they are less likely to affect the ego-vehicle. In addition, unknowns within the ROI, but frequently observed (\eg auto-rickshaws, banner) were also not included as anomalies. All such unknowns, but non-anomaly objects, were labeled as void in both train and test sets. Examples of anomaly objects in our datasets include tires, bins, water-tanks, construction material, road barricades, road-maintenance dugouts, animals, road-side vendor items such as fruits \etc, traffic cones and traffic-poles, pile of stones, mud and sand, tubs, rope, deep potholes. The process involved 7 annotators over a span of 2 months. To ensure that the team of annotators is familiar with the task, they received appropriate training until they achieved 95\% accuracy with respect to the CityScapes labels. Examples of anomalies are shown in~\cref{fig:examples}

\noindent\textbf{Void label.} Pixels that are not assigned to the labels of known classes or the anomaly label, according to the aforementioned guidelines, are assigned to the void label. 

\begin{table*}[t!]

\centering\resizebox{0.98\textwidth}{!}{
\begin{tabular}{ccccccccc|cccccc}
\toprule

{} & \multirow{3}{*}{Method} & \multirow{3}{*}{\shortstack[c]{OOD\\ Data}} & \multicolumn{6}{c}{Static} & \multicolumn{6}{c}{Temporal}\\

\cmidrule(lr){4-9}
\cmidrule(lr){10-15}

{} & {} & {} & \multicolumn{3}{c}{Road Anomaly} & \multicolumn{3}{c}{Closed \& Open-set} & 
               \multicolumn{3}{c}{Road Anomaly} & \multicolumn{3}{c}{Closed \& Open-set} \\

\cmidrule(lr){4-6}
\cmidrule(lr){7-9}
\cmidrule(lr){10-12}
\cmidrule(lr){13-15}

{} & {} & {} & 
\multicolumn{1}{c}{AP $\uparrow$} & \multicolumn{1}{c}{FPR$_T\downarrow$} & \multicolumn{1}{c}{TPR$_F\uparrow$} & \multicolumn{1}{c}{\text{IoU} $\uparrow$} & 
\multicolumn{1}{c}{oIoU$_{T}$ $\uparrow$} & \multicolumn{1}{c}{oIoU$_{F}$ $\uparrow$} & \multicolumn{1}{c}{AP $\uparrow$} & \multicolumn{1}{c}{FPR$_T\downarrow$} & \multicolumn{1}{c}{TPR$_F\uparrow$} & \multicolumn{1}{c}{$\text{IoU} \uparrow$} & 
\multicolumn{1}{c}{oIoU$_{T}$ $\uparrow$} & \multicolumn{1}{c}{oIoU$_{F}$ $\uparrow$} \\

\midrule
\multicolumn{15}{c}{In-domain} \\
\midrule
   \multirow{3}{*}{\rotatebox[origin=c]{90}{\it \scriptsize pixel-level}} 
                         & JSR-Net\textdagger & \xmark & \ph4.2 & 56.1   & \ph3.8 & 56.8 & \ph9.8 & 44.1 & \ph2.3 & 58.7 & \ph2.5 & 37.3 & \ph6.3 & 29.2   \\
    {}                   & DaCUP\textdagger   & \xmark & \ph5.4 & 100    & 20.4   & 57.0 & \ph9.0 & 47.0 & \ph2.9 & 100  & 11.1   & 37.3 & \ph6.9 & 30.4  \\
    {}                   & PixOOD             & \xmark & 20.3   & 39.4   & 50.9   & 65.8 & 47.8   & 60.9 & \ph6.2 & 56.5 & 26.2   & 55.1 & 32.9   & 52.2   \\
\cmidrule(lr){1-9}
\cmidrule(lr){10-15}
    \multirow{8}{*}{\rotatebox[origin=c]{90}{\it \scriptsize mask-level}}

    {}                   & RbA                & \xmark & 75.7   & 73.4   & 93.6   & 73.1 & 36.4   & 66.4 & 36.5   & 94.9 & 75.2   & 57.8 & \ph5.5 & 54.3  \\
    {}                   & EAM                & \xmark & 77.1   & \ph5.9 & 94.4   & 73.4 & 66.5   & 67.4 & 45.2   & 92.9 & 81.8   & 59.1 & \ph6.1 & 55.5  \\
    {}                   & Pebal              & \xmark & 69.9   & \ph9.2 & 93.3   & 73.1 & 64.2   & 67.2 & 32.4   & 92.6 & 74.9   & 57.8 & \ph7.8 & 55.4  \\
\cmidrule(lr){2-9}
\cmidrule(lr){10-15}

    {}                   & RbA                & \cmark & 79.1   & \ph3.9 & 95.9   & 72.9 & 67.8   & 66.5 & 37.7   & 29.4 & 76.6   & 57.8 & 41.7   & 55.6   \\
    {}                   & EAM                & \cmark & 76.8   & \ph4.2 & 96.1   & 73.8 & 68.4   & 67.6 & 38.7   & 91.4 & 84.4   & 59.5 & \ph6.7 & 55.7   \\
    {}                   & Pebal              & \cmark & 64.5   & \ph4.4 & 95.7   & 72.9 & 67.8   & 67.1 & 23.6   & 24.7 & 77.1   & 57.8 & 46.2   & 55.7   \\
    {}                   & UNO                & \cmark & 71.4   & \ph3.0 & 96.9   & 73.7 & 68.4   & 65.9 & 30.4   & 89.7 & 84.8   & 59.0 & \ph9.8 & 55.2  \\
    {}                   & M2A                & \cmark & 32.0   & 66.9   & 71.1   & 53.9 & 31.5   & 49.5 & 10.7   & 78.6 & 47.5   & 40.3 & 16.9   & 34.1   \\
\midrule
    
    \multicolumn{15}{c}{Cross-domain} \\
\midrule

    {}                   & PixOOD             & \xmark & 11.4 &       73.7 &      33.2 &         56.3 &             20.4 &             52.8 & 4.8 &       80.7 &      25.5 &         48.7 &             14.7 &             46.9  \\
    {}                   & RbA                & \xmark & 43.3 &       97.3 &      70.5 &         57.2 &              4.1 &             55.2 & 15.7 &       98.5 &      46.2 &         41.3 &              1.1 &             40.6  \\
    {}                   & RbA                & \cmark & 56.4 &       80.7 &      78.9 &         57.5 &             11.9 &             55.1 & 24.6 &       91.6 &      54.4 &         43.7 &              3.2 &             41.9  \\
    {}                   & UNO                & \cmark & 55.5 &       92.9 &      79.1 &         68.1 &             12.0 &             65.6 & 37.2 &       92.4 &      70.3 &         57.4 &              6.6 &             54.6 \\

\bottomrule
\end{tabular}
}
\caption{Results for road anomaly, closed-set and open-set evaluation protocols under in-domain and cross-domain evaluation setups. The {\it T} ({\it F}) subscript for oIoU metric refers to operating point (anomaly score threshold) for which the methods achieves $95\%$ TPR ($5\%$ FPR).}

\label{tab:main_anomaly}
\vspace{-5mm}
\end{table*}%

\subsection{Evaluation protocols}
We establish four distinct evaluation protocols, each focusing on different aspects of the problem. 
This is enabled by the proposed \IDDDataset dataset, because it includes images annotated with $K$ known classes along with an anomaly class.

\noindent{\bf Road obstacles} evaluation protocol considers the driving surfaces to be the area of interest for evaluation.
Therefore, pixels that are not annotated as \textit{road} or as anomalies are assigned to the void label during this evaluation setup.

\noindent{\bf Road anomaly} evaluation protocol benchmarks performance across all non-void pixels, \ie, pixels labeled as any of $K+1$ classes.
In contrast to the previous protocol, this one also accounts for errors occurring outside the driving regions, which increases the task difficulty.

\noindent{\bf Open-set} evaluation protocol validates the recognition of known classes in the presence of anomalies.
This protocol penalizes \textit{both} misclassifications among known classes and incorrect detection of anomalies.

\noindent{\bf Closed-set} evaluation protocol assesses classification performance solely on the $K$ known classes and maps the anomaly label to the void label during this evaluation protocol.
This standard evaluation protocol estimates the capabilities of trained models in an ideal setting and can serve as an upper bound for open-set evaluation performance.

\subsection{Metrics}
\noindent{\bf Average precision (AP)} quantifies anomaly detection performance by measuring the area under the precision-recall curve.
This threshold-free metric is used to evaluate performance in road obstacle and road anomaly protocols.

\noindent{\bf FPR$_T$} measures the false positive rate for the threshold that yields the true positive rate of $95\%$.
This is particularly important in safety-critical applications that demand high sensitivity and recall  of all anomalous objects. 

\noindent{\bf TPR$_F$} measures
the true positive rate for the threshold that yields the false positive rate of
$5\%$. This metric forms a complementary operation point to the previous one and
targets applications that require high precision, \ie low false positive detection.

\noindent{\bf F1 score} \cite{chan21neurips} combines the anomaly detection
metrics (recall and precision) and extends them from the pixel level to the
component level. By grouping connected neighboring anomalous pixels into
cohesive components, this approach provides instance-level performance
estimates.

\noindent{\bf Intersection-over-Union} (IoU) quantifies recognition performance
by measuring the overlap between predicted and ground-truth segments. Since
traffic scene segmentation is a multiclass classification task, we report the
macro-averaged IoU over the classes of interest. We use IoU to assess
the performance in closed-set evaluation.

\noindent{\bf Open-Intersection-over-Union} (oIoU) \cite{grcic24tpami}
evaluates recognition performance of known classes in the presence of anomalous
instances. Unlike the standard IoU metric, oIoU incorporates false positives
and false negatives committed by the anomaly detector. \textit{The difference between
IoU and oIoU highlights the performance gap between closed-set and open-set
deployments.}

\subsection{Statistics}
\label{sec:dataset_stats}
The train set of \IDDDatasetTrain comprises 3436 images, while the validation set comprises 762 images. The test set \IDDDatasetStatic contains 980 annotated images. \IDDDatasetTemporal, which consists of video clips, includes a total of 21118 images, of which 1140 are annotated. The unannotated images are released to facilitate future research in using temporal images for online test-time adaptation of anomaly segmentation models to mitigate the challenges of domain shifts. The number of pixels (log-scale) per class in \IDDDatasetTrain, \IDDDatasetStatic, \IDDDatasetTemporal is shown in Fig.~\ref{fig:stats} of the supplementary. As can be seen, the distribution of pixel counts per class is similar between train and test splits. Additionally, \cref{tab:supp_light_stats} in supplementary provides statistics on the number of images captured under normal daylight and adverse lowlight conditions across different \IDDDataset splits. The frequency histogram of anomaly sizes is shown in~\cref{fig:spatial}, illustrating significant variations in anomaly sizes. Examples of images showing these variations are presented in~\cref{fig:examples}. The spatial distribution of the anomalies, along with the approximate road regions, is shown in the bottom left of~\cref{fig:spatial} for both \IDDDatasetStatic and \IDDDatasetTemporal. The plot shows that the anomalies are distributed across various regions of the road.

\begin{table}[t!]
\vspace{4mm}
\centering\resizebox{0.45\textwidth}{!}{
\begin{tabular}{ccccc|cc}
\toprule

{} & \multirow{2}{*}{Method} & \multirow{2}{*}{\shortstack[c]{OOD\\ Data}} & \multicolumn{2}{c}{Static} & \multicolumn{2}{c}{Temporal}\\

\cmidrule(lr){4-5}
\cmidrule(lr){6-7}

{} & {} & {} & 
\multicolumn{1}{c}{AP $\uparrow$} & \multicolumn{1}{c}{FPR$_T\downarrow$} &  \multicolumn{1}{c}{AP $\uparrow$} & \multicolumn{1}{c}{FPR$_T\downarrow$}  \\ 

\midrule
\multicolumn{7}{c}{In-domain} \\
\midrule
   \multirow{3}{*}{\rotatebox[origin=c]{90}{\it \scriptsize pixel-level}} 
                        & JSR-Net\textdagger & \xmark & 85.7 & \ph8.4          & 52.1 & 26.5   \\
    {}                  & DaCUP\textdagger   & \xmark & 85.5 & 100             & 56.8 & 100  \\
    {}                  & PixOOD             & \xmark & 93.1 & \ph4.3          & 83.1 & 10.1 \\

\cmidrule(lr){1-5}
\cmidrule(lr){6-7}

    \multirow{8}{*}{\rotatebox[origin=c]{90}{\it \scriptsize mask-level}}

    {}                  & RbA                & \xmark & 92.7 & 77.5            & 53.4 & 98.9  \\
    {}                  & EAM                & \xmark & 94.5 & \ph2.2          & 70.0 & 98.1  \\
    {}                  & Pebal              & \xmark & 92.3 & \ph3.4          & 54.2 & 95.6  \\
\cmidrule(lr){2-5}
\cmidrule(lr){6-7}

    {}                  & RbA                & \cmark & 95.8 & \ph1.7          & 57.2 & 33.7  \\
    {}                  & EAM                & \cmark & 95.6 & \ph1.6          & 62.1 & 96.2 \\
    {}                  & Pebal              & \cmark & 92.5 & \ph1.9          & 48.9 & 23.8   \\
    {}                  & UNO                & \cmark & 94.0 & \ph1.2          & 56.1 & 92.3  \\
    {}                  & M2A                & \cmark & 48.9 & 78.5            & 30.0 & 79.5  \\

\midrule
\multicolumn{7}{c}{Cross-domain} \\
\midrule

    {}                   & PixOOD             & \xmark & 92.3 & 5.1  & 84.3 & 10.8 \\
    {}                   & RbA                & \xmark & 62.4 &       99.1     & 32.5 &       99.3 \\
    {}                   & RbA                & \cmark & 76.1 &       68.9    & 
    37.9 &       87.9\\
    {}                  & UNO                & \cmark & 66.3 &       90.8   & 
    49.1 &       90.5  \\

\bottomrule
\end{tabular}
}
\caption{Results for road obstacle evaluation protocols under in-domain and cross-domain setups.}

\vspace{-3mm}
\label{tab:main_obstacle}
\end{table}%

\begin{figure*}[ht!]
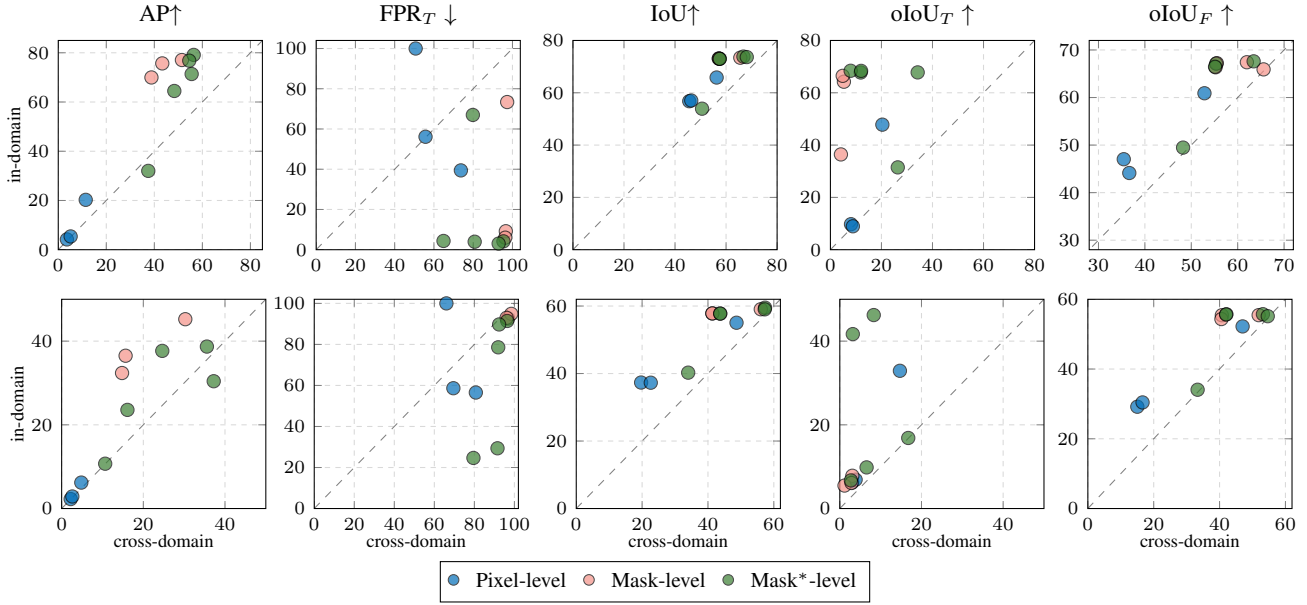

\centering
    \setlength{\tabcolsep}{-0.1em}
    \renewcommand{\arraystretch}{0.5}
    \begin{tabular}{@{}ccccc@{}}
        \input{fig/static-cross-in} \\
        \input{fig/temporal-cross-in} \\
        \multicolumn{5}{c}{ 
            \begin{tikzpicture}
            \begin{customlegend}[legend style={align=left,column sep=1ex, font=\footnotesize, fill opacity=1.0},
                legend entries={Pixel-level, Mask-level, $\text{Mask}^{*}$-level},
                legend columns=-1,
                legend image post style={scale=0.8,yshift=0.2ex},
                ]
                    \addlegendimage{only marks, mark options={draw=black}, color=NavyBlue, solid, mark=, opacity=0.7, mark size=2.5},
                    \addlegendimage{only marks, mark options={draw=black}, color=Salmon, solid, mark=, opacity=0.7, mark size=2.5}   
                    \addlegendimage{only marks, mark options={draw=black}, color=OliveGreen, solid, mark=, opacity=0.7, mark size=2.5}   
                    \end{customlegend}
            \end{tikzpicture}
        }\\
    \end{tabular}
    \caption{\textbf{Cross-domain \vs In-domain performance in road anomaly evaluation protocol}. Top row -- Static, bottom row -- Temporal. Mask$^*$ are mask-based methods trained with OOD data. The $y = x$ reference line shows relative gain or drop. The {\it T} ({\it F}) subscript for oIoU metric refers to operating point (anomaly score threshold) for which the methods achieves $95\%$ TPR ($5\%$ FPR).}
    \vspace{-3mm}
    \label{fig:exp_cross}
\end{figure*}

\section{Baselines}\label{sec:baselines}
The baselines were selected as the top performing methods on the standard and the most 
frequently used SMIYC~\cite{chan21neurips} benchmark. We broadly categorized them into two 
groups based on the granularity of regions for which an anomaly score is 
predicted, \ie, pixel-level and mask-level. The methods are briefly described 
in the following paragraphs. 

\noindent{\bf Pixel-level baselines.}
We consider two reconstruction-based methods, JSR-Net~\cite{vojir2021} and DaCUP~\cite{vojir2023} that localizes anomalies as poorly reconstructed pixels. We also include recent PixOOD~\cite{vojir2024} that uses a statistical decision strategy in pre-trained representation to detect anomalies.

\noindent{\bf Mask-level baselines.}
We consider baselines that extend the mask-level classifier \cite{cheng22cvpr}.
A seminal mask-level approach EAM \cite{grcic23cvprw} assigns anomaly scores to
masks instead of pixels and aggregates decisions to recover dense predictions.
RbA \cite{nayal23iccv} considers regions rejected by all masks as anomalous,
while Mask2Anomaly (M2A) \cite{rai23iccv} adapts the model architecture to 
enhance anomaly detection. Finally, UNO \cite{delic24bmvc} revisits the $K+1$
classifier built on top of the mask classifier and combines negative class
recognition with prediction uncertainty to improve anomaly detection. %

\section{Experimental results} \label{sec:experiments}
\begin{figure}[t]
\resizebox{0.47\textwidth}{!}{
\centering
    \setlength{\tabcolsep}{-0.1em}
    \renewcommand{\arraystretch}{0.5}
    \begin{tabular}{@{}cc@{}}

        \multicolumn{2}{c}{ 
            \begin{tikzpicture}
            \begin{customlegend}[legend style={align=left,column sep=1ex, font=\footnotesize, fill opacity=1.0},
                ybar,
                legend entries={t-Small, Small, Large, v-Large},
                legend columns=-1,
                ]
                    \addlegendimage{fill=Cyan, mark options={draw=black}, solid, mark=, opacity=1, mark size=2.5}   
                    \addlegendimage{fill=Brown, mark options={draw=black}, solid, mark=, opacity=1, mark size=2.5}   
                    \addlegendimage{fill=Melon, mark options={draw=black}, solid, mark=, opacity=1, mark size=2.5}   
                    \addlegendimage{fill=LimeGreen, mark options={draw=black}, solid, mark=, opacity=1, mark size=2.5}   
                    \end{customlegend}
            \end{tikzpicture}
        }\\

\begin{tikzpicture}
    \begin{axis} [
        ylabel shift = -4pt,
        ybar, %
        bar width=9pt,
        width=0.49\textwidth,
        height=4.8cm,
        enlarge x limits=0.3,
        ymin=0, ymax=87, %
        ylabel={F1},
        xlabel={},
        xtick=data,
        ymajorgrids=true,
        major grid style={dotted,black},
        tickpos=left,
        symbolic x coords={ PixOOD, EAM, RbA}, %
        xticklabels={PixOOD~\xmark, EAM~\cmark, RbA~\cmark},
        x tick label style={rotate=0, anchor=center, yshift=-2pt}, %
        xtick style={draw=none},
        legend style={at={(0.5,1.05)}, anchor=south, legend columns=-1, /tikz/every even column/.append style={column sep=0.25cm}},
        every node near coord/.append style={rotate=45,font=\normalsize, scale=0.6, yshift=4pt, xshift=8pt}, 
        nodes near coords, %
        nodes near coords align={vertical}
    ]
    \addplot[fill=Cyan] coordinates {
         (PixOOD, 22.4) (EAM, 17.2) (RbA, 11.9) 
    };
    \addplot[fill=Brown] coordinates {
         (PixOOD, 43.6) (EAM, 39.7) (RbA, 32.7) 
        
    };
    \addplot[fill=Melon] coordinates {
         (PixOOD, 51.2) (EAM, 73.7) (RbA, 71.4) 
        
    };
    \addplot[fill=LimeGreen] coordinates {
         (PixOOD, 20.3) (EAM, 68.2) (RbA, 69.9) 
        
    };
    
    \legend{}
    \end{axis}
\end{tikzpicture} & 
\phantom{F1}\begin{tikzpicture}
    \begin{axis} [
        ylabel shift = -4pt,    
        ybar, %
        bar width=9pt,
        width=0.49\textwidth,
        height=4.8cm,
        enlarge x limits=0.3,
        ymin=0, ymax=87, %
        xlabel={},
        xtick=data,
        ymajorgrids=true,
        major grid style={dotted,black},
        tickpos=left,
        symbolic x coords={ PixOOD, EAM, RbA}, %
        xticklabels={PixOOD~\xmark, EAM~\cmark, RbA~\cmark},
        x tick label style={rotate=0, anchor=center, yshift=-2pt}, %
        xtick style={draw=none},
        legend style={at={(0.5,1.05)}, anchor=south, legend columns=-1},
        every node near coord/.append style={rotate=45,font=\normalsize,scale=0.6, yshift=4pt, xshift=8pt}, 
        nodes near coords, %
        nodes near coords align={vertical}
    ]
    \addplot[fill=Cyan] coordinates {
         (PixOOD, 14.6) (EAM, 17.9) (RbA, 6.5) 
    };
   
    \addplot[fill=Brown] coordinates {
         (PixOOD, 44.3) (EAM, 29.1) (RbA, 20.5) 
    };
    \addplot[fill=Melon] coordinates {
         (PixOOD, 63.1) (EAM, 60.0) (RbA, 49.3) 
    };
    \addplot[fill=LimeGreen] coordinates {
         (PixOOD, 9.1) (EAM, 47.9) (RbA, 39.8) 
    };
   
   \legend{}; %
    \end{axis}
\end{tikzpicture} \\
\end{tabular}%
} %
\caption{{\bf Ablation of different anomaly sizes.} The plot shows results in \IDDDatasetStatic (left) and \IDDDatasetTemporal(right) for road anomaly evaluation protocol under in-domain setup.}%
\vspace{-3mm}
\label{fig:ablation_size_static}
\end{figure}
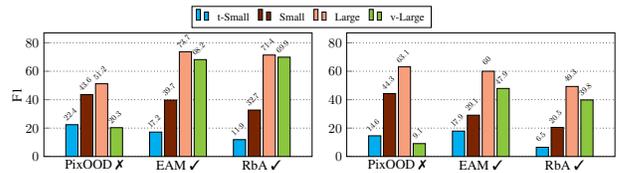 \begin{table*}[t!]

\centering
\resizebox{0.98\textwidth}{!}{
\begin{tabular}{clcrrrrrr|rrrrrr}
\toprule

{} & {} & {} & \multicolumn{6}{c}{Day} & \multicolumn{6}{c}{Lowlight}\\

\cmidrule(lr){4-9}
\cmidrule(lr){10-15}

{} & \multirow{2}{*}{Method} & \multirow{2}{*}{\shortstack[c]{OOD\\ Data}} & 
 \multicolumn{3}{c}{\IDDDatasetStatic}& \multicolumn{3}{c}{\IDDDatasetTemporal} &
 \multicolumn{3}{c}{\IDDDatasetStatic}& \multicolumn{3}{c}{\IDDDatasetTemporal} \\

\cmidrule(lr){4-6}
\cmidrule(lr){7-9}
\cmidrule(lr){10-12}
\cmidrule(lr){13-15}

{} & {} & {} & 
\multicolumn{1}{c}{AP $\uparrow$}& \multicolumn{1}{c}{FPR$_T$ $\downarrow$}&\multicolumn{1}{c}{$\overline{\text{F1}} \uparrow$}&
\multicolumn{1}{c}{AP $\uparrow$}& \multicolumn{1}{c}{FPR$_T$ $\downarrow$}&\multicolumn{1}{c}{$\overline{\text{F1}} \uparrow$}&
\multicolumn{1}{c}{AP $\uparrow$}& \multicolumn{1}{c}{FPR$_T$ $\downarrow$}&\multicolumn{1}{c}{$\overline{\text{F1}} \uparrow$}&
\multicolumn{1}{c}{AP $\uparrow$}& \multicolumn{1}{c}{FPR$_T$ $\downarrow$}&\multicolumn{1}{c}{$\overline{\text{F1}} \uparrow$}\\

\midrule
   \multirow{3}{*}{\rotatebox[origin=c]{90}{\it \scriptsize pixel-level}} 
        & JSR-Net   & \xmark & 4.26 & 57.45 & 1.84 & 2.28 & 75.69 & 0.83 & 4.20 &       39.94 & 0.61 & 2.28 & 75.69 & 0.83\\
    {}  & DaCUP     & \xmark & 5.09 & 100.00 & 1.67 & 2.90 &      100.00 &       2.25 & 7.80 &       38.80 &       3.13 &  2.80 &      100.00 &       2.78 \\
    {}  & PixOOD    & \xmark &  34.24 &       32.46 &       2.28 &  16.07 &       53.57 &       1.50 & 5.63 &       60.59 &       0.65 &  2.16 &       65.76 &       0.56 \\

\cmidrule(lr){1-9}
\cmidrule(lr){10-15}

    \multirow{8}{*}{\rotatebox[origin=c]{90}{\it \scriptsize mask-level}} 
    {}  & RbA          & \xmark & 77.06 &        5.59 &      16.34 & 39.82 &       95.77 &      11.38 & 67.99 &       97.44 &       9.28 & 24.16 &       96.20 &       6.81  \\
    {}  & EAM          & \xmark &  77.72 &        5.09 &      21.34 &  47.48 &       95.57 &      15.06 & 73.92 &       95.58 &      14.61 & 37.68 &       89.63 &      11.50 \\
    {}  & Pebal        & \xmark &  71.27 &        6.37 &      21.60 & 34.95 &       97.45 &      11.84 & 61.61 &       89.31 &       9.78 & 21.70 &       87.91 &       7.18 \\
    
\cmidrule(lr){2-9}
\cmidrule(lr){10-15}

    {}  & RbA          & \cmark & 79.64 &        3.44 &      22.14 & 40.75 &       23.03 &      12.00 & 75.40 &       74.38 &      12.30 & 28.56 &       74.54 &       8.64  \\
    {}  & EAM          & \cmark & 77.26 &        3.63 &      22.21 & 39.62 &       94.27 &      15.28 & 74.47 &       14.65 &      14.26 & 37.81 &       84.35 &      12.06 \\
    {}  & Pebal        & \cmark &  65.39 &        3.71 &       0.00 & 28.55 &       22.19 &       0.00 & 58.92 &       16.92 &       0.00 & 12.20 &       28.80 &       0.00 \\
    {}  & UNO          & \cmark &  71.71 &        2.82 &      29.19 & 31.78 &       41.77 &      18.60 & 74.47 &       14.65 &      14.26 & 30.10 &       88.17 &      14.58 \\
    {}  & Mask2Anomaly & \cmark &  35.13 &       61.02 &       9.45 & 11.71 &       77.05 &       5.61 & 19.97 &       91.35 &       6.17 & 7.78 &       86.29 &       4.32 \\

\bottomrule
\end{tabular}
}
\caption{\textbf{Ablation of different lighting conditions.} Results for in-domain road anomaly evaluation protocol under day and light-adverse conditions (night, rain, fog, dawn).}
\label{tab:ablation_day_night}
\end{table*}%
 \noindent
{\bf Road Anomaly Evaluation} results are presented in~\cref{tab:main_anomaly} and~\cref{tab:supp_main_anomaly} in supplementary. For the in-domain Static evaluation setup, most Mask2Former (mask-level) methods trained with auxiliary out-of-domain (OOD) data achieve good performance across three anomaly detection metrics: FPR$_T$ ($<$ 5\%), TPR$_F$ ($>$ 90\%) and AP ($>$ 70\%). Due to the lack of any ``objectness'' priors, the pixel-level methods
classify many random pixels as anomalous with high confidence, resulting in a poor anomaly detection metrics.  In contrast, results on the challenging in-domain Temporal setup show that both pixel-level and mask-level methods have high FPR$_T$ and low AP. This indicates that domain shifts due to differences in sensor quality adversely affect anomaly detection performance. 

Results in cross-domain Static and cross-domain Temporal setups shows a significant performance drop for all methods compared to the in-domain setup (\cf~\cref{fig:exp_cross}) resulting in low AP and high FPR$_T$. The complementary TPR$_F$ metric shows that some mask-based methods such as UNO could detect 79\% and 70\% of anomalies for cross-domain Static and Temporal setups respectively. Successfully detecting the remaining anomalies, to achieve 95\% TPR, results in high FPR$_T$ ( $>$ 90\%). This is because the methods include many known-class pixels as true-positives to correctly classify the hard anomalous cases. Qualitative examples of hard anomalies are presented in~\cref{sec:supp_qualitative_fpr} (~\cref{fig:supp_qualitative_rba,fig:qualit_rba_pxood}) and~\cref{sec:addn_qualt} (~\cref{fig:qualit_roc,fig:qualit_rba_op,fig:qualit_uno_op}) in supplementary.

\noindent
{\bf Closed and Open-set Evaluation (\cref{tab:main_anomaly}).}
The IoU metric for \IDDDataset is for all methods about $10\%$ lower than respective IoU achieved in CityScapes considering in-domain Static setup
($75.88\%$ vs. $65.83\%$ for PixOOD and $83.5-83.7\%$ vs. $\sim73\%$ for most
Mask2former methods). This difference is significantly higher (20-30\%) for in-domain Temporal and cross-domain setups. This highlights the difficulty of the cluttered
traffic environment in India and challenging domain shifts. Open-set IoU (oIoU) follows similar conclusions as road anomaly evaluation protocols. Results on in-domain Temporal and cross-domain setups show significant difference between closed set IoU (IoU) and open-set IoU at 95\% TPR (oIoU$_T$). This is attributed to misdetection of anomalies as known classes and known classes as anomalies. The drop is less significant between IoU and oIoU$_F$ but results in significantly lower TPR$_F$. It is to be noted, zero drop (IoU = oIoU) can be achieved by not detecting any anomalies (TPR/FPR = 0\%). Thus it is important to analyze both  TPR$_F$ (FPR$_T$) and oIoU$_F$ (oIoU$_T$). The open-set results signify the importance of evaluating semantic segmentation in real-world setting by jointly evaluating closed-set segmentation in the presence of anomalous object.

\noindent
{\bf Road Obstacle Evaluation (\cref{tab:main_obstacle}).} When the evaluation is limited to road regions, the pixel-level methods generally are much
better at generalizing from CityScapes to the \IDDDataset, resulting in a much
lower FPR metric for both static and temporal datasets. However, for in-domain Static evaluation setup the mask-level methods are able to outperform other
method, mainly due to strong priors baked in object-wise mask predictions that
seems to be more robust to detecting entire anomaly instances. In the 
temporal part where the different sensors act as a
form of a domain-shift the Mask2former based methods struggle to localize all
anomalies resulting in high FPR. 
The effects of domain shift are less pronounced in this setup due to the uniformity of roads, as shown in~\cref{tab:supp_main_obstacle}.

\noindent{\bf Ablations: anomaly sizes.}
The effect of anomaly sizes is shown in the~\cref{fig:ablation_size_static}, where specific anomaly size ranges
(defined in \cref{fig:spatial}) are considered. The size intervals were 
motivated by the dataset statistics and spatial resolution of the most
commonly used backbone architectures. We use the F1 metric, which is designed to
measure instance-level performance. The metric is generally improved with
larger anomaly sizes, except for the largest anomalies, where the methods
struggle to accurately and fully segment the very large instances. This is
again more apparent for pixel-level methods. Consistently with the evaluation
limited to road region, the temporal dataset with the additional challenges of
different sensors negatively effects Mask2former based methods
significantly more than the pixel-level methods. The results for all methods are in the supplementary~\cref{fig:supp-ablation_size_static}.

\noindent{\bf Ablations: lighting variations.}
This ablation compares the performance of the methods under lighting variations - day
(clear weather and good lighting conditions) and lowlight (\eg fog, rain,
dawn). The results are presented in~\cref{tab:ablation_day_night} which shows that both the pixel-based and mask-based methods struggle under lowlight conditions.

\section{Conclusions} \label{sec:conclusions}

We presented a new dataset and a benchmark for anomaly segmentation in a real-world setting. The results show that cross-domain generalization remains a challenge for current state-of-the-art anomaly segmentation methods. When trained on in-domain data, the performance of these models improves by a significant margin. This forms a strong baseline for future work in cross-domain generalization and adaptation of anomaly segmentation models. The results also showed that existing methods struggle in the presence of lower sensor quality, lower visibility, and small anomaly size. The diverse conditions provided by our benchmark offer a timely test-bed for anomaly segmentation research.

\section*{Acknowledgments}

\noindent The authors acknowledge support from respective sources as follows. Zakaria Laskar: Programme Johannes Amos
Comenius (no. CZ.02.01.01/00/22 010/0003405). Giorgos Tolias: Junior Star GACR (no. GM 21-28830M).  
Tomáš Vojíř and Jiri Matas: Toyota Motor Europe and by the Czech Science Foundation grant 25-15993S. Shankar Gangisetty and C.V. Jawahar: iHub-Data and Mobility at IIIT Hyderabad.
Matej Grcic: Croatian Recovery and Resilience Fund - NextGenerationEU (grant C1.4 R5-I2.01.0001). Iaroslav Melekhov and Juho Kannala: Research Council of Finland (grants 352788, 353138, 362407), Wallenberg AI, Autonomous Systems and Software Program (WASP) funded by the Knut and Allice Wallenberg Foundation.

We thank Ram Sharma from CVIT, IIIT Hyderabad, Mahender Reddy and the annotation team in Annotations and Data Capture-Mobility, IIIT Hyderabad for their work in the annotation process.

{
    \small
    \bibliographystyle{ieeenat_fullname}
    \bibliography{main}

\begin{thebibliography}{32}
\providecommand{\natexlab}[1]{#1}
\providecommand{\url}[1]{\texttt{#1}}
\expandafter\ifx\csname urlstyle\endcsname\relax
  \providecommand{\doi}[1]{doi: #1}\else
  \providecommand{\doi}{doi: \begingroup \urlstyle{rm}\Url}\fi

\bibitem[Blum et~al.(2021)Blum, Sarlin, Nieto, Siegwart, and Cadena]{blum21ijcv}
Hermann Blum, Paul{-}Edouard Sarlin, Juan~I. Nieto, Roland Siegwart, and Cesar Cadena.
\newblock The fishyscapes benchmark: Measuring blind spots in semantic segmentation.
\newblock \emph{IJCV}, 2021.

\bibitem[Bogdoll et~al.(2024)Bogdoll, Hamdard, R{\"{o}}{\ss}ler, Geisler, Bayram, Wang, Imhof, de~Campos, Tabarov, Yang, Gottschalk, and Z{\"{o}}llner]{bogdoll24arxiv}
Daniel Bogdoll, Iramm Hamdard, Lukas~Namgyu R{\"{o}}{\ss}ler, Felix Geisler, Muhammed Bayram, Felix Wang, Jan Imhof, Miguel de Campos, Anushervon Tabarov, Yitian Yang, Hanno Gottschalk, and J.~Marius Z{\"{o}}llner.
\newblock Anovox: {A} benchmark for multimodal anomaly detection in autonomous driving.
\newblock \emph{CoRR}, abs/2405.07865, 2024.

\bibitem[Chan et~al.(2021)Chan, Lis, Uhlemeyer, Blum, Honari, Siegwart, Fua, Salzmann, and Rottmann]{chan21neurips}
Robin Chan, Krzysztof Lis, Svenja Uhlemeyer, Hermann Blum, Sina Honari, Roland Siegwart, Pascal Fua, Mathieu Salzmann, and Matthias Rottmann.
\newblock Segmentmeifyoucan: {A} benchmark for anomaly segmentation.
\newblock In \emph{NeurIPS Dataset and Benchmarks}, 2021.

\bibitem[Chen et~al.(2018)Chen, Zhu, Papandreou, Schroff, and Adam]{chen18eccv}
Liang{-}Chieh Chen, Yukun Zhu, George Papandreou, Florian Schroff, and Hartwig Adam.
\newblock Encoder-decoder with atrous separable convolution for semantic image segmentation.
\newblock In \emph{ECCV}, 2018.

\bibitem[Cheng et~al.(2022)Cheng, Misra, Schwing, Kirillov, and Girdhar]{cheng22cvpr}
Bowen Cheng, Ishan Misra, Alexander~G. Schwing, Alexander Kirillov, and Rohit Girdhar.
\newblock Masked-attention mask transformer for universal image segmentation.
\newblock In \emph{CVPR}, 2022.

\bibitem[Cordts et~al.(2016)Cordts, Omran, Ramos, Rehfeld, Enzweiler, Benenson, Franke, Roth, and Schiele]{cordts16cvpr}
Marius Cordts, Mohamed Omran, Sebastian Ramos, Timo Rehfeld, Markus Enzweiler, Rodrigo Benenson, Uwe Franke, Stefan Roth, and Bernt Schiele.
\newblock The cityscapes dataset for semantic urban scene understanding.
\newblock In \emph{CVPR}, 2016.

\bibitem[Dai et~al.(2020)Dai, Sakaridis, Hecker, and Gool]{dai20ijcv}
Dengxin Dai, Christos Sakaridis, Simon Hecker, and Luc~Van Gool.
\newblock Curriculum model adaptation with synthetic and real data for semantic foggy scene understanding.
\newblock \emph{IJCV}, 2020.

\bibitem[Delić et~al.(2024)Delić, Grcic, and Šegvić]{delic24bmvc}
Anja Delić, Matej Grcic, and Siniša Šegvić.
\newblock Outlier detection by ensembling uncertainty with negative objectness.
\newblock In \emph{BMVC}, 2024.

\bibitem[Everingham et~al.(2010)Everingham, Gool, Williams, Winn, and Zisserman]{everingham10ijcv}
Mark Everingham, Luc~Van Gool, Christopher K.~I. Williams, John~M. Winn, and Andrew Zisserman.
\newblock The pascal visual object classes {(VOC)} challenge.
\newblock \emph{IJCV}, 2010.

\bibitem[Grcic and Segvic(2024)]{grcic24tpami}
Matej Grcic and Sinisa Segvic.
\newblock Hybrid open-set segmentation with synthetic negative data.
\newblock \emph{IEEE TPAMI}, 2024.

\bibitem[Grcic et~al.(2023)Grcic, Saric, and Segvic]{grcic23cvprw}
Matej Grcic, Josip Saric, and Sinisa Segvic.
\newblock On advantages of mask-level recognition for outlier-aware segmentation.
\newblock In \emph{CVPR}, 2023.

\bibitem[Hendrycks and Dietterich(2019)]{hendrycks2019robustness}
Dan Hendrycks and Thomas Dietterich.
\newblock Benchmarking neural network robustness to common corruptions and perturbations.
\newblock \emph{Proceedings of the International Conference on Learning Representations}, 2019.

\bibitem[Hendrycks and Gimpel(2017)]{hendrycks17iclr}
Dan Hendrycks and Kevin Gimpel.
\newblock A baseline for detecting misclassified and out-of-distribution examples in neural networks.
\newblock In \emph{ICLR}, 2017.

\bibitem[Hendrycks et~al.(2022)Hendrycks, Basart, Mazeika, Zou, Kwon, Mostajabi, Steinhardt, and Song]{hendrycks22icml}
Dan Hendrycks, Steven Basart, Mantas Mazeika, Andy Zou, Joseph Kwon, Mohammadreza Mostajabi, Jacob Steinhardt, and Dawn Song.
\newblock Scaling out-of-distribution detection for real-world settings.
\newblock In \emph{Int. Conf. on Mach. Learn.} {PMLR}, 2022.

\bibitem[Huang et~al.(2020)Huang, Wang, Cheng, Zhou, Geng, and Yang]{Huang2020}
Xinyu Huang, Peng Wang, Xinjing Cheng, Dingfu Zhou, Qichuan Geng, and Ruigang Yang.
\newblock {The ApolloScape Open Dataset for Autonomous Driving and Its Application}.
\newblock In \emph{IEEE TPAMI}, pages 2702--2719, 2020.

\bibitem[Kuznetsova et~al.(2020)Kuznetsova, Rom, Alldrin, Uijlings, Krasin, Pont{-}Tuset, Kamali, Popov, Malloci, Kolesnikov, Duerig, and Ferrari]{kuznetsova20ijcv}
Alina Kuznetsova, Hassan Rom, Neil Alldrin, Jasper R.~R. Uijlings, Ivan Krasin, Jordi Pont{-}Tuset, Shahab Kamali, Stefan Popov, Matteo Malloci, Alexander Kolesnikov, Tom Duerig, and Vittorio Ferrari.
\newblock The open images dataset {V4}.
\newblock \emph{IJCV}, 2020.

\bibitem[Lis et~al.(2019)Lis, Nakka, Fua, and Salzmann]{lis19iccv}
Krzysztof Lis, Krishna~Kanth Nakka, Pascal Fua, and Mathieu Salzmann.
\newblock Detecting the unexpected via image resynthesis.
\newblock In \emph{ICCV}, 2019.

\bibitem[Maag et~al.(2022)Maag, Chan, Uhlemeyer, Kowol, and Gottschalk]{maag22accv}
Kira Maag, Robin Chan, Svenja Uhlemeyer, Kamil Kowol, and Hanno Gottschalk.
\newblock Two video data sets for tracking and retrieval of out of distribution objects.
\newblock In \emph{ACCV}, 2022.

\bibitem[Nayal et~al.(2023)Nayal, Yavuz, Henriques, and G{\"{u}}ney]{nayal23iccv}
Nazir Nayal, Misra Yavuz, Jo{\~{a}}o~F. Henriques, and Fatma G{\"{u}}ney.
\newblock Rba: Segmenting unknown regions rejected by all.
\newblock In \emph{ICCV}, 2023.

\bibitem[Neuhold et~al.(2017)Neuhold, Ollmann, Bul{\`{o}}, and Kontschieder]{neuhold17iccv}
Gerhard Neuhold, Tobias Ollmann, Samuel~Rota Bul{\`{o}}, and Peter Kontschieder.
\newblock The mapillary vistas dataset for semantic understanding of street scenes.
\newblock In \emph{ICCV}, 2017.

\bibitem[Parikh et~al.(2024)Parikh, Saluja, Jawahar, and Sarvadevabhatla]{parikh24icra}
Chirag Parikh, Rohit Saluja, C.~V. Jawahar, and Ravi~Kiran Sarvadevabhatla.
\newblock {IDD-X:} {A} multi-view dataset for ego-relative important object localization and explanation in dense and unstructured traffic.
\newblock In \emph{{ICRA}}, pages 14815--14821. {IEEE}, 2024.

\bibitem[Pinggera et~al.(2016)Pinggera, Ramos, Gehrig, Franke, Rother, and Mester]{pinggera16iros}
Peter Pinggera, Sebastian Ramos, Stefan Gehrig, Uwe Franke, Carsten Rother, and Rudolf Mester.
\newblock Lost and found: detecting small road hazards for self-driving vehicles.
\newblock In \emph{Int. Conf. on Intelligent Robots and Systems}, 2016.

\bibitem[Rai et~al.(2023)Rai, Cermelli, Fontanel, Masone, and Caputo]{rai23iccv}
Shyam~Nandan Rai, Fabio Cermelli, Dario Fontanel, Carlo Masone, and Barbara Caputo.
\newblock Unmasking anomalies in road-scene segmentation.
\newblock In \emph{ICCV}, 2023.

\bibitem[Sakaridis et~al.(2021)Sakaridis, Dai, and Gool]{sakaridis21iccv}
Christos Sakaridis, Dengxin Dai, and Luc~Van Gool.
\newblock {ACDC:} the adverse conditions dataset with correspondences for semantic driving scene understanding.
\newblock In \emph{ICCV}, 2021.

\bibitem[Shaik et~al.(2024)Shaik, Malreddy, Billa, Chaudhary, Manchanda, and Varma]{shaik24wacv}
Furqan~Ahmed Shaik, Abhishek~Reddy Malreddy, Nikhil~Reddy Billa, Kunal Chaudhary, Sunny Manchanda, and Girish Varma.
\newblock {IDD-AW:} {A} benchmark for safe and robust segmentation of drive scenes in unstructured traffic and adverse weather.
\newblock In \emph{{WACV}}, 2024.

\bibitem[Varma et~al.(2019)Varma, Subramanian, Namboodiri, Chandraker, and Jawahar]{varma19wacv}
Girish Varma, Anbumani Subramanian, Anoop~M. Namboodiri, Manmohan Chandraker, and C.~V. Jawahar.
\newblock {IDD:} {A} dataset for exploring problems of autonomous navigation in unconstrained environments.
\newblock In \emph{Winter Conf. Appl. of Comput. Vis.}, 2019.

\bibitem[Vojir et~al.(2021)Vojir, \v{S}ipka, Aljundi, Chumerin, Reino, and Matas]{vojir2021}
Tomas Vojir, Tom\'a\v{s} \v{S}ipka, Rahaf Aljundi, Nikolay Chumerin, Daniel~Olmeda Reino, and Jiri Matas.
\newblock {Road Anomaly Detection by Partial Image Reconstruction With Segmentation Coupling}.
\newblock In \emph{ICCV}, pages 15651--15660, 2021.

\bibitem[Voj{\'\i}\v{r} and Matas(2023)]{vojir2023}
Tom\'a\v{s} Voj{\'\i}\v{r} and Ji\v{r}{\'\i} Matas.
\newblock {Image-Consistent Detection of Road Anomalies As Unpredictable Patches}.
\newblock In \emph{{WACV}}, pages 5491--5500, 2023.

\bibitem[Vojíř et~al.(2024)Vojíř, Šochman, and Matas]{vojir2024}
Tomáš Vojíř, Jan Šochman, and Jiří Matas.
\newblock {PixOOD: Pixel-Level Out-of-Distribution Detection}.
\newblock In \emph{ECCV}, 2024.

\bibitem[Yu et~al.(2020)Yu, Chen, Wang, Xian, Chen, Liu, Madhavan, and Darrell]{yu20cvpr}
Fisher Yu, Haofeng Chen, Xin Wang, Wenqi Xian, Yingying Chen, Fangchen Liu, Vashisht Madhavan, and Trevor Darrell.
\newblock {BDD100K:} {A} diverse driving dataset for heterogeneous multitask learning.
\newblock In \emph{CVPR}, 2020.

\bibitem[Zendel et~al.(2022)Zendel, Sch\"orghuber, Rainer, Murschitz, and Beleznai]{Zendel2022}
Oliver Zendel, Matthias Sch\"orghuber, Bernhard Rainer, Markus Murschitz, and Csaba Beleznai.
\newblock Unifying panoptic segmentation for autonomous driving.
\newblock In \emph{CVPR}, pages 21351--21360, 2022.

\bibitem[Zhou et~al.(2019)Zhou, Zhao, Puig, Xiao, Fidler, Barriuso, and Torralba]{zhou19ijcv}
Bolei Zhou, Hang Zhao, Xavier Puig, Tete Xiao, Sanja Fidler, Adela Barriuso, and Antonio Torralba.
\newblock Semantic understanding of scenes through the {ADE20K} dataset.
\newblock \emph{IJCV}, 2019.

\end{thebibliography}
}

\clearpage
\setcounter{page}{1}

\twocolumn[{%
\renewcommand\twocolumn[1][]{#1}%
\maketitlesupplementary

\pgfplotstableread[row sep=\\,col sep=&]{
  Class & Train & Static & Temporal \\
  1 & 2655816829 &  610812964 & 757220930\\
  2 & 40832626 & 13055869 & 6942879 \\
  3 & 372391332 & 131842967 & 110630805\\
  4 & 117241224 & 35522412 & 26264621\\
  5 & 43411094 & 11012641 & 796057\\
  6 & 81584995 & 21523877 & 22125102\\
  7 & 485741 & 236530 & 171925 \\
  8 & 13434914 & 1000802 & 1619492 \\
  9 & 2584622669 & 369376535 & 476717440 \\
  10 & 149627087 & 15993597 & 14563101\\
  11 & 2875216777 & 428612022 & 490239780 \\
  12 & 55937252 & 8999708 & 6739329 \\
  13 & 52477656 & 9683517 & 8741689 \\
  14 & 393796009 & 50402000 & 38605960 \\
  15 & 137055310 & 21541675 & 28032950 \\
  16 & 94958964 & 9718216 & 9164502 \\
  17 & 180313 & 1500 & 11117 \\
  18 & 66208552 & 16098392 & 14047067\\
  19 & 1662664 & 493775 & 311158\\
  20 & 0.1 & 42794949 & 28705066\\
  21 & 959803592 & 160750228 & 309811430\\
  }\mydata
\begin{center}
    \centering    
    \captionsetup{type=figure}
    \begin{tikzpicture}[font=\small]
   \definecolor{roadc}{RGB}{128, 64, 128} %
   \definecolor{sidewalkc}{RGB}{244, 35, 232}
   \definecolor{buildingc}{RGB}{70, 70, 70}
   \definecolor{wallc}{RGB}{102, 102, 156}
   \definecolor{fencec}{RGB}{190, 153, 153}
   \definecolor{polec}{RGB}{153, 153, 153}
   \definecolor{trafficlightc}{RGB}{250, 170, 30}
   \definecolor{trafficsignc}{RGB}{220, 220, 0}
   \definecolor{vegetationc}{RGB}{107, 142, 35}
   \definecolor{terrainc}{RGB}{152, 251, 152}
   \definecolor{skyc}{RGB}{70, 130, 180}
   \definecolor{personc}{RGB}{220, 20, 60}
   \definecolor{riderc}{RGB}{255, 0, 0}
   \definecolor{carc}{RGB}{0, 0, 142}
   \definecolor{truckc}{RGB}{0, 0, 70}
   \definecolor{busc}{RGB}{0, 60, 100}
   \definecolor{trainc}{RGB}{0, 80, 100}
   \definecolor{motorcyclec}{RGB}{0, 0, 230}
   \definecolor{bicyclec}{RGB}{119, 11, 32}
   \definecolor{anomalyc}{RGB}{169, 187, 214}
   \definecolor{voidc}{RGB}{0, 0, 0}
  \begin{axis}[
  ybar,
  log origin=infty, %
  ymode=log, %
  ymin=1, %
  ylabel={\# of pixels(log-scale)}, 
  width=\textwidth,
  height=4.8cm,
  bar width=3pt,
  enlarge x limits=0.03,
  symbolic x coords={1,2,3,4,5,6,7,8,9,10,11,12,13,14,15,16,17,18,19,20,21},
  xtick=data,
  xticklabels={
    road~{\textcolor{roadc}{\rule{9pt}{3pt}}},
    sidewalk~{\textcolor{sidewalkc}{\rule{9pt}{3pt}}},
    building~{\textcolor{buildingc}{\rule{9pt}{3pt}}}, 
    wall~{\textcolor{wallc}{\rule{9pt}{3pt}}}, 
    fence~{\textcolor{fencec}{\rule{9pt}{3pt}}},
    pole~{\textcolor{polec}{\rule{9pt}{3pt}}},
    traffic light~{\textcolor{trafficlightc}{\rule{9pt}{3pt}}},
    traffic sign~{\textcolor{trafficsignc}{\rule{9pt}{3pt}}},
    vegetation~{\textcolor{vegetationc}{\rule{9pt}{3pt}}},
    terrain~{\textcolor{terrainc}{\rule{9pt}{3pt}}},
    sky~{\textcolor{skyc}{\rule{9pt}{3pt}}},
    person~{\textcolor{personc}{\rule{9pt}{3pt}}},
    rider~{\textcolor{riderc}{\rule{9pt}{3pt}}},
    car~{\textcolor{carc}{\rule{9pt}{3pt}}},
    truck~{\textcolor{truckc}{\rule{9pt}{3pt}}},
    bus~{\textcolor{busc}{\rule{9pt}{3pt}}},
    train~{\textcolor{trainc}{\rule{9pt}{3pt}}},
    motorcycle~{\textcolor{motorcyclec}{\rule{9pt}{3pt}}},
    bicycle~{\textcolor{bicyclec}{\rule{9pt}{3pt}}},
    anomaly~{\textcolor{anomalyc}{\rule{9pt}{3pt}}},
    void~{\textcolor{voidc}{\rule{9pt}{3pt}}}},
  x tick label style={rotate=90, anchor=east, xshift=-3pt},
  legend style={
    at={(0.5,1.1)},
    anchor=south,
    legend columns=-1,
    /tikz/every even column/.append style={column sep=1cm}
    }, 
  legend cell align={left}
  ]
  \addplot table[x=Class, y=Train]{\mydata};
  \addlegendentry{\IDDDatasetTrain}
  \addplot table[x=Class, y=Static]{\mydata};
  \addlegendentry{\IDDDatasetStatic}
  \addplot table[x=Class, y=Temporal]{\mydata};
  \addlegendentry{\IDDDatasetTemporal}
  \end{axis}
\end{tikzpicture}
    
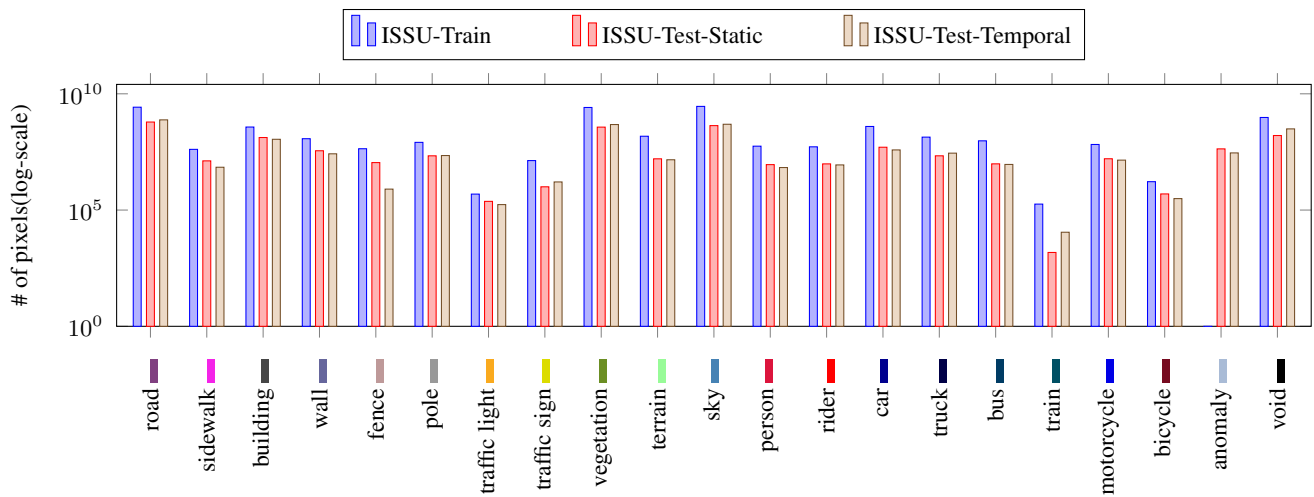
\captionof{figure}{\textbf{Dataset statistics}. The number of annotated pixels per class and their associated class labels for each part (\IDDDatasetTrain, \IDDDatasetStatic, and \IDDDatasetTemporal) of the proposed dataset.}
    \label{fig:stats}
\end{center}%
}]

In the supplementary, we provide additional results in Sec.~\ref{sec:supp_se}, implementation details of benchmarked methods in Sec.~\ref{sec:supp_impl}, dataset composition process in Sec.~\ref{sec:supp_dataset_curation} and comparison with existing anomaly segmentation datasets in Sec.~\ref{sec:supp_metric_comp}.
\section{Additional Results}\label{sec:supp_se}
In this section, we provide extended results and additional analysis. \Cref{sec:supp_stats} shows the train and test statistics, \Cref{sec:supp_results_cd} presents the results of the cross-domain evaluation, while \cref{sec:supp_qualitative_fpr,sec:addn_qualt} provides qualitative examples of undetected anomalies which are the primary contributors to the high FPR metric. Furthermore, \cref{sec:supp_results_ab_size,sec:supp_f1} include additional ablation studies and detailed analyses.

\subsection{Statistics}\label{sec:supp_stats}
The number of pixels (log-scale) per class in \IDDDatasetTrain, \IDDDatasetStatic, \IDDDatasetTemporal is shown in Fig.~\ref{fig:stats}. As can be seen, the distribution of pixel counts per class is similar between train and test splits. Additionally, \cref{tab:supp_light_stats} provides statistics on the number of normal and adverse images across different \IDDDataset splits.
\begin{table}[t!]
    \centering
    \begin{tabular}{lcc}
    \toprule
         Dataset & Day & Lowlight \\
        \midrule
         \IDDDatasetTrain    & 2690  & 746 \\
         \IDDDatasetStatic   & 848 & 132 \\
         \IDDDatasetTemporal & 868 & 270 \\
        \bottomrule
    \end{tabular}
    \caption{\textbf{Statistics of Day and Lowlight across the train and test splits of the proposed dataset}.}
    \label{tab:supp_light_stats}
\end{table}

\subsection{Cross-domain Results}\label{sec:supp_results_cd}
Complete results for the cross-domain evaluation, \ie, training on CityScapes and evaluating on the proposed \IDDDataset, are provided in \cref{tab:supp_main_anomaly} and \cref{tab:supp_main_obstacle} for the road anomaly and road obstacle evaluation protocols, respectively.

The effects of cross-domain evaluation are less pronounced for the road obstacle evaluation, \ie, where only the road region and anomalies are considered due to the high visual similarity of road regions across domains. In this setting, pixel-level methods demonstrate better robustness.
\begin{table*}[t!]

\centering\resizebox{0.98\textwidth}{!}{
\begin{tabular}{ccccccccc|cccccc}
\toprule

{} & \multirow{2}{*}{Method} & \multirow{2}{*}{\shortstack[c]{OOD\\ Data}} & \multicolumn{6}{c}{Static} & \multicolumn{6}{c}{Temporal}\\

\cmidrule(lr){4-9}
\cmidrule(lr){10-15}

{} & {} & {} & \multicolumn{3}{c}{Road Anomaly} & \multicolumn{3}{c}{Closed \& Open-set} & 
               \multicolumn{3}{c}{Road Anomaly} & \multicolumn{3}{c}{Closed \& Open-set} \\

\cmidrule(lr){4-6}
\cmidrule(lr){7-9}
\cmidrule(lr){10-12}
\cmidrule(lr){13-15}

{} & {} & {} & 
\multicolumn{1}{c}{AP $\uparrow$} & \multicolumn{1}{c}{FPR$_T$ $\downarrow$} & \multicolumn{1}{c}{TPR$_F$ $\uparrow$} & \multicolumn{1}{c}{\text{IoU} $\uparrow$} & 
\multicolumn{1}{c}{oIoU$_{T}$ $\uparrow$} & \multicolumn{1}{c}{oIoU$_{F}$ $\uparrow$} & \multicolumn{1}{c}{AP $\uparrow$} & \multicolumn{1}{c}{FPR$_T$ $\downarrow$} & \multicolumn{1}{c}{TPR$_F$ $\uparrow$} & \multicolumn{1}{c}{$\text{IoU} \uparrow$} & 
\multicolumn{1}{c}{oIoU$_{T}$ $\uparrow$} & \multicolumn{1}{c}{oIoU$_{F}$ $\uparrow$} \\

\midrule
   \multirow{3}{*}{\rotatebox[origin=c]{90}{\it \scriptsize pixel-level}} 
                         & JSR-Net\textdagger & \xmark &  3.60 &       55.71 &       5.06 &         45.57 &              8.07 &             36.64 & 2.21 &       69.51 &       5.10 &         19.70 &              3.03 &             15.02   \\
    {}                   & DaCUP\textdagger   & \xmark &  5.16 &       50.69 &      16.35 &         46.35 &              8.81 &             35.45 & 2.61 &       66.03 &      13.88 &         22.63 &              3.87 &             16.61  \\
    {}                   & PixOOD             & \xmark & 11.44 &       73.73 &      33.19 &         56.30 &             20.36 &             52.84 & 4.81 &       80.74 &      25.53 &         48.67 &             14.72 &             46.99  \\
\cmidrule(lr){1-9}
\cmidrule(lr){10-15}
    \multirow{8}{*}{\rotatebox[origin=c]{90}{\it \scriptsize mask-level}}

    {}                   & RbA                & \xmark & 43.31 &       97.30 &      70.47 &         57.17 &              4.12 &             55.24 & 15.66 &       98.46 &      46.21 &         41.33 &              1.15 &             40.56  \\
    {}                   & EAM                & \xmark & 51.49 &       96.32 &      68.83 &         65.58 &              4.82 &             61.98 & 30.28 &       96.12 &      53.51 &         56.04 &              2.86 &             51.87 \\
    {}                   & Pebal              & \xmark & 38.80 &       96.62 &      71.09 &         57.17 &              5.29 &             55.51 & 14.79 &       96.84 &      46.86 &         41.33 &              3.07 &             40.69 \\
\cmidrule(lr){2-9}
\cmidrule(lr){10-15}

    {}                   & RbA                & \cmark & 56.39 &       80.75 &      78.98 &         57.50 &             11.88 &             55.12 & 24.64 &       91.56 &      54.40 &         43.72 &              3.18 &             41.97  \\
    {}                   & EAM                & \cmark & 54.54 &       95.40 &      71.74 &         66.80 &              7.94 &             63.44 &  35.57 &       96.42 &      61.97 &         57.33 &              2.72 &             53.16 \\
    {}                   & Pebal              & \cmark & 48.32 &       64.88 &      79.66 &         57.50 &             34.20 &             55.36 &  16.11 &       79.54 &      55.01 &         43.72 &              8.31 &             42.07 \\
    {}                   & UNO                & \cmark & 55.54 &       92.96 &      79.15 &         68.11 &             12.03 &             65.58 & 37.24 &       92.37 &      70.35 &         57.24 &              6.56 &             54.63 \\
    {}                   & M2A                & \cmark & 37.48 &       79.82 &      69.00 &         50.59 &             26.40 &             48.23 & 10.66 &       91.92 &      33.16 &         33.99 &             16.76 &             33.33  \\

\bottomrule
\end{tabular}
}
\caption{\textbf{Cross-domain evaluation of road anomaly, closed-set and open-set}.}

\label{tab:supp_main_anomaly}
\end{table*}%

\begin{table}[htp]

\centering\resizebox{0.45\textwidth}{!}{
\begin{tabular}{ccccc|cc}
\toprule

{} & \multirow{2}{*}{Method} & \multirow{2}{*}{\shortstack[c]{OOD\\ Data}} & \multicolumn{2}{c}{Static} & \multicolumn{2}{c}{Temporal}\\

\cmidrule(lr){4-5}
\cmidrule(lr){6-7}

{} & {} & {} & 
\multicolumn{1}{c}{AP $\uparrow$} & \multicolumn{1}{c}{FPR$_T$ $\downarrow$} &  \multicolumn{1}{c}{AP $\uparrow$} & \multicolumn{1}{c}{FPR$_T$ $\downarrow$}  \\

\midrule
   \multirow{3}{*}{\rotatebox[origin=c]{90}{\it \scriptsize pixel-level}} 
                         & JSR-Net\textdagger & \xmark & 80.70 & 11.91 & 25.45 & 41.63   \\
    {}                   & DaCUP\textdagger   & \xmark & 85.95 & 9.23  & 69.52 & 20.42  \\
    {}                   & PixOOD             & \xmark & 92.30 & 5.10  & 84.34 & 10.84 \\

\cmidrule(lr){1-5}
\cmidrule(lr){6-7}

    \multirow{8}{*}{\rotatebox[origin=c]{90}{\it \scriptsize mask-level}}

    {}                  & RbA                & \xmark & 62.40 &       99.11     & 32.48 &       99.28 \\
    {}                  & EAM                & \xmark & 57.96 &       93.83         & 37.15 &       95.44 \\
    {}                  & Pebal              & \xmark & 62.85 &       98.08   & 34.21 &       97.97  \\
\cmidrule(lr){2-5}
\cmidrule(lr){6-7}

    {}                  & RbA                & \cmark & 76.14 &       68.89    & 
    37.86 &       87.93\\
    {}                  & EAM                & \cmark & 61.35 &       93.44   & 
    43.03 &       98.26 \\
    {}                  & Pebal              & \cmark & 73.58 &       40.79   & 
    29.36 &       67.09   \\
    {}                  & UNO                & \cmark & 66.25 &       90.81   & 
    49.10 &       90.50  \\
    {}                  & M2A                & \cmark & 63.29 &       45.84     & 30.74 &       81.35  \\

\bottomrule
\end{tabular}
}
\caption{\textbf{Cross-domain evaluation of road obstacle}}
\label{tab:supp_main_obstacle}
\end{table}%

\subsection{Qualitative Results}\label{sec:supp_qualitative_fpr}
\begin{figure*}[ht]
    \centering
    \includegraphics[width=0.98\textwidth]{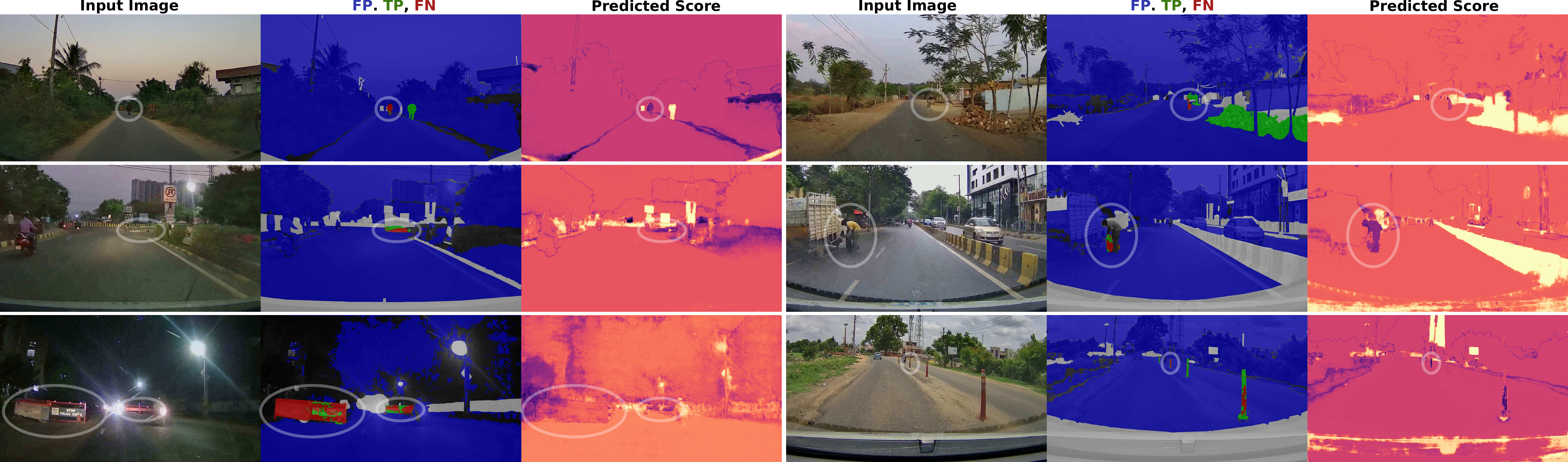}
    \caption{Qualitative results of the RbA(\xmark) at $95\%$ TPR threshold. The figure shows examples of anomalies that are not detected (fully or partially) at this threshold where most of the image pixels are falsely labeled as anomalies, resulting in very high FPR at $95\%$ TPR metric. The pixel classifications at the $95\%$ TPR threshold are coded by color overlay in the middle images -- false positive (blue), true positive (green), false negative (red), void (white) and true negative (without overlay).}
    \label{fig:supp_qualitative_rba}
\end{figure*}

To analyze the high FPR metric, particularly for Mask2Former-based methods, we conducted a visual analysis of the results from the RbA method (representative of Mask2Former-based approaches) in \cref{fig:supp_qualitative_rba}. By setting the anomaly score threshold such that the TPR metric reaches $95\%$, we observe several examples of fully (or partially) undetected anomalous instances. This behavior leads to a high FPR at this operating point, as the method includes many known-class pixels to correctly classify the ``hard'' anomalous cases. 

\begin{figure*}[!ht]
\centering
    \begin{tabular}{cccc}  %

        \color{green}{\tiny{TP}},\color{red}{\tiny{FN}},\color{blue}{\tiny{FP}} & \color{green}{\tiny{TP}},\color{red}{\tiny{FN}},\color{blue}{\tiny{FP}} & \color{green}{\tiny{TP}},\color{red}{\tiny{FN}},\color{blue}{\tiny{FP}} & \color{green}{\tiny{TP}},\color{red}{\tiny{FN}},\color{blue}{\tiny{FP}} \\  
        
        \includegraphics[width=0.23\textwidth]{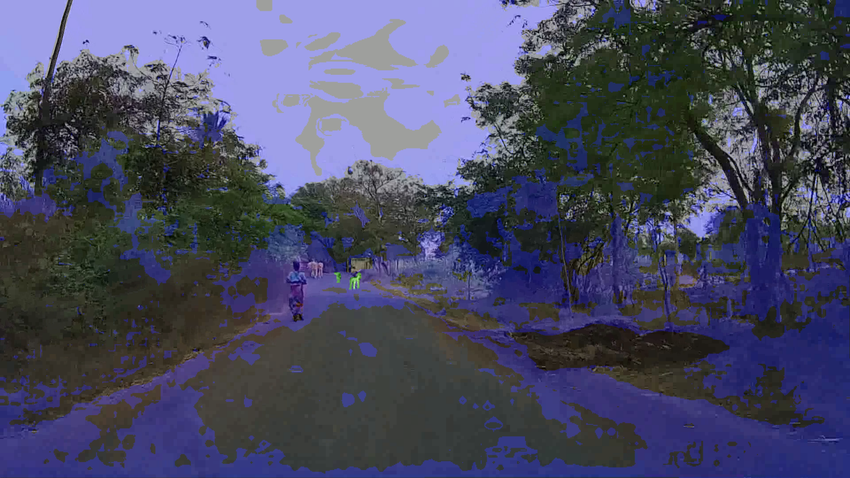}&
        \includegraphics[width=0.23\textwidth]{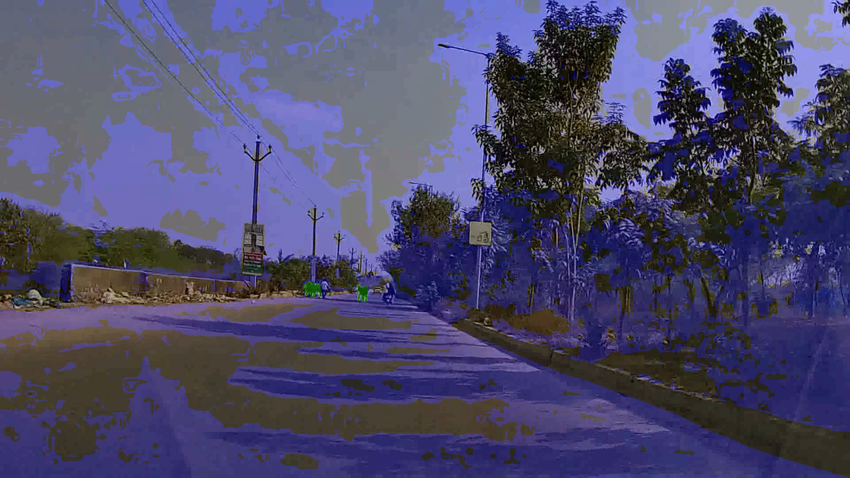}&
        \includegraphics[width=0.23\textwidth]{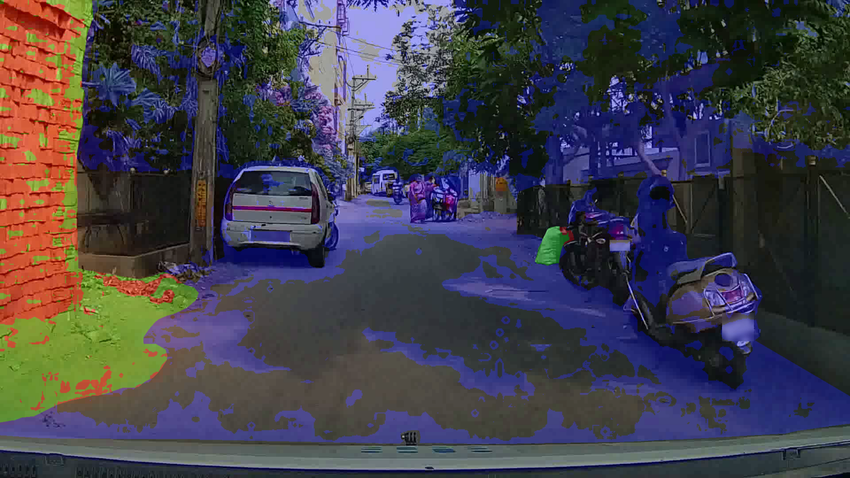}&
        \includegraphics[width=0.23\textwidth]{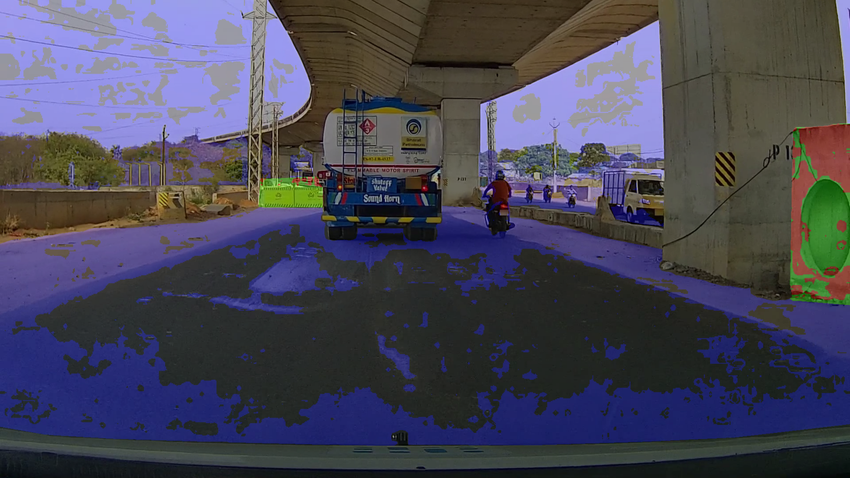} \\

        \includegraphics[width=0.23\textwidth]{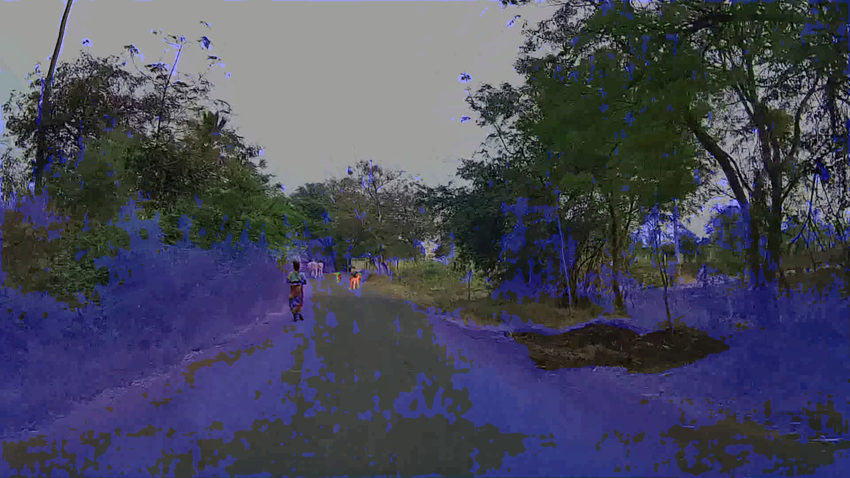} &
        \includegraphics[width=0.23\textwidth]{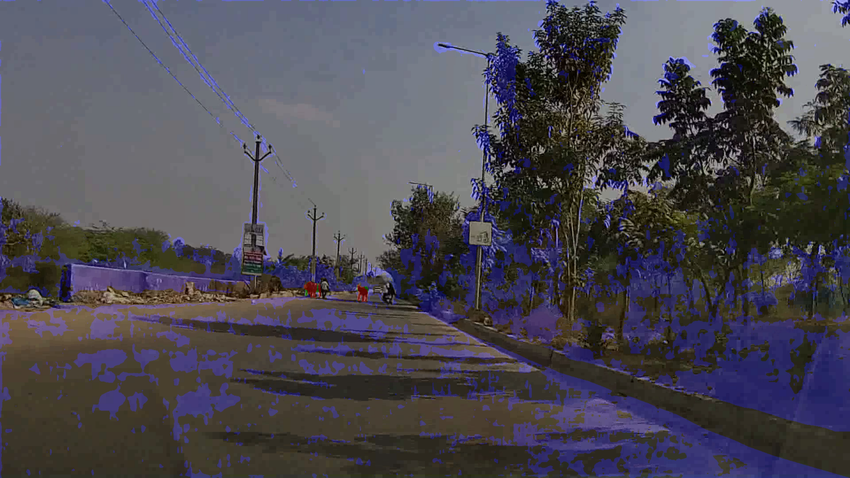} &
        \includegraphics[width=0.23\textwidth]{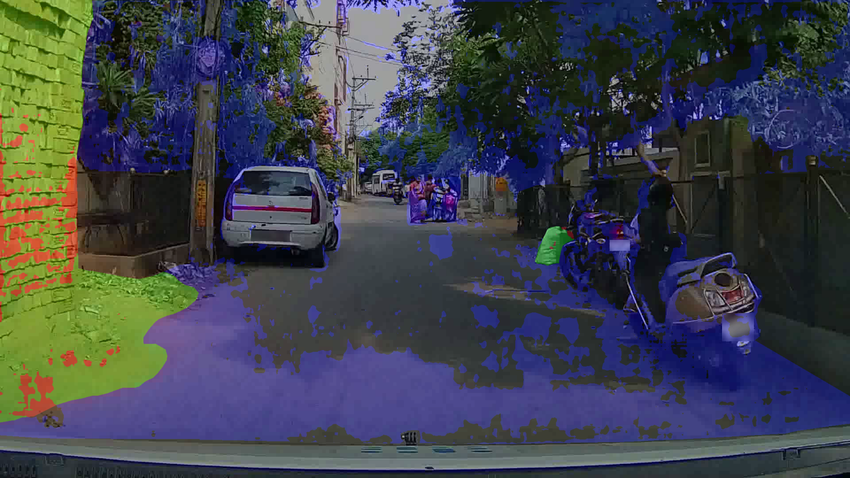} &
        \includegraphics[width=0.23\textwidth]{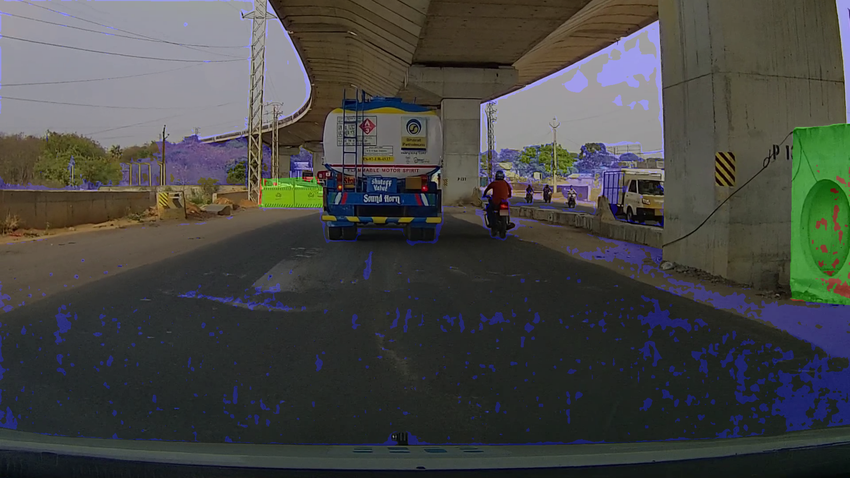} \\

         & \hspace{-40mm} \small{(a) in-domain Temporal, small anomaly} & & \hspace{-40mm} \small{(b) in-domain Temporal, very large anomaly} \\

        \color{green}{\tiny{TP}},\color{red}{\tiny{FN}},\color{blue}{\tiny{FP}} & \color{green}{\tiny{TP}},\color{red}{\tiny{FN}},\color{blue}{\tiny{FP}} & \color{green}{\tiny{TP}},\color{red}{\tiny{FN}},\color{blue}{\tiny{FP}} & \color{green}{\tiny{TP}},\color{red}{\tiny{FN}},\color{blue}{\tiny{FP}} \\  
        
        \includegraphics[width=0.23\textwidth]{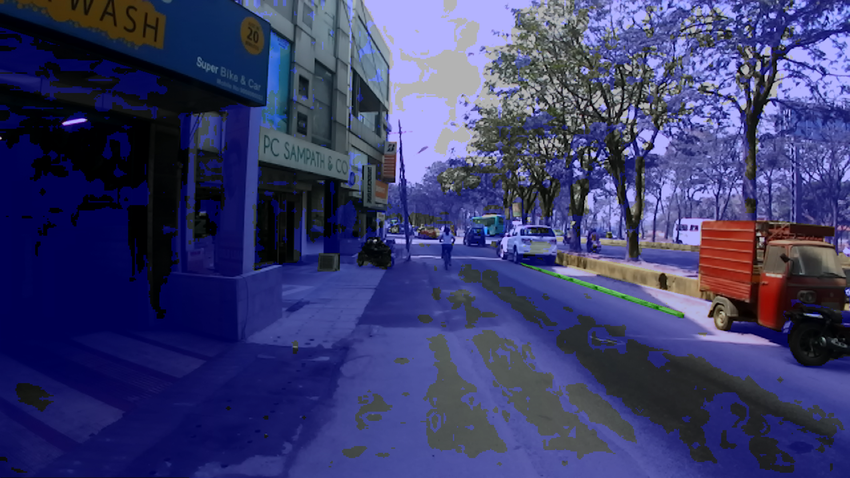}&
        \includegraphics[width=0.23\textwidth]{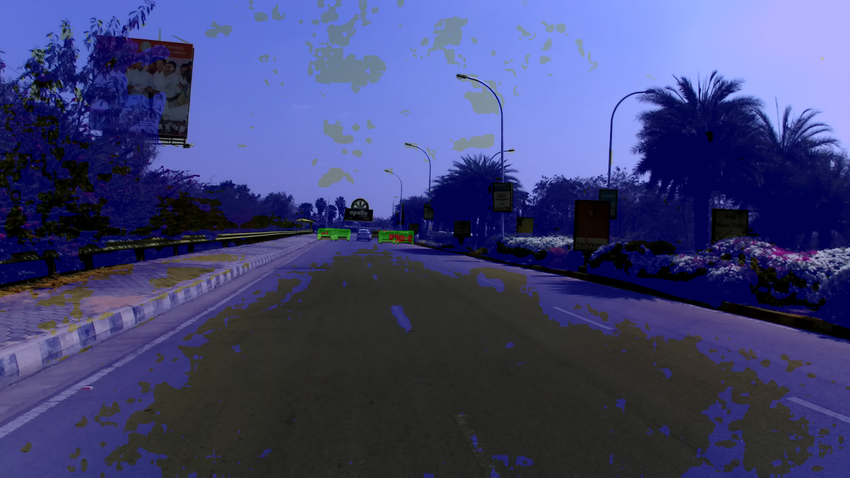}&
        \includegraphics[width=0.23\textwidth]{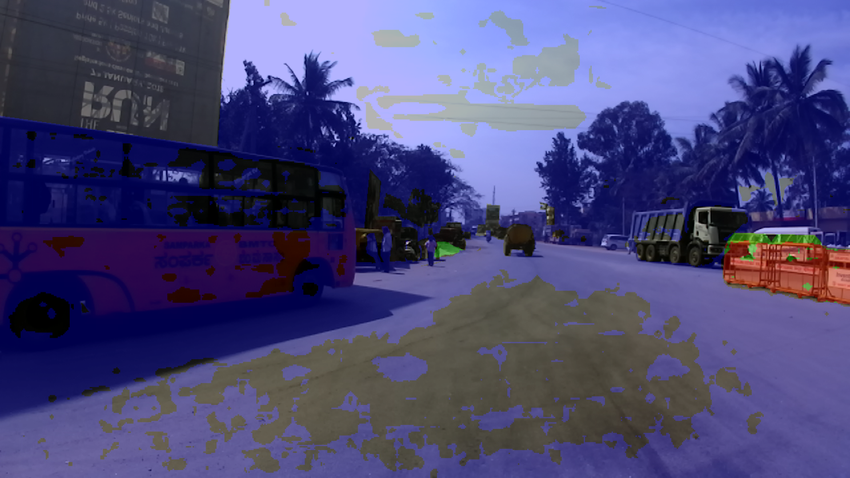}&
        \includegraphics[width=0.23\textwidth]{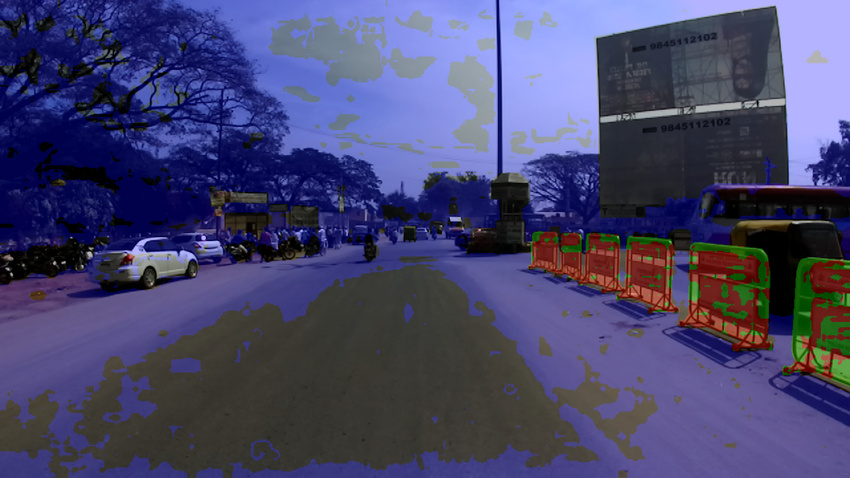} \\

        \includegraphics[width=0.23\textwidth]{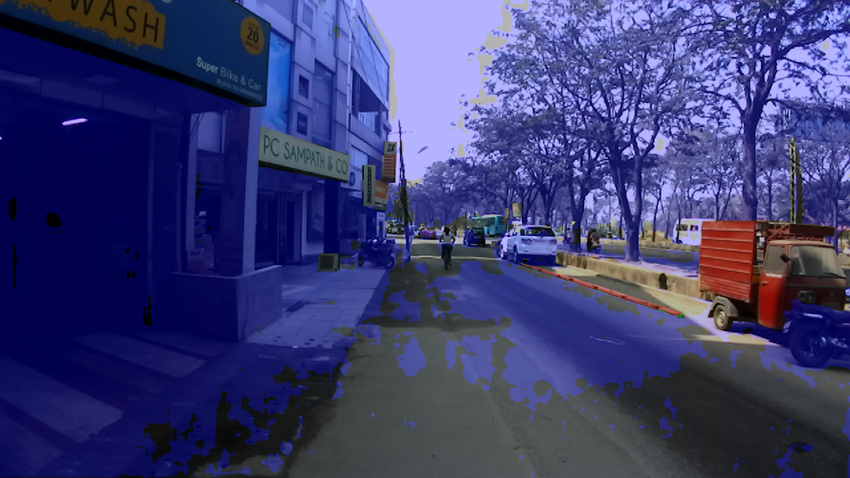} &
        \includegraphics[width=0.23\textwidth]{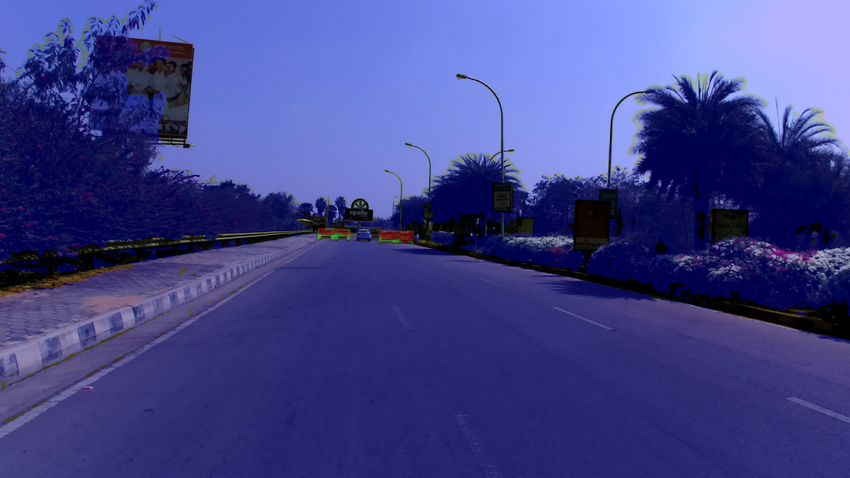} &
        \includegraphics[width=0.23\textwidth]{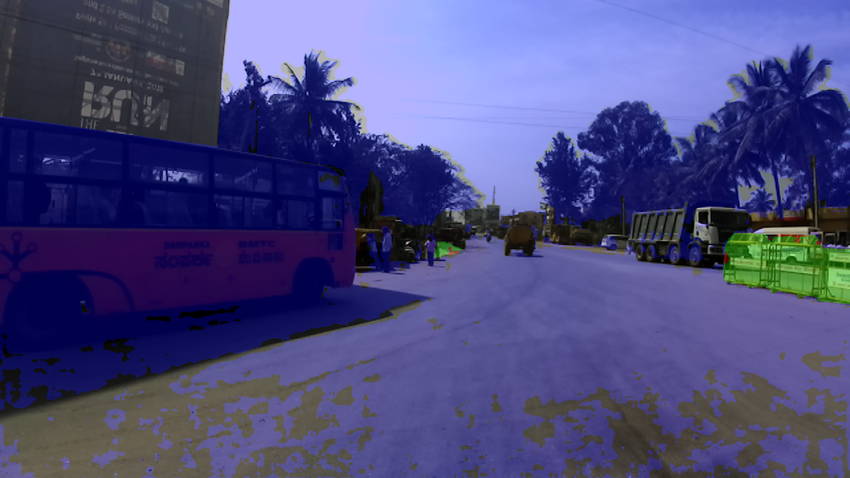} &
        \includegraphics[width=0.23\textwidth]{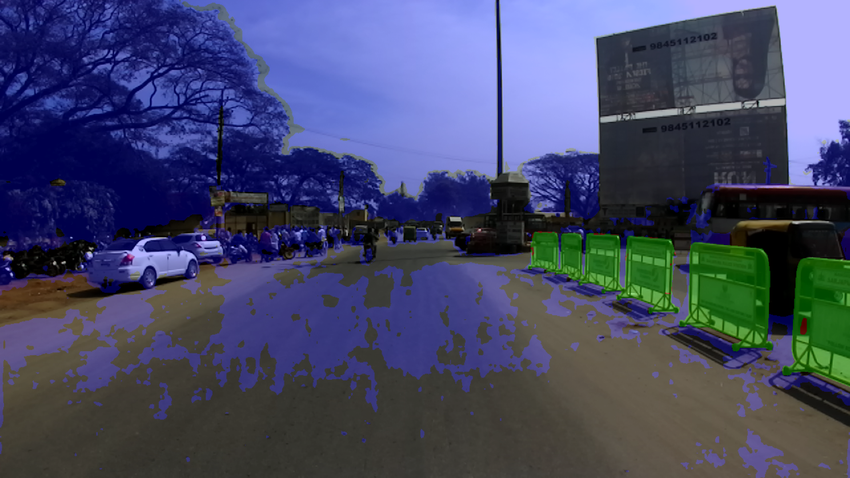} \\

         & \hspace{-40mm} \small{(a) cross-domain, small anomaly} & & \hspace{-40mm} \small{(b) cross-domain, very large anomaly} \\
        
    \end{tabular}
    \vspace{-4mm}
\caption{\textbf{Qualitative Results} shown for PixOOD ($1^{st}$ and $3^{rd}$ row) and RbA (\cmark) ($2^{nd}$ and $4^{th}$ row) across in-domain Temporal (cross-sensor) and cross-domain setups  for small and very large anomalies. Anomaly detection threshold is set based on operation point $95\%$TPR} 
\vspace{-5mm}
\label{fig:qualit_rba_pxood}
\end{figure*}

We considered cross-sensor (in-domain Temporal) and cross-domain setups, and qualitatively compared two methods: PixOOD and RbA (\cmark) in~\cref{fig:qualit_rba_pxood}. The results are shown for very large and small anomalies with \textcolor{green}{TP}, \textcolor{red}{FN} and \textcolor{blue}{FP} pixels colored accordingly. These qualitative visualizations support the findings in~\cref{fig:supp-ablation_size_static} -- pixel-level PixOOD struggles in detecting very large anomalies while being better than mask-level RbA in detecting small anomaly objects. 
The cross-sensor and cross-domain shift is challenging for both methods as shown by the known classes misclassified as anomalies (\textcolor{blue}{FP} pixels).

\vspace{-2mm}

\subsection{Ablation: Anomaly Size}\label{sec:supp_results_ab_size}
\begin{figure*}[ht]
\begin{tikzpicture}
    \begin{axis} [
        ylabel shift = -4pt,
        ybar, %
        bar width=7pt,
        width=0.98\textwidth,
        height=5cm,
        enlarge x limits=0.1,
        ymin=0, ymax=90, %
        ylabel={F1},
        xlabel={},
        xtick=data,
        ymajorgrids=true,
        major grid style={dotted,black, opacity=0.5},
        tickpos=left,
        symbolic x coords={ JSR-Net, DaCUP, PixOOD, RbA, RbAt, EAM, EAMt, M2A, UNO}, %
        xticklabels={JSR-Net~\xmark, DaCUP~\xmark, PixOOD~\xmark, RbA~\xmark, RbA~\cmark, EAM~\xmark, EAM~\cmark, M2A~\cmark, UNO~\cmark},
        x tick label style={rotate=0, anchor=center, yshift=-2pt, font=\small}, %
        xtick style={draw=none},
        legend style={at={(0.5,1.05)}, anchor=south, legend columns=-1},
        every node near coord/.append style={rotate=45,font=\normalsize,scale=0.6, yshift=4pt, xshift=8pt}, 
        nodes near coords, %
        nodes near coords align={vertical}
    ]
    \addplot[fill=Cyan] coordinates {
         (JSR-Net, 21.3) (DaCUP, 0.2) (PixOOD, 22.4) (RbA, 16.7) (RbAt, 11.9) (EAM, 12.1) (EAMt, 17.2) (M2A, 17.2) (UNO, 28.5)
    };
    
    \addplot[fill=Brown] coordinates {
         (JSR-Net, 27.9) (DaCUP, 3.9) (PixOOD, 43.6) (RbA, 36.2) (RbAt, 32.7) (EAM, 35.7) (EAMt, 39.7) (M2A, 30.4) (UNO, 51.1)
    };
    \addplot[fill=Melon] coordinates {
         (JSR-Net, 35.9) (DaCUP, 13.7) (PixOOD, 51.2) (RbA, 69.7) (RbAt, 71.4) (EAM, 69.5) (EAMt, 73.7) (M2A, 51.4) (UNO, 77.8)
    };
    \addplot[fill=LimeGreen] coordinates {
         (JSR-Net, 5.0) (DaCUP, 5.0) (PixOOD, 20.3) (RbA, 66.7) (RbAt, 69.9) (EAM, 67.9) (EAMt, 68.2) (M2A, 29.8) (UNO, 68.3)
    };

    \legend{t-Small, Small, Large, v-Large}
    \end{axis}
\end{tikzpicture}
\begin{tikzpicture}
    \begin{axis} [
        ylabel shift = -4pt,    
        ybar, %
        bar width=7pt,
        width=0.98\textwidth,
        height=5cm,
        enlarge x limits=0.1,
        ymin=0, ymax=90, %
        ylabel={F1},
        xlabel={},
        xtick=data,
        ymajorgrids=true,
        major grid style={dotted,black, opacity=0.5},
        tickpos=left,
        symbolic x coords={ JSR-Net, DaCUP, PixOOD, RbA, RbAt, EAM, EAMt, M2A, UNO}, %
        xticklabels={JSR-Net~\xmark, DaCUP~\xmark, PixOOD~\xmark, RbA~\xmark, RbA~\cmark, EAM~\xmark, EAM~\cmark, M2A~\cmark, UNO~\cmark},
        x tick label style={rotate=0, anchor=center, yshift=-2pt, font=\small}, %
        xtick style={draw=none},
        legend style={at={(0.5,1.05)}, anchor=south, legend columns=-1},
        every node near coord/.append style={rotate=45,font=\normalsize,scale=0.6, yshift=4pt, xshift=8pt}, 
        nodes near coords, %
        nodes near coords align={vertical}
    ]

    \addplot[fill=Cyan] coordinates {
         (JSR-Net, 33.5) (DaCUP, 7.0) (PixOOD, 14.6) (RbA, 14.7) (RbAt, 6.5) (EAM, 12.2) (EAMt, 17.9) (M2A, 8.5) (UNO, 35.2)
    };
    
    \addplot[fill=Brown] coordinates {
         (JSR-Net, 30.6) (DaCUP, 14.7) (PixOOD, 44.3) (RbA, 31.6) (RbAt, 20.5) (EAM, 26.6) (EAMt, 29.1) (M2A, 19.5) (UNO, 49.3)
    };
    \addplot[fill=Melon] coordinates {
         (JSR-Net, 24.5) (DaCUP, 22.0) (PixOOD, 63.1) (RbA, 56.8) (RbAt, 49.3) (EAM, 57.1) (EAMt, 60.0) (M2A, 39.5) (UNO, 73.2)
    };
    \addplot[fill=LimeGreen] coordinates {
         (JSR-Net, 2.3) (DaCUP, 2.6) (PixOOD, 9.1) (RbA, 38.6) (RbAt, 39.8) (EAM, 47.6) (EAMt, 47.9) (M2A, 14.1) (UNO, 41.9)
    };
    
    \legend{}; %
    \end{axis}
\end{tikzpicture}

\caption{\textbf{Ablation for different anomaly sizes}. Top (bottom) plot shows results for \IDDDatasetStatic (\IDDDatasetTemporal), respectively. The different anomaly sizes are defined in~\cref{fig:spatial}. The corresponding tick (\cmark / \xmark) defines trained with / without OOD data.}
\label{fig:supp-ablation_size_static}
\end{figure*}
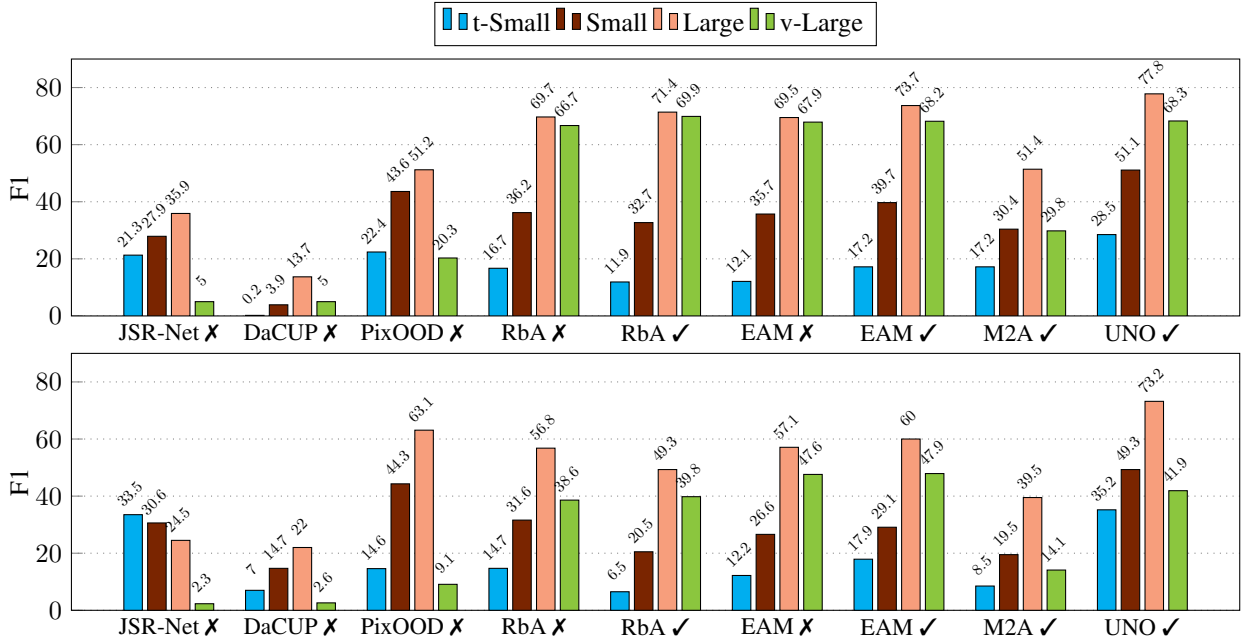

\Cref{fig:supp-ablation_size_static} presents an ablation study of the performance of all methods with respect to different anomaly sizes. The findings, consistent across all methods, align with the results presented in the main paper (\cf~\cref{fig:ablation_size_static}).

\subsection{Ablation: Effect of Anomaly Sizes to Metrics}\label{sec:supp_f1}
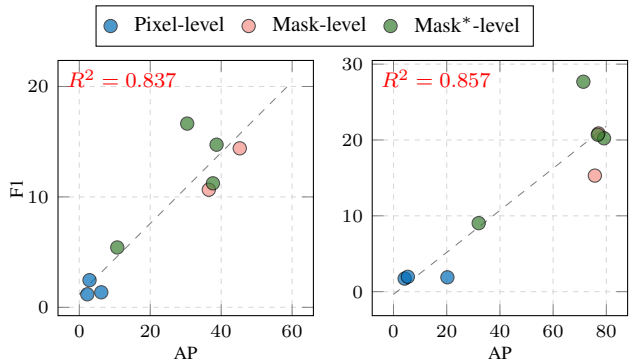
\begin{figure}[t!]
    \begin{tikzpicture}
        \begin{axis}[
            height=.1\textwidth,
            hide axis,
            xmin=0, xmax=1, ymin=0, ymax=1,
            legend columns=3,
            legend style={at={(0.,0)}, anchor=north, font=\footnotesize, column sep=1ex, row sep=0pt, fill opacity=1.0},
            legend entries={Pixel-level, Mask-level, $\text{Mask}^{*}$-level},
        ]
            \addlegendimage{only marks, mark options={draw=black}, color=NavyBlue, solid, mark=, opacity=0.7, mark size=2.5}
            \addlegendimage{only marks, mark options={draw=black}, color=Salmon, solid, mark=, opacity=0.7, mark size=2.5}   
            \addlegendimage{only marks, mark options={draw=black}, color=OliveGreen, solid, mark=, opacity=0.7, mark size=2.5}
        \end{axis}
    \end{tikzpicture}
    \centering
    \begin{tikzpicture}
        \begin{axis}[
            width=0.285\textwidth,
            height=0.285\textwidth,
            ylabel={\scriptsize F1},
            ylabel shift = -5pt,
            xlabel={\scriptsize AP},
            tick label style={font=\scriptsize},
            ylabel near ticks, xlabel near ticks, 
            xlabel style={yshift=3pt},
            grid=both,
            grid style={color=lightgray!60, dash pattern=on 2pt off 2pt},
            legend style={at={(0.5, -0.2)}, anchor=north, cells={anchor=west}, font=\tiny, fill opacity=1.0, row sep=-1pt, inner sep=2pt},
        ]

        \addplot[domain=0:60, dashed, mark=, opacity=0.5] {0.321*x + 1.171};

        \addplot[only marks, mark options={draw=black}, color=NavyBlue, solid, opacity=0.7, mark size=2.5] 
        table[meta=label] {
        x y label
        6.2	1.36	a	
        2.26	1.18	a	
        2.88	2.46	a
        };

        \addplot[only marks, mark options={draw=black}, color=Salmon, solid, opacity=0.7, mark size=2.5] 
        table[meta=label] {
        x y label
        36.51	10.65	a
        45.24	14.4	a
        };

        \addplot[only marks, mark options={draw=black}, color=OliveGreen, solid, opacity=0.7, mark size=2.5] 
        table[meta=label] {
        x y label
        37.67	11.23	a
        38.7	14.73	a
        30.42	16.64	a
        10.71	5.42	a
        };
        \node[anchor=north west, font=\footnotesize, color=red] at (rel axis cs: 0.00, 1.00) {$R^2 = 0.837$};
        \end{axis}
    \end{tikzpicture}
    \hfill
    \begin{tikzpicture}
        \begin{axis}[
            width=0.285\textwidth,
            height=0.285\textwidth,
            tick label style={font=\scriptsize},
            ylabel near ticks, xlabel near ticks, 
            xlabel={\scriptsize AP},
            xlabel style={yshift=3pt},
            grid=both,
            grid style={color=lightgray!60, dash pattern=on 2pt off 2pt},
        ]

        \addplot[domain=0:80, dashed, mark=, opacity=0.5] {0.277*x - 0.361};

        \addplot[only marks, mark options={draw=black}, color=NavyBlue, solid, opacity=0.7, mark size=2.5] 
        table[meta=label] {
        x y label
        20.26	1.89	a
        4.18	1.72	a
        5.39	1.96	a
        };

        \addplot[only marks, mark options={draw=black}, color=Salmon, solid, opacity=0.7, mark size=2.5] 
        table[meta=label] {
        x y label
        75.68	15.3	a
        77.06	20.86	a
        };

        \addplot[only marks, mark options={draw=black}, color=OliveGreen, solid, opacity=0.7, mark size=2.5] 
        table[meta=label] {
        x y label
        79.13	20.22	a
        76.79	20.69	a
        71.37	27.68	a
        32.02	9.03	a
        };
        \node[anchor=north west, font=\footnotesize, color=red] at (rel axis cs: 0.00, 1.00) {$R^2 = 0.857$};
        \end{axis}
    
    \end{tikzpicture}
     \centering

\caption{\textbf{Correlation of AP-F1}. We fit a regression line and report the correlation coefficient $R^2$ between the F1 and AP metrics. The correlation coefficient is defined as $R^2=1-\text{SSR}/\text{SST}$ (\cf~\cref{sec:supp_f1}) showing how well the regression line explains the variability of the data. The reported values (0.837 and 0.857) indicate a strong correlation for both datasets.}
\label{fig:supp_apf1_corr}    
\end{figure}

The component-level F1 metric was introduced by Chan \etal~\cite{chan21neurips} to account for small-sized anomalies. Correlation plot in Fig.~\ref{fig:supp_apf1_corr} between pixel-level metric, AP and component-level F1, shows that both these metrics are highly correlated. We hypothesize this is due to the diversity of anomaly size in our dataset. Detailed component-level metrics - F1, sIoU and PPV are provided in~\cref{tab:cd_os} for completeness following common practice~\cite{chan21neurips}.

In order to show the correlation between the F1 and AP metrics in the proposed dataset, we fit a regression line that minimizes the total squared difference (SSR) between the observed data points $\left(x_i,y_i\right)$ and the predicted values $y_{\text{pred},i}$, \ie, $y_{\text{pred}}=mx+c$, where $m$ is the slop, and $c$ is the intercept. The correlation coefficient $R^2$ measures how well the regression line explains the variability of the data. The $R^2$ value is defined as:

\begin{equation}
R^2 = 1 - \frac{\text{SSR}}{\text{SST}}
\end{equation}

\noindent where SST is a total sum of squares that measures the variability in the data relative to the mean, \ie,~$\text{SST}=\sum_{i=1}^n\left(y_i - \bar{y}\right)^2$; the residual sum of squares, SSR is a measure of the discrepancy between the actual data points and the values predicted by a regression model. It quantifies the amount of variation in the dependent variable $y$ that the model does not explain, \ie, $\text{SSR}=\sum_{i=1}^n\left(y_i - y_{\text{pred}, i}\right)^2$

\begin{table*}[t]
\centering
\setlength{\tabcolsep}{4.5pt}
\begin{tabular}{lccccc|ccc}
\toprule

\multicolumn{9}{c}{Road Anomaly} \\
\midrule

{} & 
\multirow{2}{*}{Method} & \multirow{2}{*}{\shortstack[c]{OOD\\ Data}} & \multicolumn{3}{c}{Static} & \multicolumn{3}{c}{Temporal}\\

\cmidrule(lr){4-6}
\cmidrule(lr){7-9}

  {} &  {} &  {} &
\multicolumn{1}{c}{F1 $\uparrow$} & \multicolumn{1}{c}{sIoU $\uparrow$} & \multicolumn{1}{c}{PPV $\uparrow$} &  \multicolumn{1}{c}{F1 $\uparrow$} & \multicolumn{1}{c}{sIoU $\uparrow$} & \multicolumn{1}{c}{PPV $\uparrow$}  \\

\midrule
\multirow{3}{*}{\rotatebox[origin=c]{90}{\it \scriptsize pixel-level}} 
& JSRNet\textdagger   & \xmark    & 3.2 /  1.7  & 13.8 / 14.6 & 8.2  /  3.2  &  1.2  /  1.2  &  12.0 / 13.5 & 4.4  /  2.6  \\
& DaCUP\textdagger    & \xmark        & 1.2 /  2.0  & 8.7  /  7.0  &  7.5  /  6.6  &  0.9  /  2.5  &  4.3  /  8.3  &  5.4  /  6.1 \\
& PixOOD  & \xmark      & 1.8 /  1.9  & 15.6 / 27.5 & 13.4 / 7.6  &  1.4  / 1.4  &  14.1 / 24.7 & 7.8  /  3.7 \\
\cmidrule(lr){1-6}
\cmidrule(lr){7-9}

\multirow{8}{*}{\rotatebox[origin=c]{90}{\it \scriptsize mask-level}}
& RbA    & \xmark       & 11.2 / 15.3 & 28.5 / 36.7 & 19.9 / 18.2 & 5.7  /  10.7 & 17.5 / 25.6 & 12.4 / 17.8 \\
& EAM    & \xmark       & 20.2 / 20.9 & 29.3 / 35.8 & 23.2 / 23.2 & 11.7 / 14.4 & 19.1 / 25.4 & 18.0 / 20.3 \\
& Pebal  & \xmark       & 11.8 / 17.6 & 27.2 / 34.2 & 21.7 / 23.0 & 6.3  /  11.3 & 14.1 / 23.2 & 17.2 / 20.5 \\

\cmidrule(lr){2-6}
\cmidrule(lr){7-9}

& RbA  & \cmark     & 9.6 /  20.2 & 33.2 / 36.9 & 15.5 / 25.7 & 5.3  /  11.2 & 20.0 / 21.6 & 11.9 / 23.9 \\
& EAM  & \cmark     & 21.5 / 20.7 & 30.4 / 39.1 & 25.2 / 23.0 & 10.6 / 14.7 & 26.0 / 27.8 & 13.5 / 20.2 \\
& Pebal & \cmark    & 13.0 / 0.0  &  29.4 / 0.0  &  25.2 / 0.0  &  0.0  /  0.0  &  0.0  /  0.0  &  0.0  /  0.0 \\
& UNO   & \cmark        & 27.8 / 27.7 & 27.8 / 44.3 & 43.3 / 29.1 & 18.6 / 16.6 & 22.8 / 37.9 & 28.4 / 17.8 \\
& M2A  & \cmark     & 10.8 / 9.0  &  27.3 / 25.5 & 18.3 / 17.5 & 4.4  /  5.4  &  8.8  /  16.1 & 15.3 / 15.4  \\

\bottomrule

\multicolumn{9}{c}{Road Obstacle} \\
\midrule

 {} & 
\multirow{2}{*}{Method} & \multirow{2}{*}{\shortstack[c]{OOD\\ Data}} & \multicolumn{3}{c}{Static} & \multicolumn{3}{c}{Temporal}\\

\cmidrule(lr){4-6}
\cmidrule(lr){7-9}

{} &   {} &  {} &
\multicolumn{1}{c}{F1 $\uparrow$} & \multicolumn{1}{c}{sIoU $\uparrow$} & \multicolumn{1}{c}{PPV $\uparrow$} &  \multicolumn{1}{c}{mF1 $\uparrow$} & \multicolumn{1}{c}{sIoU $\uparrow$} & \multicolumn{1}{c}{PPV $\uparrow$}  \\

\midrule
\multirow{3}{*}{\rotatebox[origin=c]{90}{\it \scriptsize pixel-level}} 
& JSRNet\textdagger   & \xmark     & 31.2 / 24.3 & 55.4 / 62.4 & 33.2 / 23.8 & 11.5 / 18.6 & 25.4 / 45.1 & 28.1 / 26.6 \\
& DaCUP\textdagger    & \xmark      & 28.1 / 28.2 & 62.8 / 53.0 & 22.3 / 24.5 & 31.0 / 24.1 & 47.7 / 41.6 & 32.7 / 25.5 \\
& PixOOD  & \xmark &  28.0 / 27.9 & 58.9 / 64.6 & 27.3 / 22.9 & 33.6 / 28.2 & 50.5 / 53.7 & 35.8 / 27.6 \\

\cmidrule(lr){1-6}
\cmidrule(lr){7-9}

\multirow{8}{*}{\rotatebox[origin=c]{90}{\it \scriptsize mask-level}}

& RbA     & \xmark   & 25.6 / 29.5 & 37.7 / 51.7 & 38.6 / 30.1 & 13.7 / 22.5 & 26.0 / 41.8 & 25.1 / 27.4 \\
& EAM     & \xmark   & 36.4 / 32.7 & 31.7 / 55.5 & 51.8 / 28.4 & 19.3 / 27.5 & 28.8 / 43.3 & 27.7 / 28.2 \\
& Pebal   & \xmark   & 25.6 / 30.8 & 37.7 / 51.1 & 38.7 / 32.4 & 13.7 / 22.5 & 26.6 / 42.5 & 25.0 / 27.2 \\

\cmidrule(lr){2-6}
\cmidrule(lr){7-9}

& RbA    & \cmark & 17.5 / 37.0 & 40.9 / 50.9 & 24.6 / 40.1 & 9.3 / 23.5 & 26.6 / 39.1 & 18.0 / 31.9 \\
& EAM    & \cmark & 36.3 / 41.4 & 35.5 / 56.7 & 47.9 / 38.7 & 23.6 / 28.0 & 30.8 / 48.4 & 32.4 / 26.8 \\
& Pebal  & \cmark & 19.0 / 37.4 & 40.3 / 50.5 & 26.8 / 41.1 & 10.8 / 25.5 & 23.1 / 35.3 & 22.8 / 38.6 \\
& UNO    & \cmark    & 38.5 / 41.5 & 29.0 / 62.1 & 67.9 / 35.0 & 26.2 / 31.1 & 31.9 / 52.2 & 34.6 / 28.7 \\
& M2A    & \cmark & 24.5 / 22.4 & 36.9 / 39.3 & 35.0 / 31.0 & 12.4 / 15.4 & 21.9 / 25.6 & 25.3 / 31.2 \\

\bottomrule
\end{tabular}
\vspace{-3mm}
\caption{Component-level metrics for road anomaly (top) and  obstacle (bottom) tracks in the form \textbf{cross-domain/in-domain}.}
\vspace{-5mm}
\label{tab:cd_os}
\end{table*}%

\begin{table}[ht!]
  \centering
  \small
  \setlength{\tabcolsep}{3.7pt}
  \renewcommand*{\arraystretch}{1}
  \begin{tabular}{@{}lccc}
    \toprule
    \multirow{2}{*}{Class} & \multicolumn{3}{c}{Mapping} \\
     & CityScapes (C) & Anomaly (A) & Void (V) \\
    \midrule
    road & \cmark &  & \\
    parking & \cmark & & \\
    drivable fallback & \cmark & & \\
    sidewalk & \cmark & & \\
    non-drivable fallback & & & \cmark \\
    person & \cmark & & \\
    animal & & \cmark & \\
    rider & \cmark & & \\
    motorcycle & \cmark & & \\
    bicycle & \cmark & & \\
    auto-rickshaw & & & \cmark \\
    car & \cmark & & \\
    truck & \cmark & & \\
    bus & \cmark & & \\
    caravan & & & \cmark \\
    vehicle-fallback & & \cmark & \cmark\\
    curb & & \cmark & \cmark\\
    wall & \cmark & & \\
    fence & \cmark & & \\
    guard rail & & \cmark & \cmark \\
    billboard & & & \cmark \\
    traffic-sign & \cmark & \cmark & \\
    traffic-light & \cmark & & \\
    pole & \cmark & & \\
    obs-str-bar-fallback & & \cmark & \cmark \\
    building & \cmark & & \\
    bridge & & & \cmark \\
    vegetation & \cmark & & \\
    sky & \cmark & & \\
    fallback-background & & & \cmark \\
    \bottomrule
  \end{tabular}
  \caption{\textbf{Dataset annotation protocol}. The mapping between the level-4 label hierarchy of IDD dataset and corresponding CityScapes (C), Anomaly (A), and Void (V) labels in our proposed datasets is indicated by the \cmark~tick.}
  \label{tab:class-mapping}
\end{table}

\section{Implementation Details}
\label{sec:supp_impl}

\noindent
\textbf{Pixel-level baselines.}
We implement JSR-Net\footnote{\url{https://github.com/vojirt/JSRNet}} and DaCUP\footnote{\url{https://github.com/vojirt/DaCUP}} baselines by extending the publicly available code releases.
Both baselines extend the DeepLabV3 segmentation model with specialized plug-in modules for anomaly detection.
Thus, we follow the optimization procedure and hyperparameters reported in the original papers \cite{vojir2021} and \cite{vojir2023}.
Similarly, we extend the publicly available code
of the PixOOD\footnote{\url{https://github.com/vojirt/PixOOD}} baseline.
This baseline relies on a generic feature extractor, so we use ViT-L trained with DINOv2 as suggested in \cite{vojir2024}.
Other hyperparameters follow the reported values as well.

\noindent
\textbf{Mask-level baselines.}
All mask-level baselines extend the Mask2Former architecture with anomaly detection capabilities. 
In the case of EAM and UNO\footnote{\url{https://github.com/matejgrcic/Open-set-M2F}} we use the default Mask2Former upsampling and SWIN-L backbone pretrained on ImageNet-22k, as suggested in the corresponding manuscripts \cite{grcic23cvprw,delic24bmvc}.
In the case of the RbA\footnote{\url{https://github.com/NazirNayal8/RbA}} baseline, we use SWIN-B and a single transformer decoder layer.
This architecture was validated as optimal for RbA \cite{nayal23iccv}.
We use the same architecture when adapting the pixel-level baseline PEBAL to mask-level predictions.
Finally, we use a frozen ResNet-50 feature extractor pretrained on ImageNet for the Mask2Anomaly baseline.
Again, this backbone was validated as optimal for Mask2Anomaly \cite{rai23iccv}.
We use the default hyperparameter values reported in the corresponding manuscripts for all baselines.

\section{Dataset Composition}\label{sec:supp_dataset_curation}

\begin{table}[ht!]
    \centering
    \small
    \setlength{\tabcolsep}{2.3pt}
    \renewcommand*{\arraystretch}{1}
    \begin{tabular}{lcccrc}
    \toprule
        Datasets & Eval & F1$\uparrow$ & AP$\uparrow$ & FPR$\downarrow$ & oIoU$_T\uparrow$ \\
    \midrule
        LostAndFound'16~\cite{pinggera16iros}    & RO & 61.7 & 89.2 & 0.6 & N/A \\
        SOS'22~\cite{maag22accv}                 & RO & 53.6 & 89.5 & 0.3 & N/A \\
        WOS'22~\cite{maag22accv}                 & RO & 48.5 & 93.8 & 0.8 & N/A \\
        SMIYC-RoadObstacle'21~\cite{chan21neurips} & RO & 75.0 & 95.1 & 0.1 & N/A \\
    \midrule
        Street-hazards'22~\cite{hendrycks22icml} & RA & N/A & 58.1 & 13.0 & 59.8\\
        Fishyscapes-static'21~\cite{blum21ijcv}  & RA & N/A & 96.8 & 0.3 & N/A \\
        
        Fishyscapes-LaF'21~\cite{blum21ijcv}     & RA & N/A & 74.8 & 1.3 & N/A \\
        SMIYC-RoadAnomaly'21~\cite{chan21neurips} & RA & 60.9 & 94.5 & 4.1 & N/A \\
        
        \midrule
        \IDDDatasetStatic'24                     & RO & 41.5 & 95.8 & 1.2 & N/A \\   
        \IDDDatasetTemporal'24                   & RO & 31.1 & 83.1 & 10.1 & N/A \\
        \IDDDatasetStatic'24                     & RA & 27.7 & 79.2 & 3.0 & 68.4 \\   
        \IDDDatasetTemporal'24                   & RA & 18.5 & 45.2 & 24.7 & 46.2 \\
    \bottomrule
    \end{tabular}
    \caption{\textbf{The datasets performance comparison}. For different evaluation protocols - road obstacle (RO) and road anomaly (RA), best values obtained by any method across different metrics: F1, AP, FPR at 95\%TPR (FPR), open-IoU at 95\%TPR (oIoU$_T$) are presented.  }
    \label{tab:supp_dataset_compare_metric}
\end{table}

\begin{figure*}
    \centering
    \includegraphics[width=.9\textwidth, height=0.8\textheight]{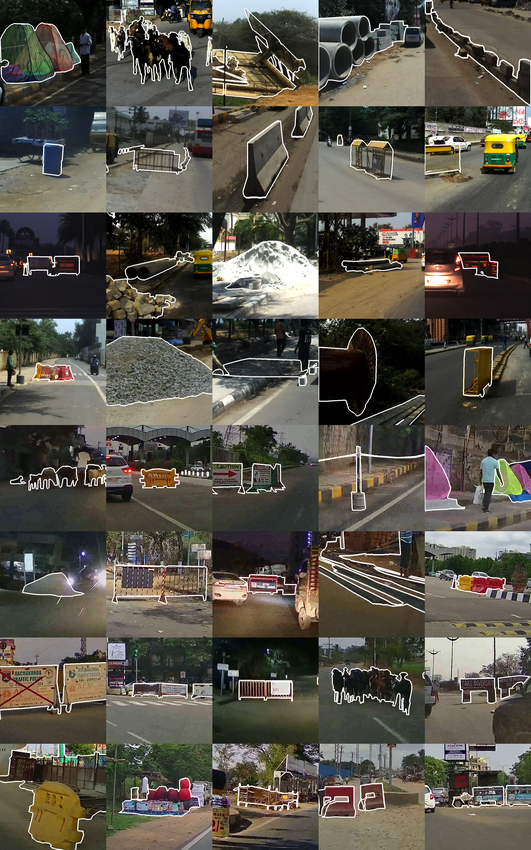}
    \caption{\textbf{Qualitative results}. Example images with anomaly objects from \IDDDatasetStatic (first 4 rows) and \IDDDatasetTemporal (bottom 4 rows).}
    \label{fig:tile}
\end{figure*}

\begin{figure*}
    \centering
    \includegraphics[width=.8\textwidth, height=0.4\textheight]{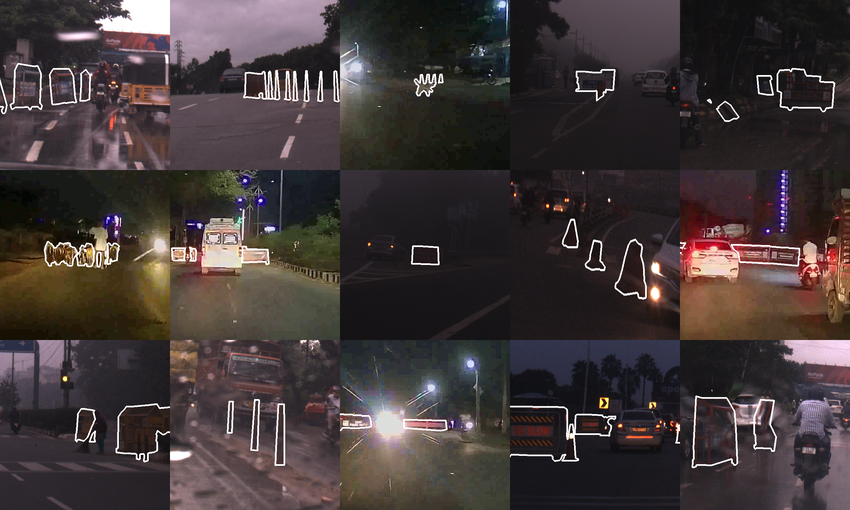}
    \caption{\textbf{Qualitative results}. Example images with anomaly objects in challenging lighting conditions.}
    \label{fig:supp_light_images}
\end{figure*}

\IDDDatasetTrain and \IDDDatasetStatic are composed from the train and validation sets of IDD~\cite{varma19wacv} which already has semantic segmentation annotations as per level-4 IDD label hierarchy that consists of 30 classes. We mapped these classes to CityScapes (C), anomaly (A) and void (V) classes as shown in Tab.~\ref{tab:class-mapping}. Certain IDD classes are mapped to multiple classes, however, the mapping is such that an input pixel can only map to one of the 3 classes (C / A / V) making the assignment unique.

The main requirement for the annotation is to ensure only the test set \IDDDatasetStatic contains anomalies. This is done by first identifying the anomaly objects in IDD and creating two subsets: one that does not include any of the listed anomaly objects forming \IDDDatasetTrain and the remaining subset forms \IDDDatasetStatic. To identify anomaly objects, we asked the annotators to find images with objects in IDD classes that are mapped to A, lies within 2 meters of the road and likely to cause damage or alter the trajectory of a vehicle. A list of such objects are mentioned in Sec.~\ref{sec:dataset_anno} and shown in Fig.~\ref{fig:tile}. The shortlisted images with anomaly objects are used to form \IDDDatasetStatic and the remaining subset constitutes \IDDDatasetTrain. Objects in A that are outside 2 meters of the road, or unlikely to adversely affect a vehicle, are annotated as void. Similarly, objects in "traffic-sign" IDD class are mapped to both C and A. Objects that are mapped to A consist of traffic cones and traffic poles that are considered anomalies in existing anomaly segmentation datasets~\cite{chan21neurips}.

 \IDDDatasetTemporal is composed using videos from IDD-X~\cite{parikh24icra}. From the original 1140 videos, we selected a subset of 103 videos that depicted the anomaly objects present in \IDDDatasetStatic. The particular clip showing the anomaly object is cropped and will be released as part of \IDDDatasetTemporal to facilitate methods to utilize temporal information. The clip is chosen in a way such that first and last frame in the clip observes the relevant anomaly. The average clip length is 8.5 seconds at a frame rate of 25 FPS resulting in around 21K images. For each clip, we selected around 10 frames for anomaly and closed-set label annotation. The frame selection is done in a way to ensure the anomaly is approximately observed at uniform temporal and spatial resolutions with respect to the ego-vehicle. The selected subset of frames are annotated into one of the 3 classes (C / A / V).

 To include images with challenging lighting conditions, we expanded \IDDDatasetTrain and \IDDDatasetStatic with images from IDD-AW~\cite{shaik24wacv}. The images in IDD-AW are also annotated as per level-4 IDD label hierarchy and consists of images collected in adverse weather conditions such as fog, rain, lowlight, snow. We excluded images collected in snow conditions due to the absence of anomaly objects. Similarly, \IDDDatasetTemporal also consists of rain and lowlight images present in the original IDD-X dataset. 
 The number of such images with challenging lighting variations is listed in Tab.~\ref{tab:supp_light_stats} and example images shown in Fig.~\ref{fig:supp_light_images}.

\section{Dataset comparison}
\label{sec:supp_metric_comp}
In~\cref{tab:supp_dataset_compare_metric}, we compare \IDDDatasetStatic and \IDDDatasetTemporal with existing datasets based on the best performance achieved by any method on the respective datasets. The results indicate that for both evaluation protocols - road obstacle (RO) and road anomaly (RA), \IDDDatasetStatic is comparably challenging to existing datasets. However, \IDDDatasetTemporal proves to be significantly more difficult, showing a notable gap in the best performance achieved.

The best values obtained for the metrics (F1 / AP / FPR) on the challenging SMIYC-RA'21~\cite{chan21neurips} , FSL\&F~\cite{blum21ijcv} are ({\color{BrickRed}{60.9~/~94.5~/~4.1}}), ({\color{BrickRed}-~/~74.8~/~2.7})~\cite{vojir2024, delic24bmvc} respectively. In comparison, the corresponding values on \IDDDatasetStatic and \IDDDatasetTemporal are ({\color{ForestGreen}{27.7~/~79.2~/~3.0}}) and ({\color{ForestGreen}{18.5~/~45.2~/~24.7}}). Given in-domain training data, \IDDDatasetStatic is as challenging as FSL\&F while being significantly diverse (\cf~\cref{tab:overview}). \IDDDatasetTemporal is much more challenging. A detailed comparison with other datasets is provided in Supplementary.

\section{Additional Qualitative Results}
\label{sec:addn_qualt}
\begin{figure*}[!ht]
\centering
    \begin{tabular}{cccc}  %

        \includegraphics[width=0.23\textwidth]{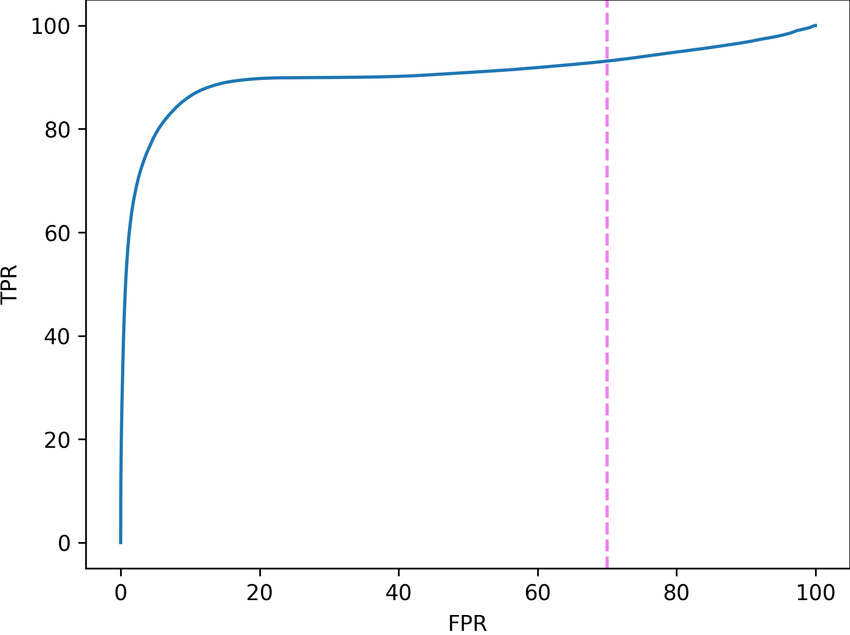}&
        \includegraphics[width=0.23\textwidth]{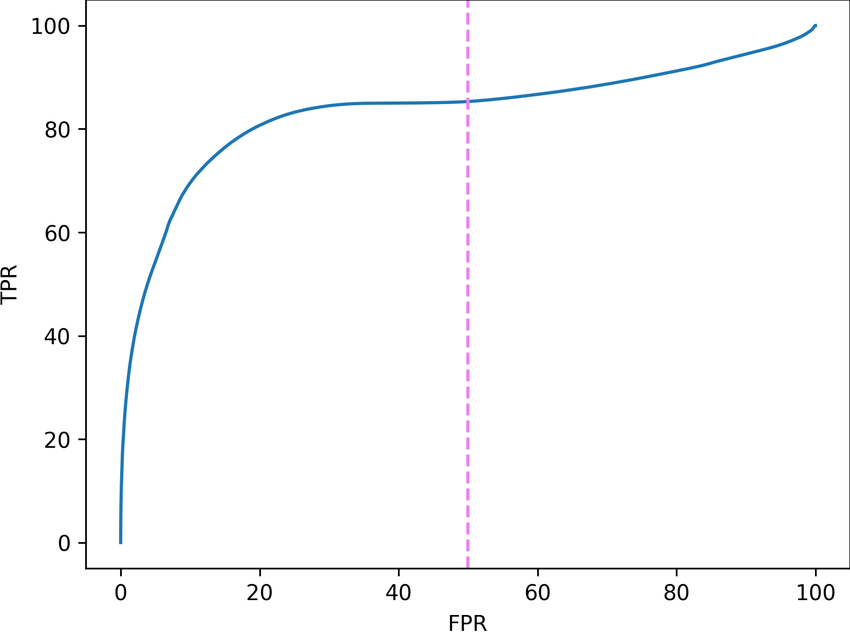}&
        \includegraphics[width=0.23\textwidth]{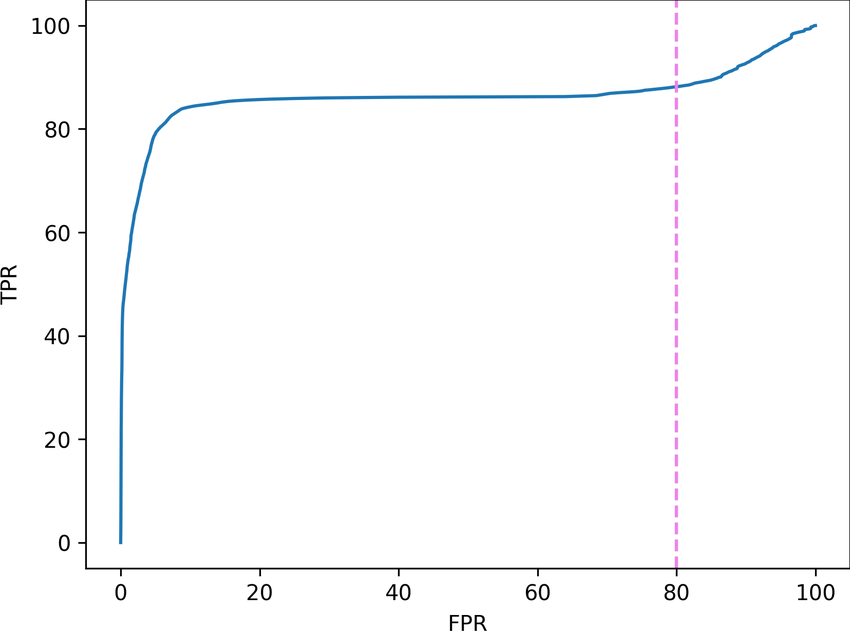}&
        \includegraphics[width=0.23\textwidth]{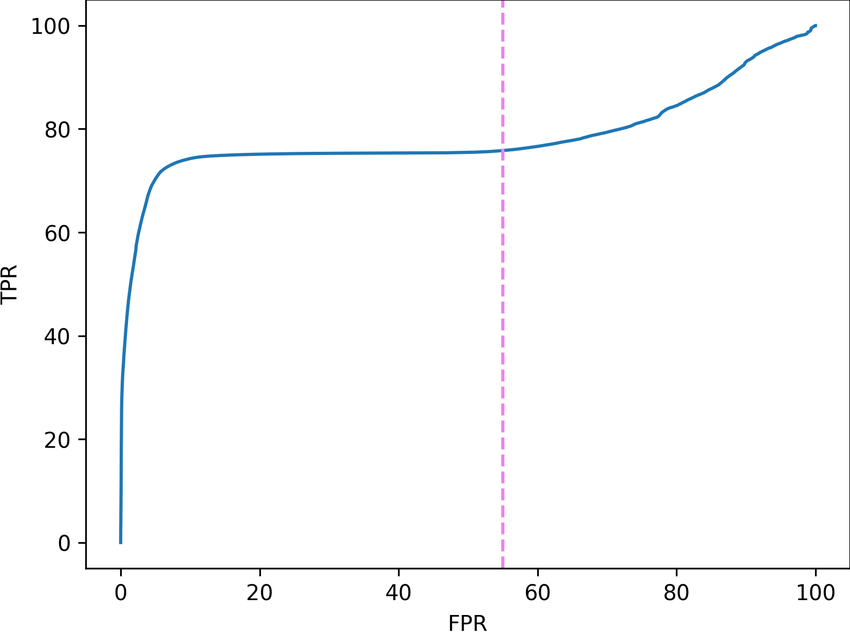} \\

        (a) RbA (\cmark) Static & (b) RbA (\cmark) Temporal & (c) UNO (\cmark) Static & (d) UNO (\cmark) Temporal  \\
        
    \end{tabular}
    \vspace{-4mm}
\caption{\textbf{ROC curves} shown for RbA (\cmark) and UNO (\cmark) across \textbf{cross-domain} setup on Static and Temporal splits. X-axis: FPR$_T$, Y-axis: TPR$_F$. Anomalies not detected until the critical point indicated by vertical line are shown in~\ref{fig:qualit_rba_op} and~\ref{fig:qualit_uno_op}.}
\vspace{-5mm}
\label{fig:qualit_roc}
\end{figure*}

\begin{figure*}[!ht]
\centering
    \begin{tabular}{cccc}  %

        \color{green}{\tiny{TP}},\color{red}{\tiny{FN}},\color{blue}{\tiny{FP}} & \color{green}{\tiny{TP}},\color{red}{\tiny{FN}},\color{blue}{\tiny{FP}} & \color{green}{\tiny{TP}},\color{red}{\tiny{FN}},\color{blue}{\tiny{FP}} & \color{green}{\tiny{TP}},\color{red}{\tiny{FN}},\color{blue}{\tiny{FP}} \\  
                
        \includegraphics[width=0.23\textwidth]{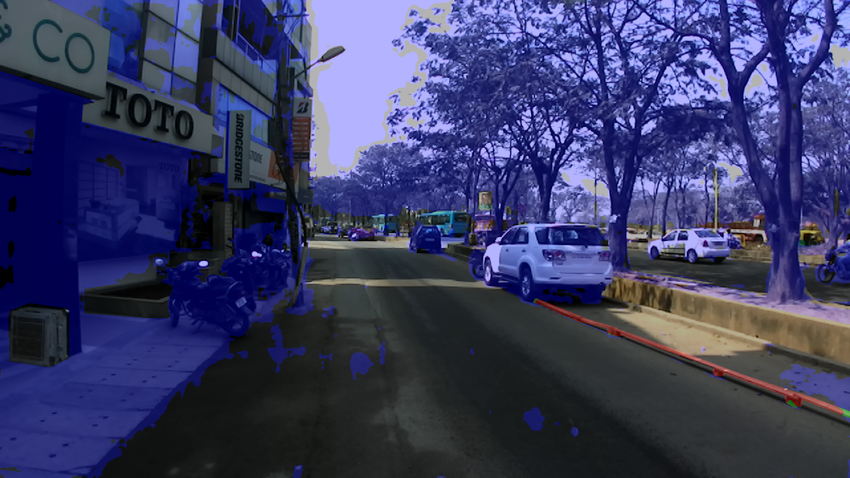}&
        \includegraphics[width=0.23\textwidth]{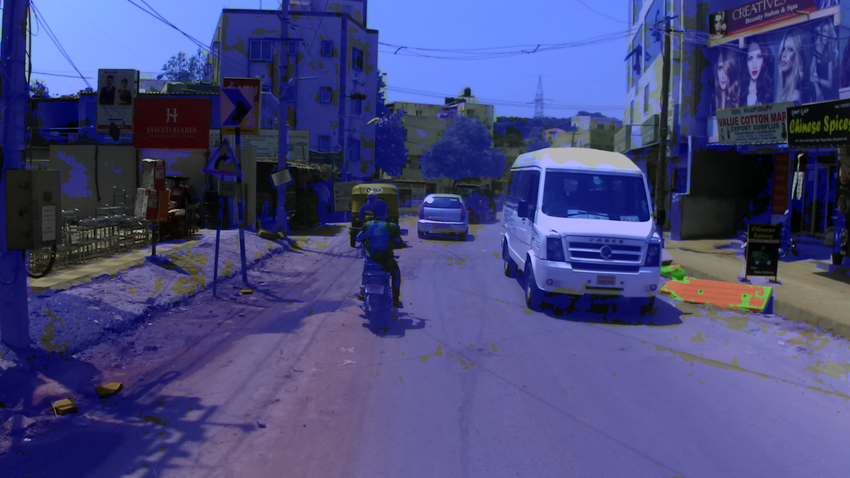}&
        \includegraphics[width=0.23\textwidth]{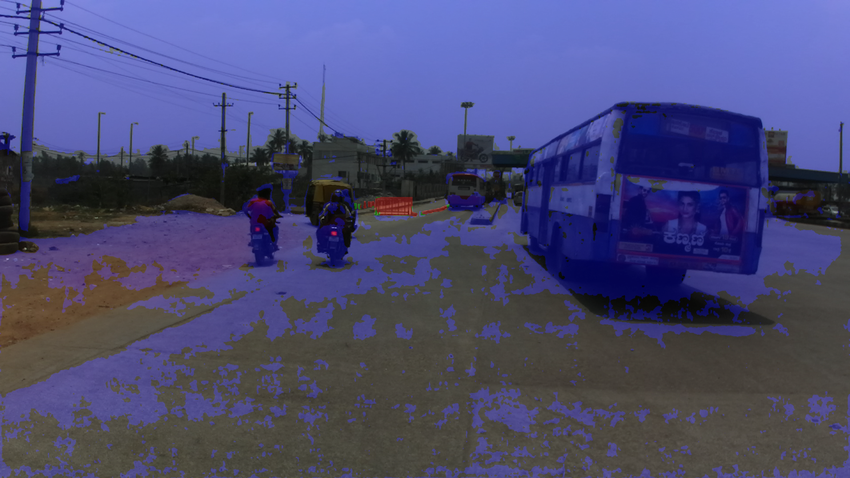}&
        \includegraphics[width=0.23\textwidth]{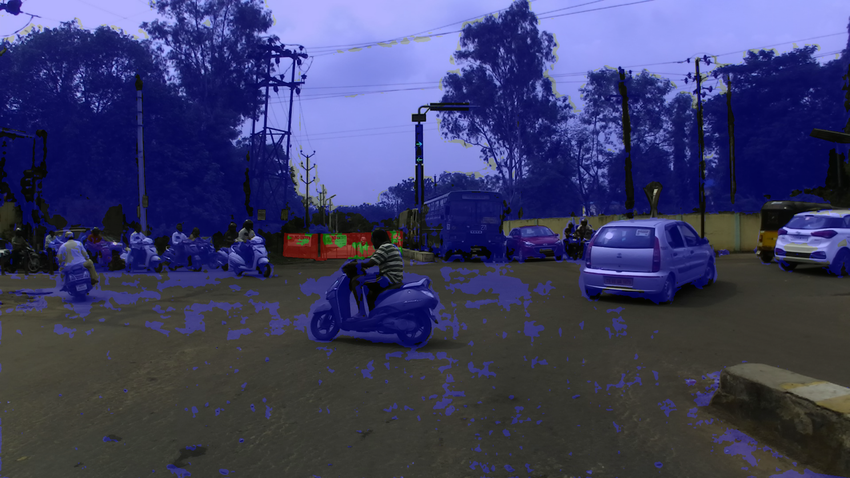} \\

        \includegraphics[width=0.23\textwidth]{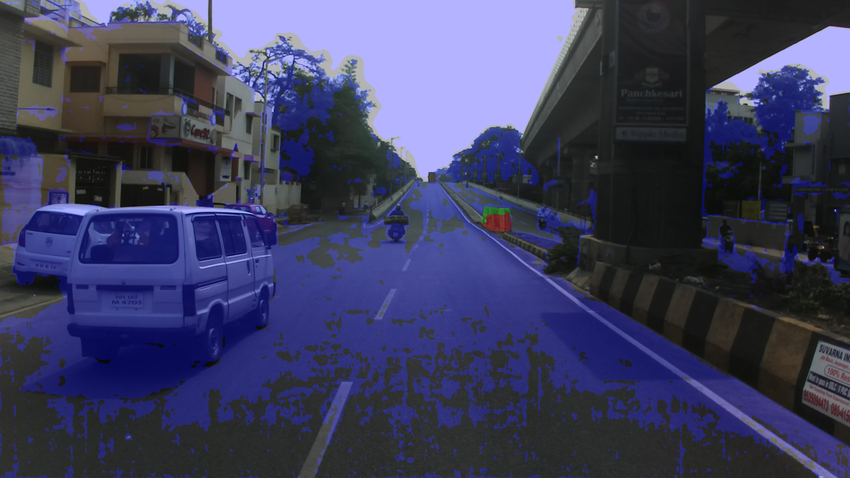}&
        \includegraphics[width=0.23\textwidth]{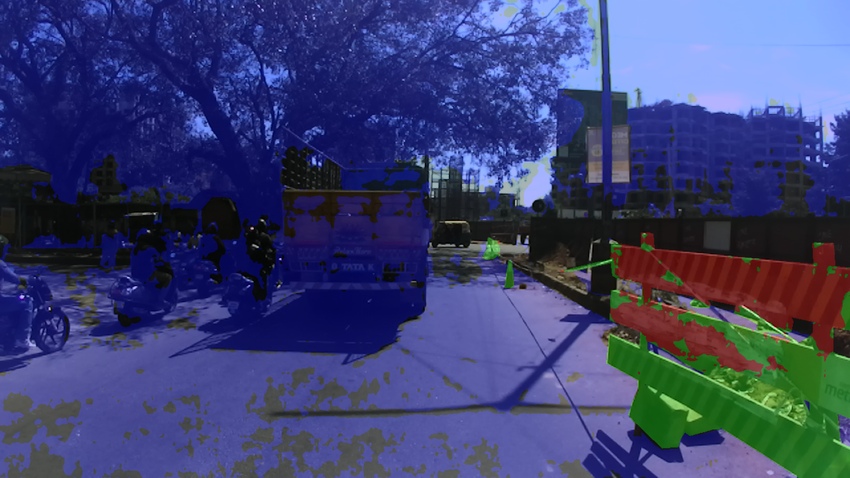}&
        \includegraphics[width=0.23\textwidth]{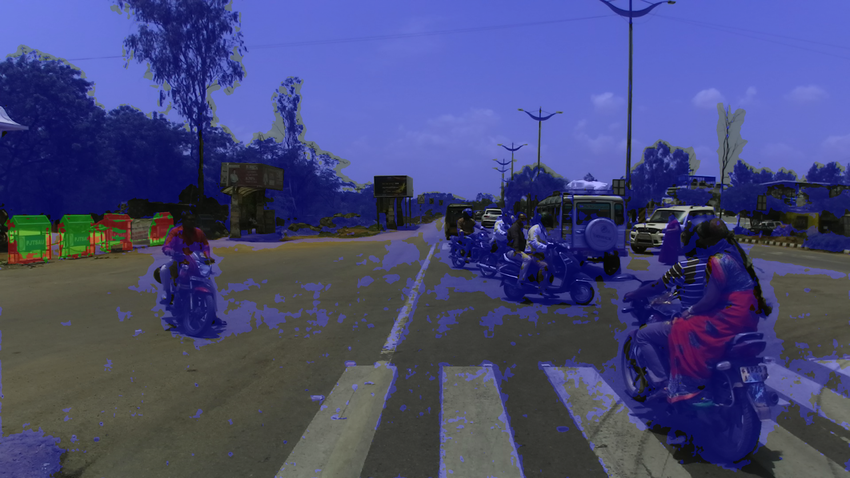}&
        \includegraphics[width=0.23\textwidth]{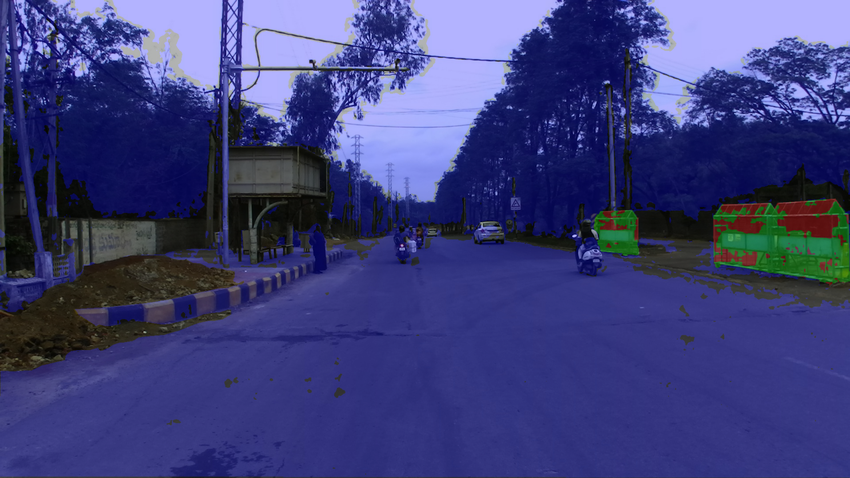} \\

        \includegraphics[width=0.23\textwidth]{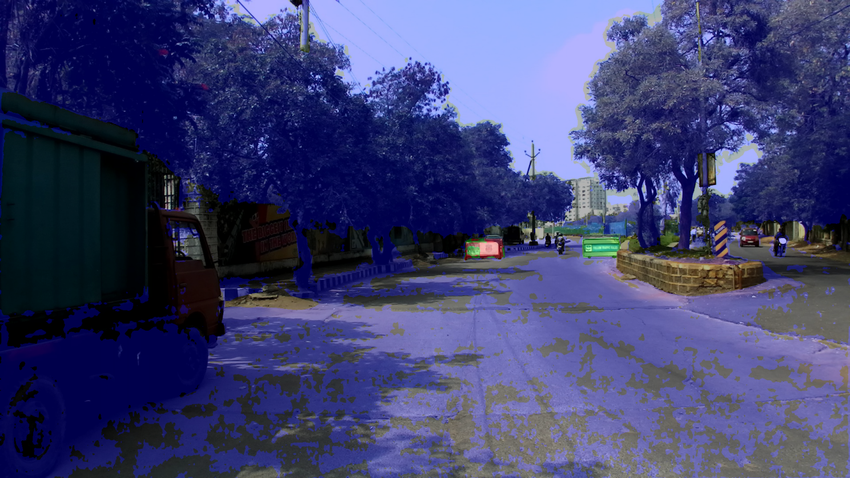}&
        \includegraphics[width=0.23\textwidth]{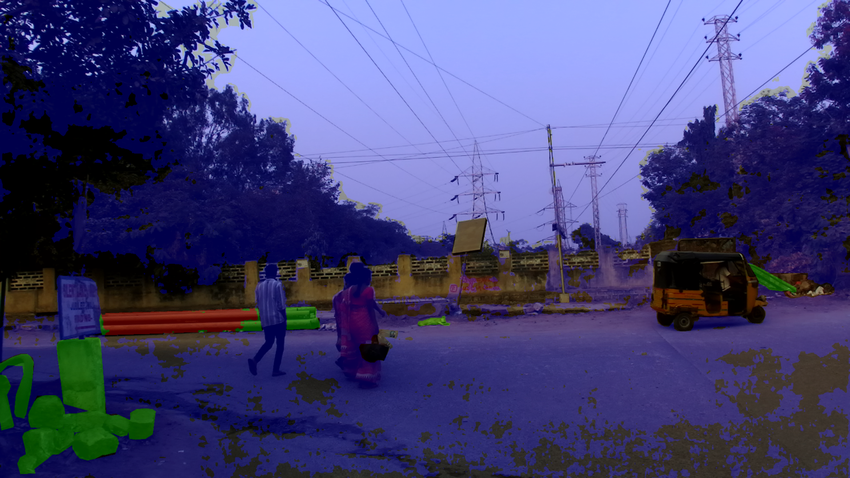}&
        \includegraphics[width=0.23\textwidth]{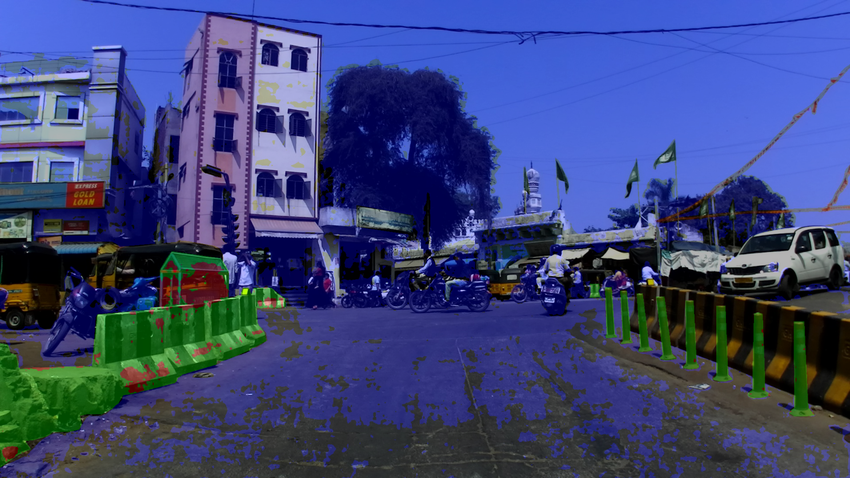}&
        \includegraphics[width=0.23\textwidth,height=0.1\textheight]{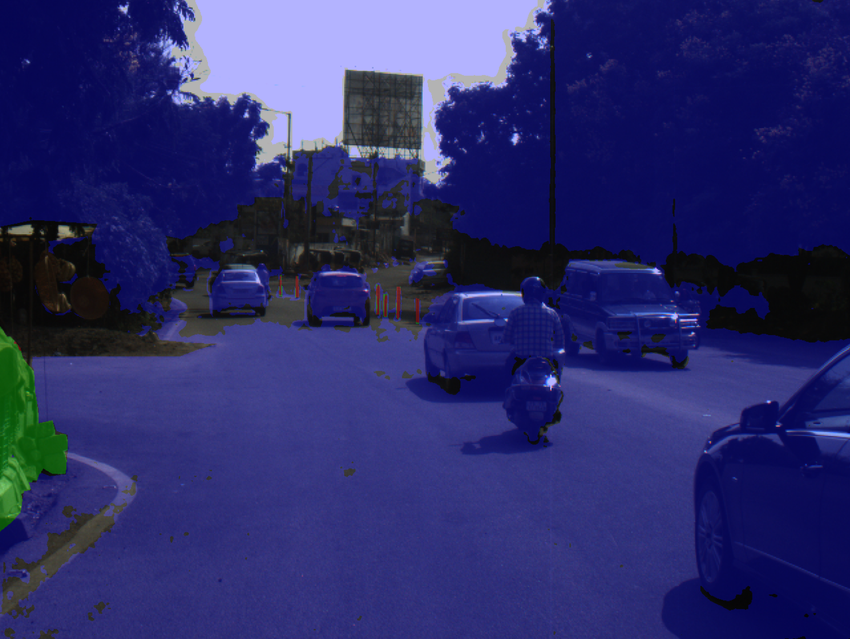} \\
        &  & \hspace{-40mm} (a) RbA (\cmark) Static  &  \\

        \color{green}{\tiny{TP}},\color{red}{\tiny{FN}},\color{blue}{\tiny{FP}} & \color{green}{\tiny{TP}},\color{red}{\tiny{FN}},\color{blue}{\tiny{FP}} & \color{green}{\tiny{TP}},\color{red}{\tiny{FN}},\color{blue}{\tiny{FP}} & \color{green}{\tiny{TP}},\color{red}{\tiny{FN}},\color{blue}{\tiny{FP}} \\  

        \includegraphics[width=0.23\textwidth]{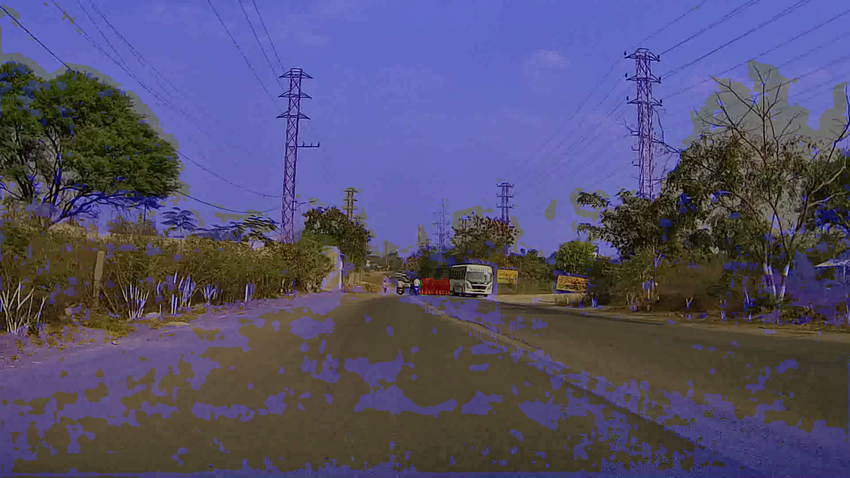}&
        \includegraphics[width=0.23\textwidth]{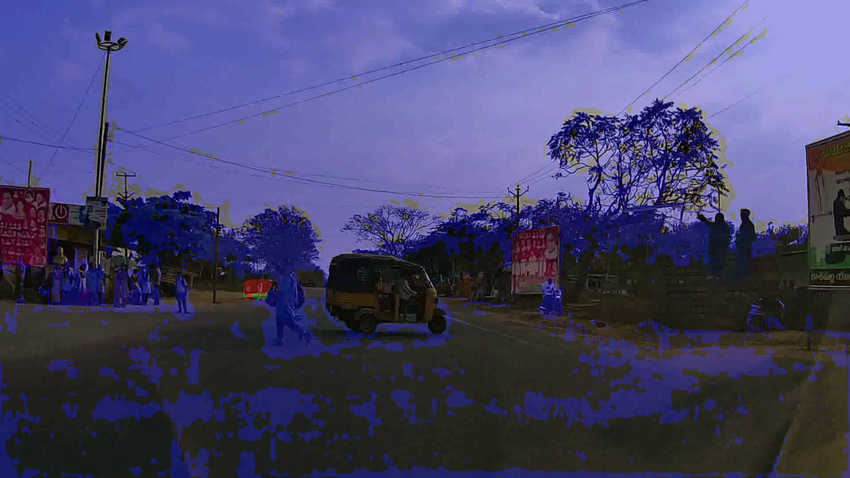}&
        \includegraphics[width=0.23\textwidth]{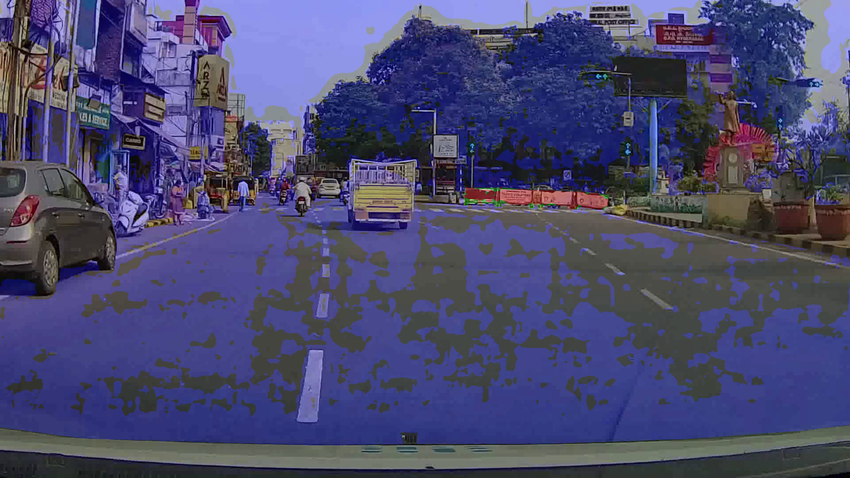}&
        \includegraphics[width=0.23\textwidth]{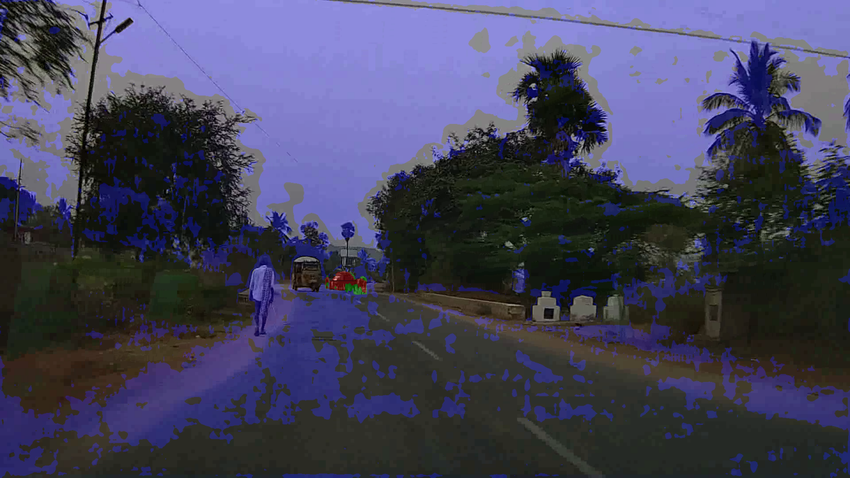} \\

        \includegraphics[width=0.23\textwidth]{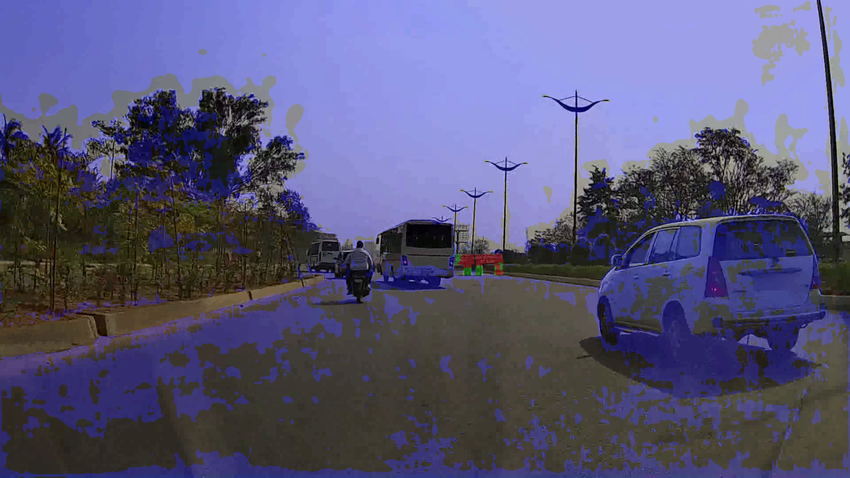}&
        \includegraphics[width=0.23\textwidth]{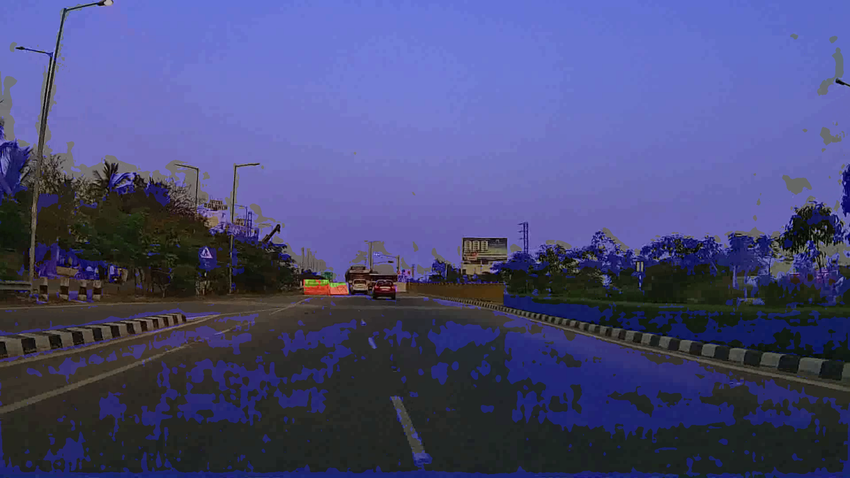}&
        \includegraphics[width=0.23\textwidth]{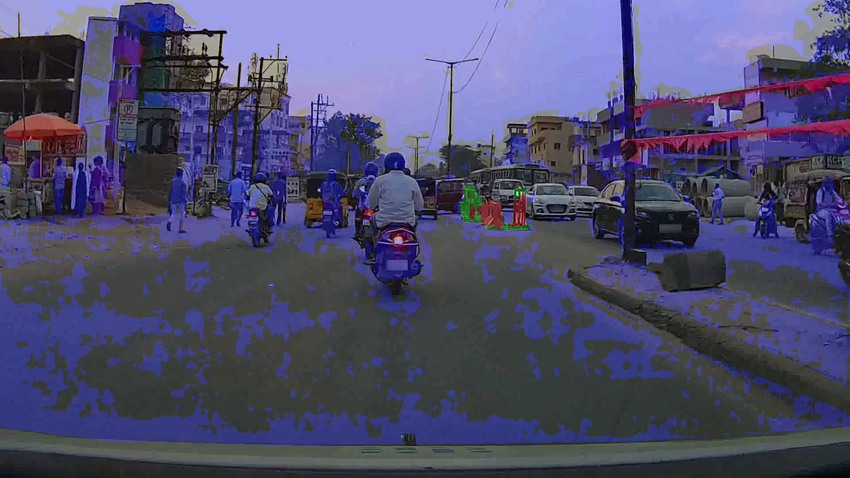}&
        \includegraphics[width=0.23\textwidth]{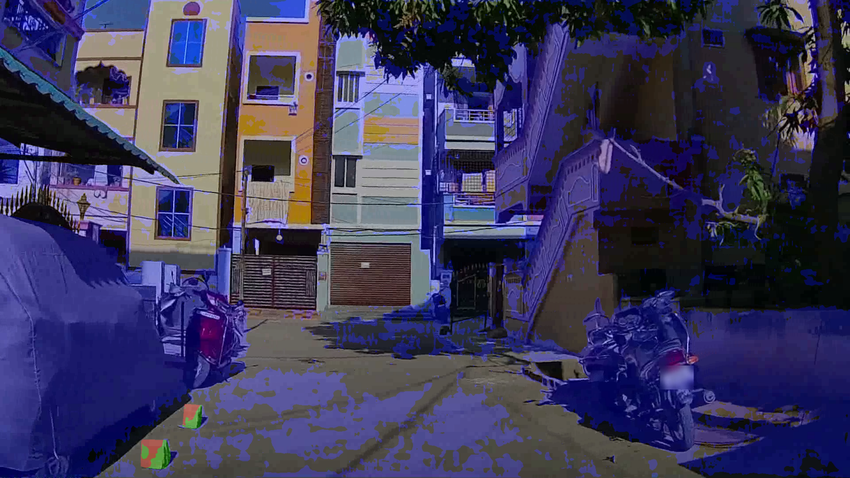} \\

        \includegraphics[width=0.23\textwidth]{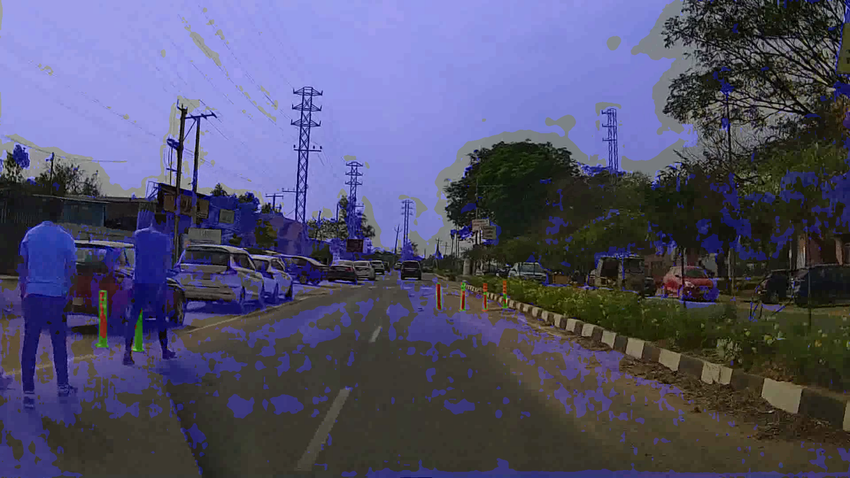}&
        \includegraphics[width=0.23\textwidth]{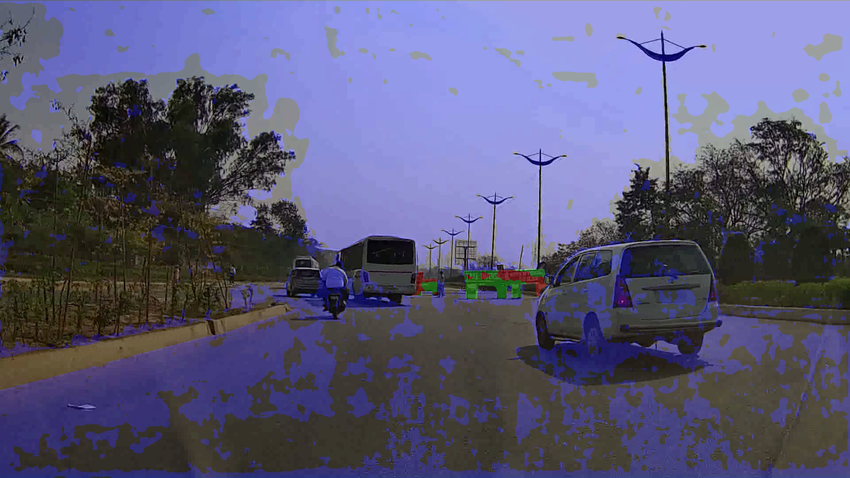}&
        \includegraphics[width=0.23\textwidth]{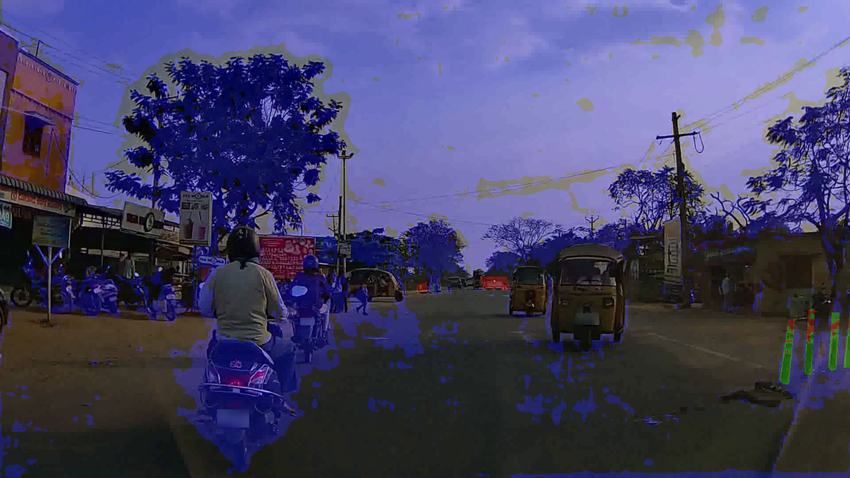}&
        \includegraphics[width=0.23\textwidth]{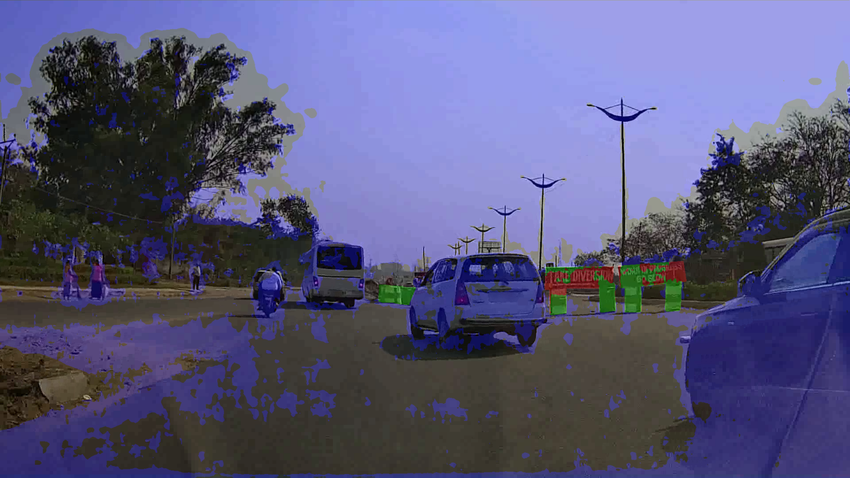} \\
        
        &  & \hspace{-40mm} (a) RbA (\cmark) Temporal  &  \\
        
    \end{tabular}
    \vspace{-4mm}
\caption{\textbf{Cross-domain qualitative results} of RbA (\cmark) in (a) Static and (b) Temporal splits. Anomaly detection threshold is set based on~\cref{fig:qualit_roc} (a) and (b).}
\vspace{-5mm}
\label{fig:qualit_rba_op}
\end{figure*}

\begin{figure*}[!ht]
\centering
    \begin{tabular}{cccc}  %

        \color{green}{\tiny{TP}},\color{red}{\tiny{FN}},\color{blue}{\tiny{FP}} & \color{green}{\tiny{TP}},\color{red}{\tiny{FN}},\color{blue}{\tiny{FP}} & \color{green}{\tiny{TP}},\color{red}{\tiny{FN}},\color{blue}{\tiny{FP}} & \color{green}{\tiny{TP}},\color{red}{\tiny{FN}},\color{blue}{\tiny{FP}} \\  
        
        \includegraphics[width=0.23\textwidth]{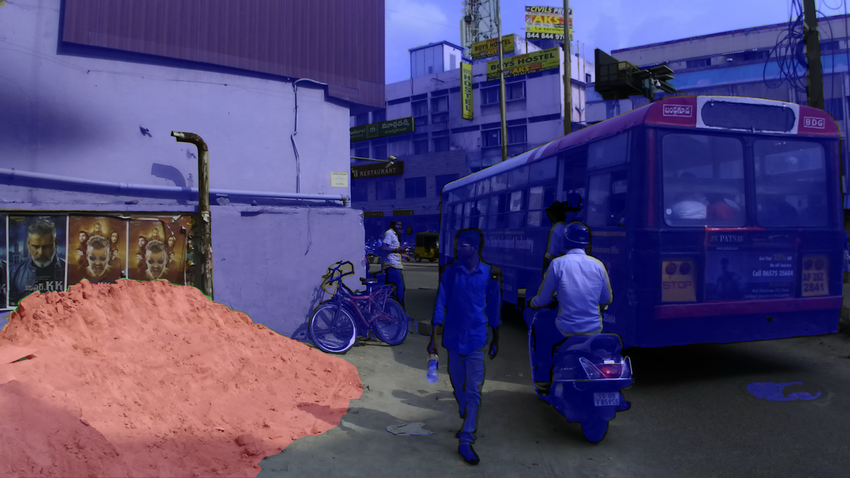}&
        \includegraphics[width=0.23\textwidth]{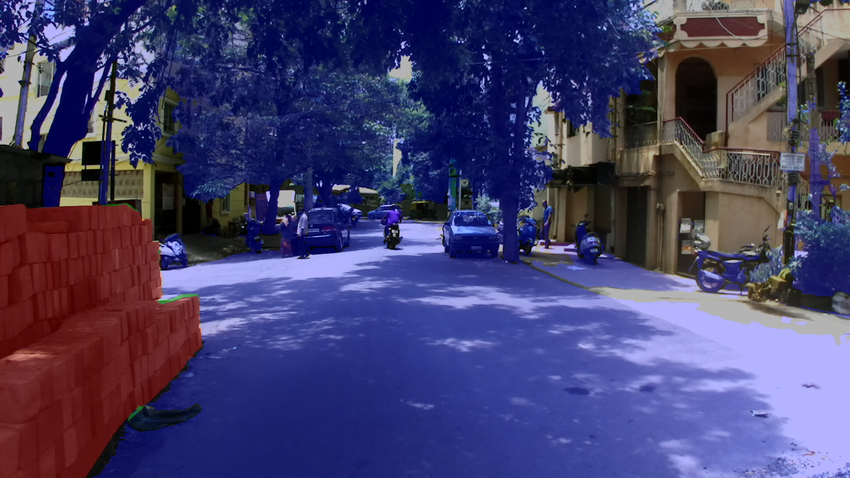}&
        \includegraphics[width=0.23\textwidth]{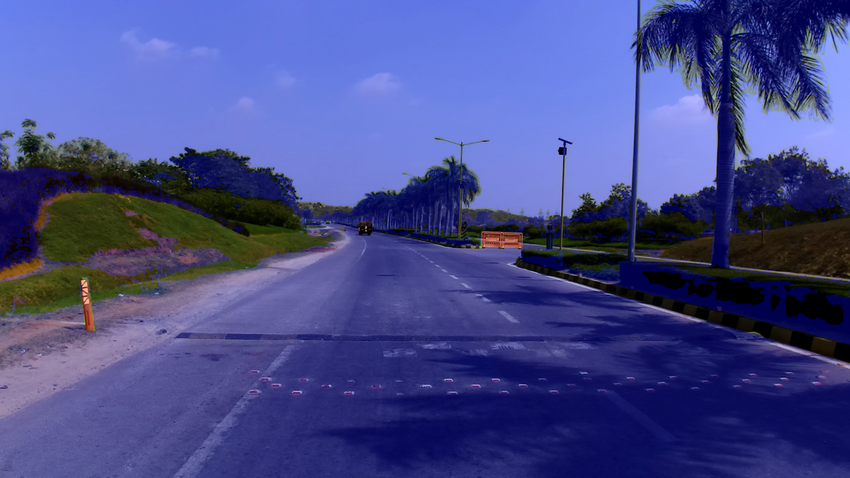}&
        \includegraphics[width=0.23\textwidth,height=0.1\textheight]{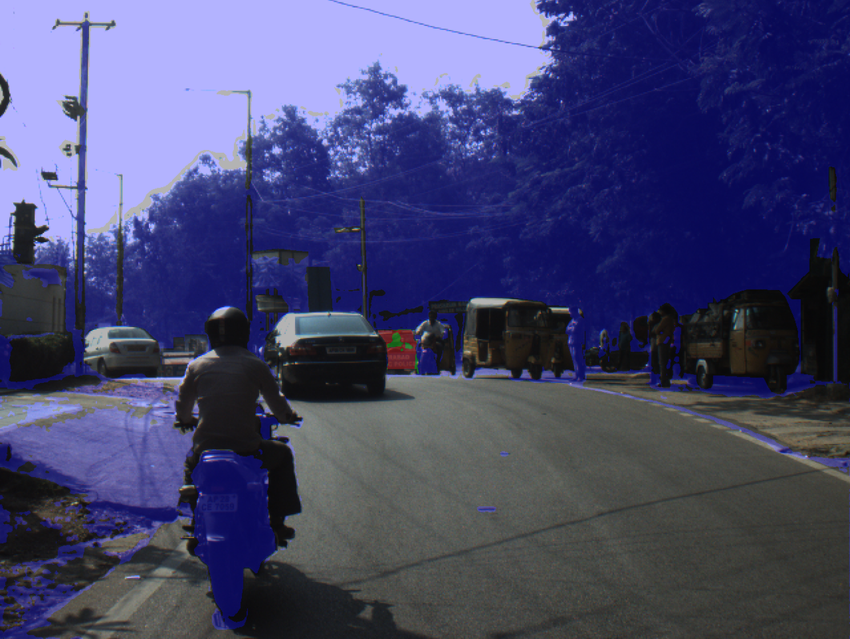} \\

        \includegraphics[width=0.23\textwidth]{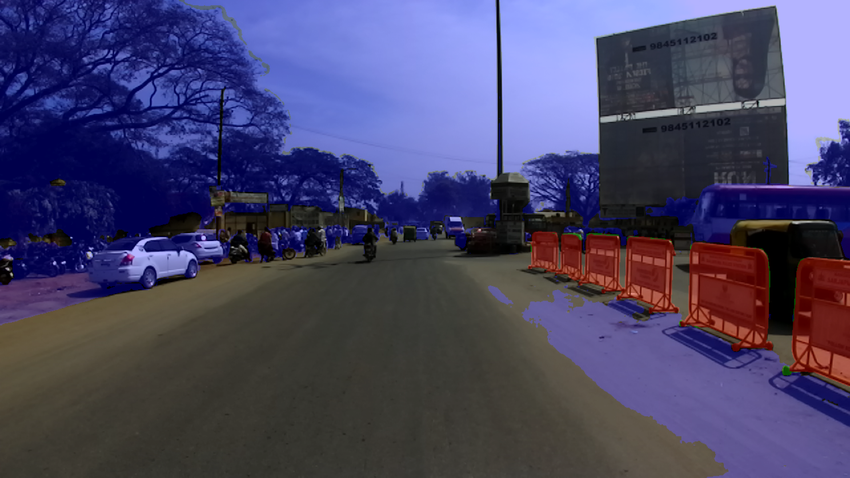}&
        \includegraphics[width=0.23\textwidth]{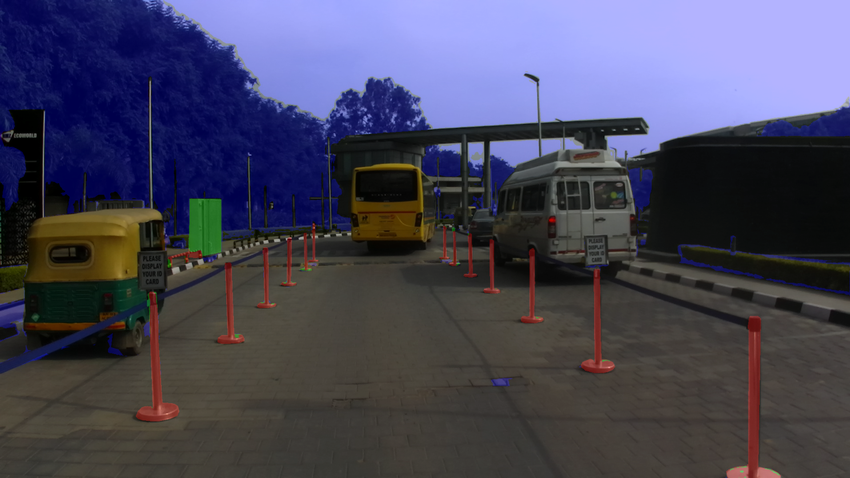}&
        \includegraphics[width=0.23\textwidth]{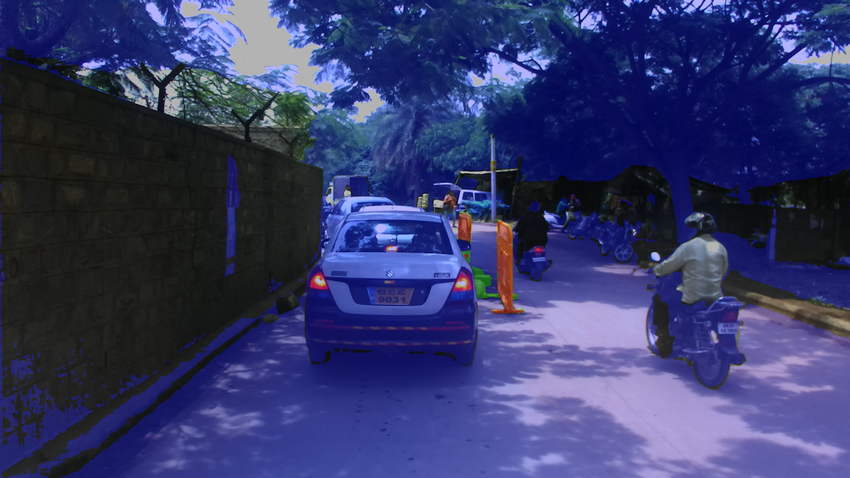}&
        \includegraphics[width=0.23\textwidth]{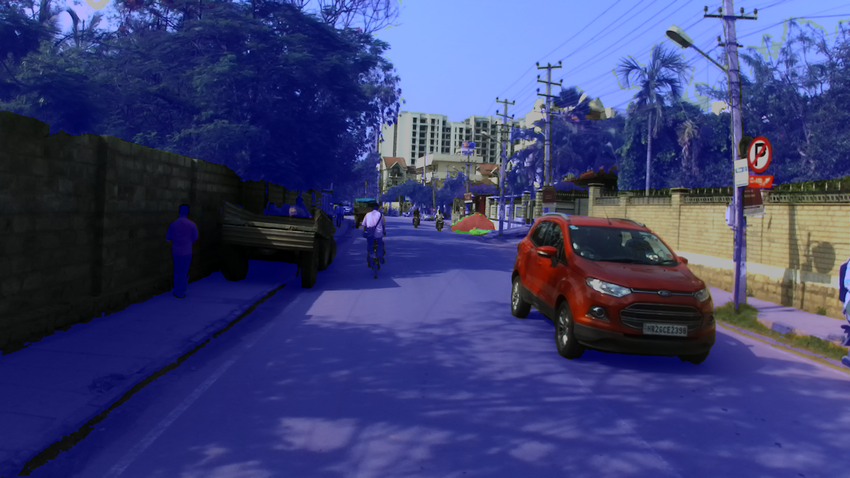} \\

        \includegraphics[width=0.23\textwidth]{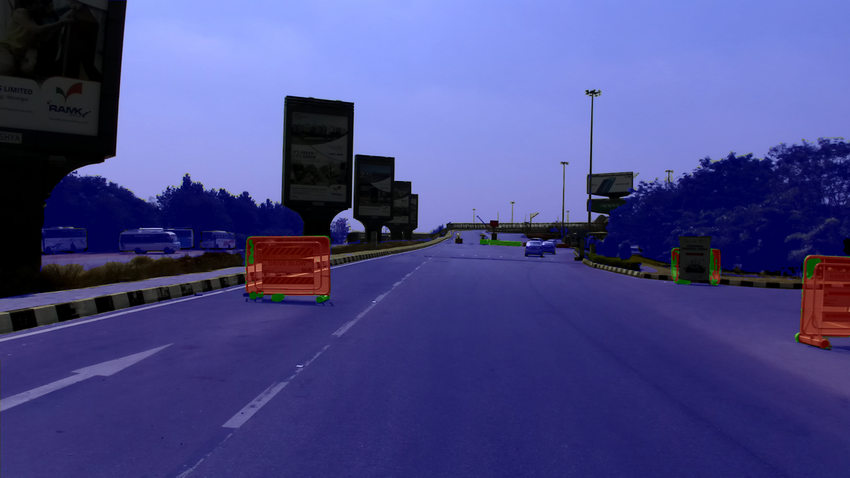}&
        \includegraphics[width=0.23\textwidth]{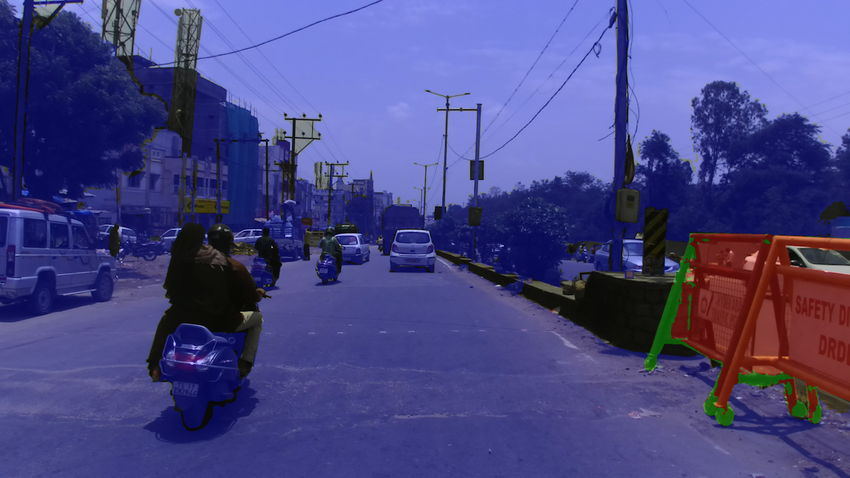}&
        \includegraphics[width=0.23\textwidth]{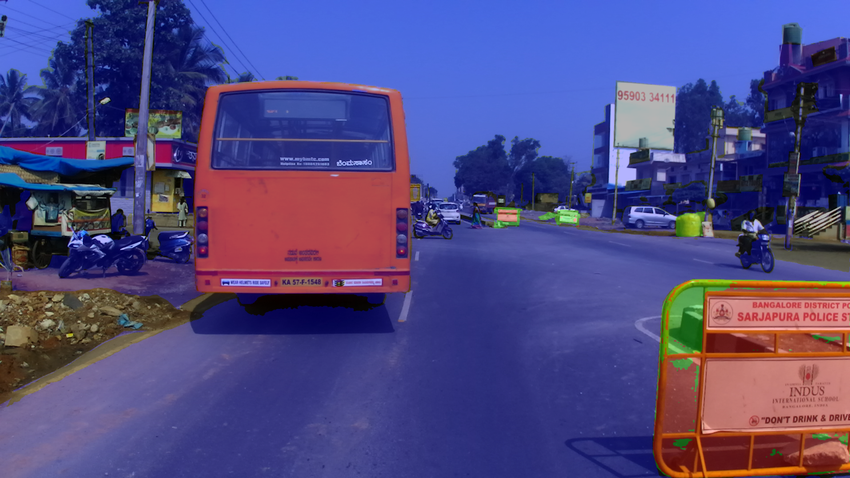}&
        \includegraphics[width=0.23\textwidth,height=0.1\textheight]{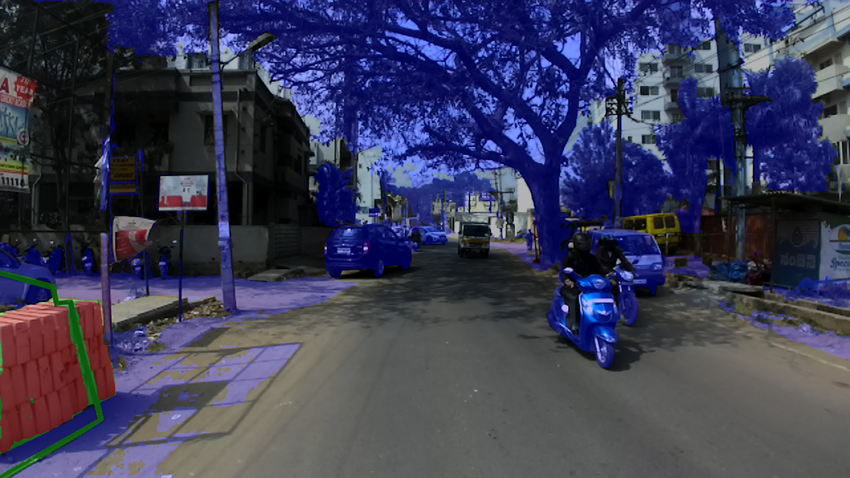} \\
        &  & \hspace{-40mm} (a) UNO (\cmark) Static  &  \\

        \color{green}{\tiny{TP}},\color{red}{\tiny{FN}},\color{blue}{\tiny{FP}} & \color{green}{\tiny{TP}},\color{red}{\tiny{FN}},\color{blue}{\tiny{FP}} & \color{green}{\tiny{TP}},\color{red}{\tiny{FN}},\color{blue}{\tiny{FP}} & \color{green}{\tiny{TP}},\color{red}{\tiny{FN}},\color{blue}{\tiny{FP}} \\  

        \includegraphics[width=0.23\textwidth]{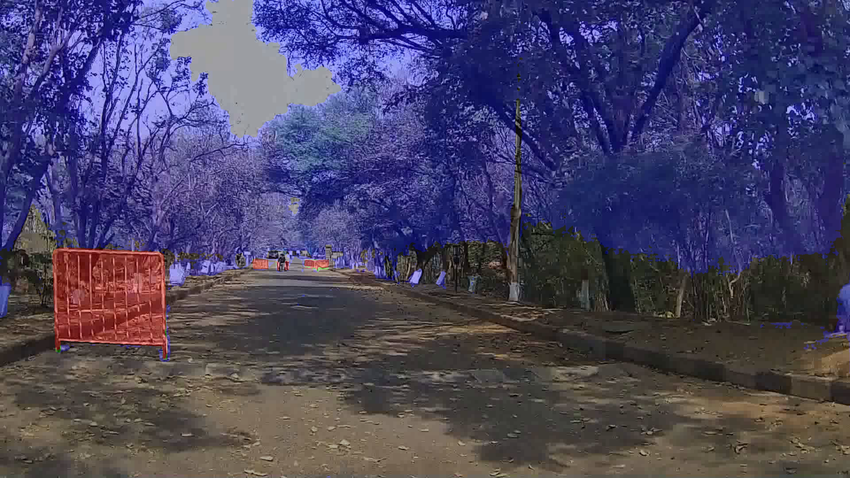}&
        \includegraphics[width=0.23\textwidth]{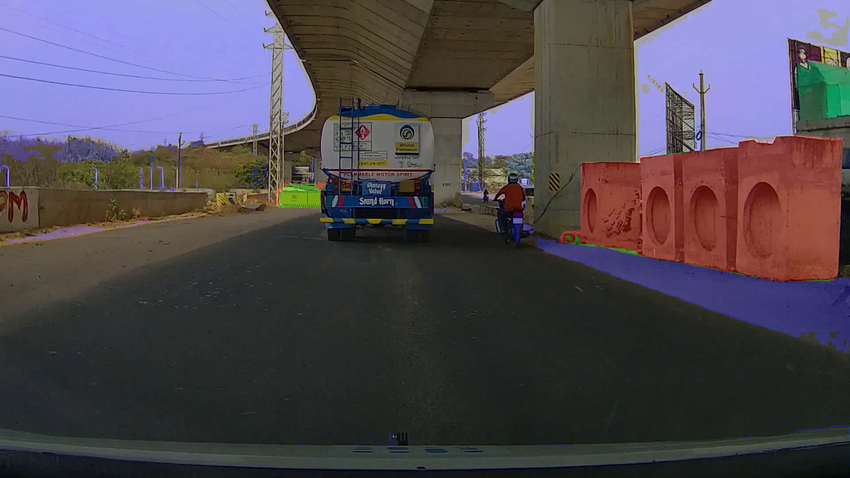}&
        \includegraphics[width=0.23\textwidth]{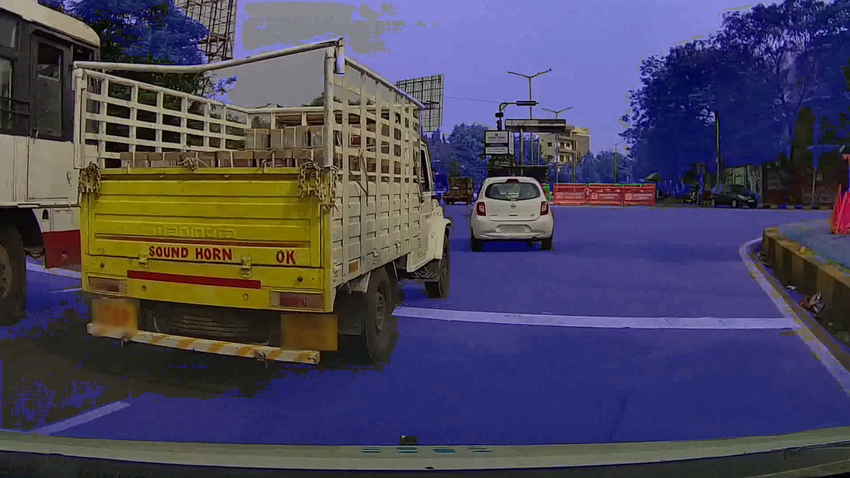}&
        \includegraphics[width=0.23\textwidth]{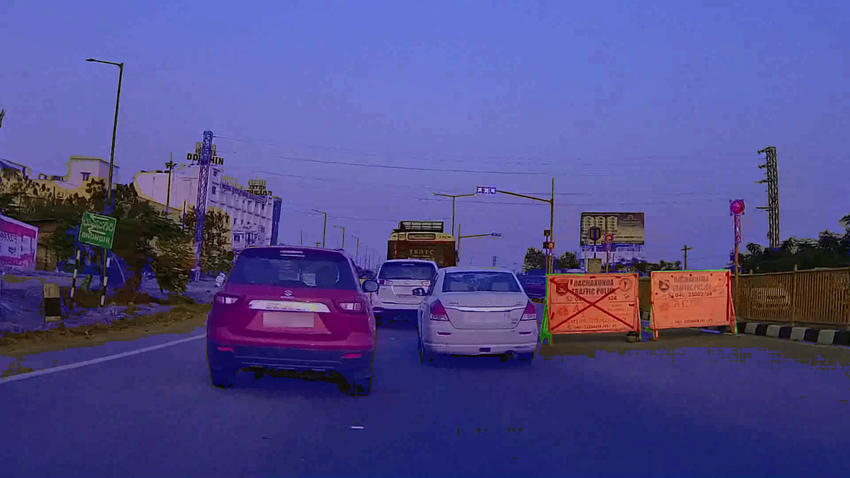} \\

        \includegraphics[width=0.23\textwidth]{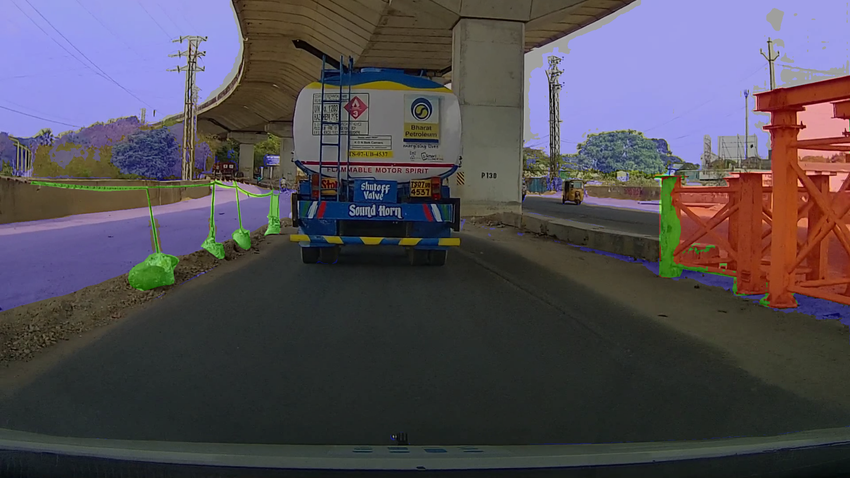}&
        \includegraphics[width=0.23\textwidth]{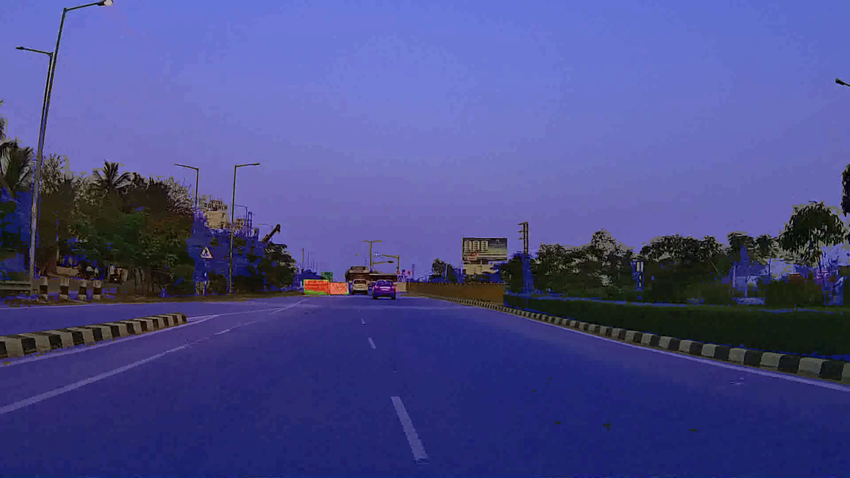}&
        \includegraphics[width=0.23\textwidth]{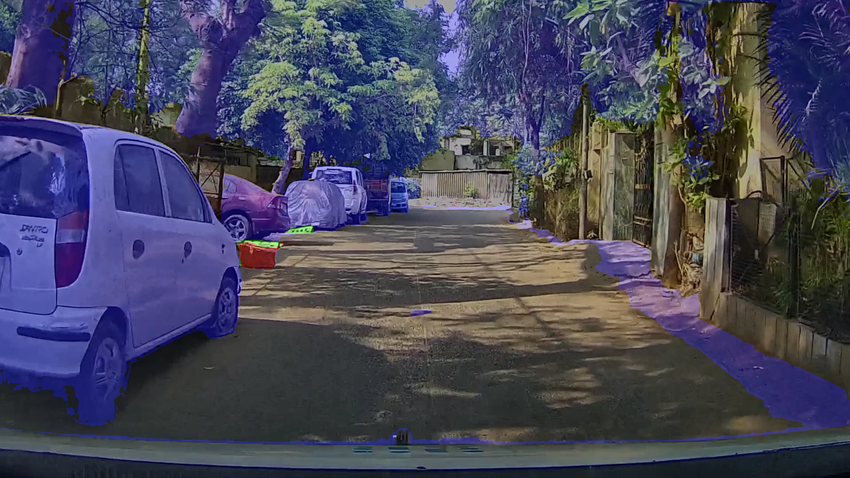}&
        \includegraphics[width=0.23\textwidth]{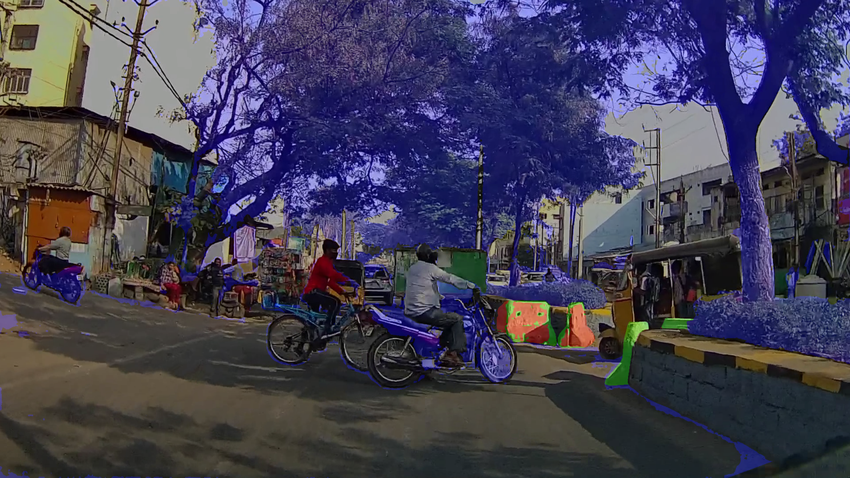} \\

        \includegraphics[width=0.23\textwidth]{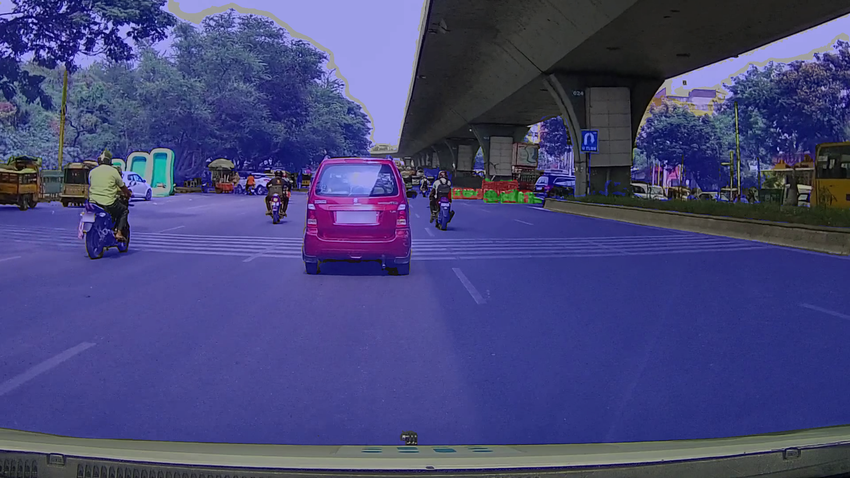}&
        \includegraphics[width=0.23\textwidth]{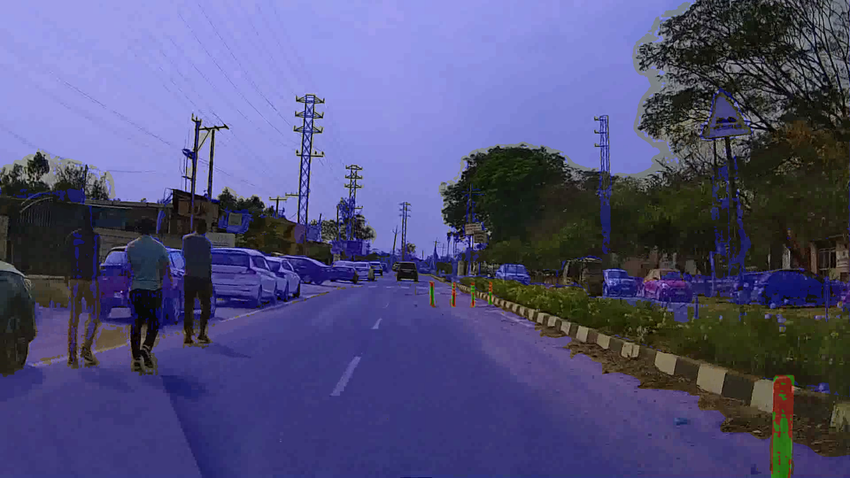}&
        \includegraphics[width=0.23\textwidth]{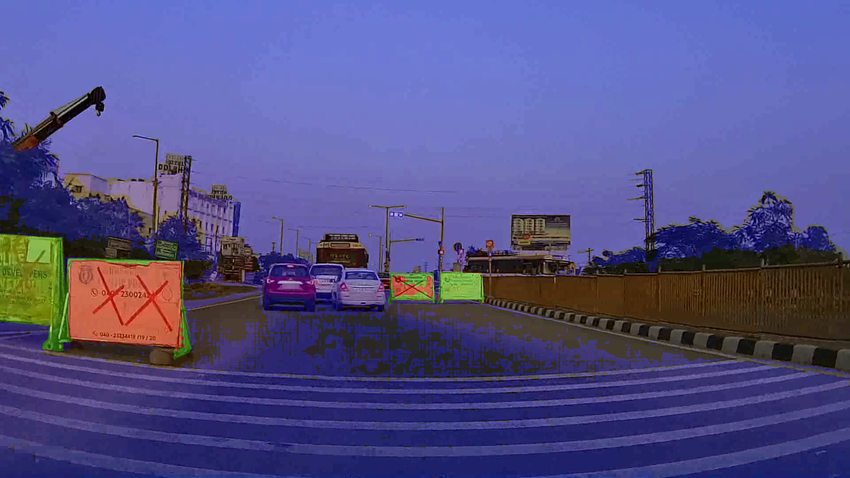}&
        \includegraphics[width=0.23\textwidth]{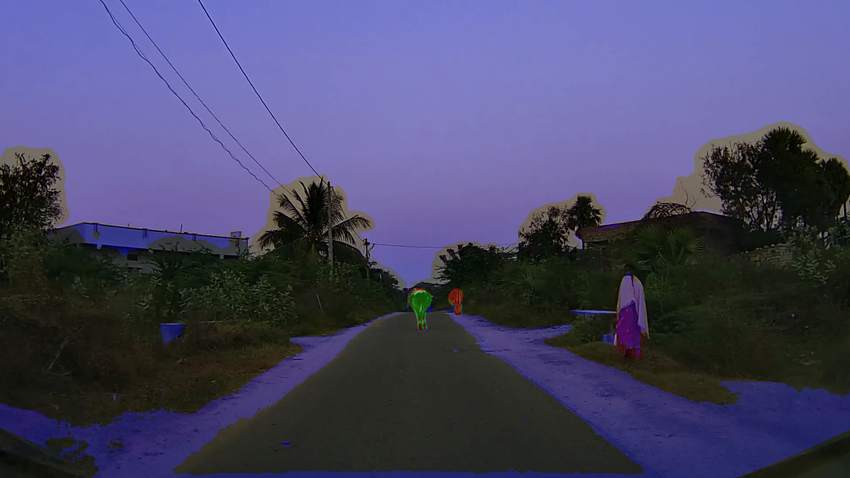} \\
        
        &  & \hspace{-40mm} (a) UNO (\cmark) Temporal  &  \\
        
    \end{tabular}
    \vspace{-4mm}
\caption{\textbf{Cross-domain qualitative results} of UNO (\cmark) in (a) Static and (b) Temporal splits. Anomaly detection threshold is set based on~\cref{fig:qualit_roc} (c) and (d).}
\vspace{-5mm}
\label{fig:qualit_uno_op}
\end{figure*}

We provide additional examples of failure cases for RbA (\cmark) and UNO (\cmark) in this section. First, we plot ROC curves of both methods in cross-domain Static and Temporal setups in~\cref{fig:qualit_roc}. Across both setups, these methods attain a TPR of ~80\% at FPR $\leq$ 15\%, beyond which the TPR deos not improve until a certain critical operating point is reached (indicated by vertical line in~\cref{fig:qualit_roc}). Examples of anomalies detected beyond this critical operating point are presented in~\cref{fig:qualit_rba_op} and~\cref{fig:qualit_uno_op}. 
\vspace{-2mm}

\vspace{-2mm}

\vspace{-2mm}

\end{document}